\documentclass[3p,times]{elsarticle}
\usepackage{amsmath}
\usepackage{xcolor}
\usepackage{booktabs}
\usepackage{subcaption}
\usepackage{verbatim}
\usepackage{multirow}

\journal{arXiv}
\begin{document}

\begin{frontmatter}
\title {Image Blind Denoising Using Dual Convolutional Neural Network with Skip Connection}

\author[mymainaddress]{Wencong Wu}

\author[mymainaddress]{Shicheng Liao}

\author[mymainaddress]{Guannan Lv}

\author[mymainaddress]{Peng Liang}

\author[mymainaddress]{Yungang Zhang\corref{mycorrespondingauthor}}
\ead{yungang.zhang@ynnu.edu.cn}

\cortext[mycorrespondingauthor]{Corresponding author}

\address[mymainaddress]{School of Information Science and Technology, Yunnan Normal University, Kunming 650500, Yunnan Province, China}

\begin{abstract}
In recent years, deep convolutional neural networks have shown fascinating performance in the field of image denoising. However, deeper network architectures are often accompanied with large numbers of model parameters, leading to high training cost and long inference time, which limits their application in practical denoising tasks. In this paper, we propose a novel dual convolutional blind denoising network with skip connection (DCBDNet), which is able to achieve a desirable balance between the denoising effect and network complexity. The proposed DCBDNet consists of a noise estimation network and a dual convolutional neural network (CNN). The noise estimation network is used to estimate the noise level map, which improves the flexibility of the proposed model. The dual CNN contains two branches: a u-shaped sub-network is designed for the upper branch, and the lower branch is composed of the dilated convolution layers. Skip connections between layers are utilized in both the upper and lower branches. The proposed DCBDNet was evaluated on several synthetic and real-world image denoising benchmark datasets. Experimental results have demonstrated that the proposed DCBDNet can effectively remove gaussian noise in a wide range of levels, spatially variant noise and real noise. With a simple model structure, our proposed DCBDNet still can obtain competitive denoising performance compared to the state-of-the-art image denoising models containing complex architectures. Namely, a favorable trade-off between denoising performance and model complexity is achieved. Codes are available at https://github.com/WenCongWu/DCBDNet.
\end{abstract}


\end{frontmatter}

\section{Introduction}\label{Introduction}
In the process of image acquisition and transmission, noise is hardly to be avoided, which seriously affects the visual quality of images, therefore noise removal is an extremely important step for many image processing tasks \cite{Chatterjee2010}. All denoising methods aim to obtain the clean image $x$ from its noisy observation $y$ by eliminating the noise $n$, namely $x = y - n$. Over the decades, many denoising methods have been proposed. For instance, the non-local means (NLM) \cite{Buades2005} uses the average of all pixels in an image to achieve noise filtering. In GNLM  \cite{LiS2016}, the grey theory \cite{Deng1982} is integrated into the NLM to further improve its denoising performance. Targeted image denoising (TID) \cite{LuoCN2015} improves denoising performance by using a dataset that contains similar image patches. Block-matching and 3-dimensional filtering (BM3D) \cite{Dabov2007} enhances the sparse representation by collaborative alteration for image denoising. The learned simultaneous sparse coding (LSSC) \cite{Mairal2009} combines the non-local means and sparse coding approaches to realize noise removal. Other denoising methods based on non-local self-similarity (NSS) \cite{Buades2008, Xu2015}, sparse learning \cite{Elad2006, Zha2017, XuZ2018}, gradient learning \cite{Osher2005, Weiss2007, Beck2009, Zuo2014}, Markov random field (MRF) \cite{Lan2006, Roth2009, Barbu2009, BarbuT2009}, total-variation (TV) \cite{Chambolle2004, Louchet2014}, and weighted nuclear norm minimization (WNNM) \cite{Gu2014, Xu2017} were also proposed.

Although the above-mentioned traditional denoising methods have achieved favorable denoising performance, most of these approaches have two shortcomings: (1) the complex optimization algorithms of these methods lead to massive running time consumption, and (2) many models require a considerable number of manual setting parameters, which increases the uncertainty of denoising results. There are some denoising methods based on image priors such as the fields of experts \cite{Roth2009}, cascade of shrinkage fields (CSF) model \cite{SchmidtR2014}, or the trainable nonlinear reaction diffusion (TNRD) \cite{Chen2017}, they do not need time-consuming optimization procedures, or less manually set parameters are required during model training, however the performances of these methods are limited by the specific forms of image priors. Inspired by the CSF and TNRD, Zhang et al. \cite{Zhang2017} proposed a successful denoising convolutional neural network (DnCNN). However, the DnCNN performs well only in a limited range of noise levels. Since the advent of DnCNN, deep neural networks (DNN) based models have become the most popular denoising methods \cite{Tian2020a, Wu2023}. However, the early DNN based denoising models like DnCNN need to train a specific model for a specific noise level, the limited flexibility of these models makes them hard to be utilized in real denoising scenes.

To make the DNN based denoising models more flexible, many techniques have been developed. Some researchers try to use a tunable noise level map as the network input \cite{Zhang2018, ZhangL2021}, therefore it is capable of handling a wide range of noise levels including spatially variant and invariant noise. However, the noise level map has to be manually pre-defined. To realize blind denoising, many models try to incorporate a noise estimation sub-network to obtain noise distribution in an image, such as the convolutional blind denoising network (CBDNet) \cite{Guo2019}, the blind universal image fusion denoiser (BUIFD) \cite{Helou2020}, and the variational denoising network (VDN) \cite{YueYZM2019}, promising denoising performances have been obtained.

Although increasing the depth of the neural networks may improve their learning ability, the problem of huge network parameters and high computational costs will also be brought, and it has been verified that the accuracy of the network will decline when the network depth is increased constantly, therefore many researchers have proposed to expand the width of the networks instead of the depth. Especially, in \cite{Pan2018, Tian2020, Tian2021}, the dual CNNs are designed for image denoising, illustrating that the competitive denoising performance can also be obtained by increasing the width of the deep neural network models.

Recently, many research works have revealed that the skip connection \cite{He2016, Huang2017}, dilated convolution \cite{Yu2016}, and the U-Net \cite{Ronneberger2015} can be helpful for deep image denoising models. The dilated convolution and the U-Net can enlarge the receptive fields of convolution layers, and skip connection has the ability to accelerate network training, also they can provide better feature preservation during feature transposing. Some recently proposed denoising models such as IRCNN \cite{ZhangZ2017}, DSNet \cite{Peng2019}, BRDNet \cite{Tian2020}, DRUNet \cite{ZhangL2021}, RDUNet \cite{Gurrola-Ramos2021}, and the DDUNet \cite{Jia2021} have achieved remarkable denoising performances by using these techniques.

Inspired by the success of the above-mentioned techniques in image denoising, in this paper, we propose a novel dual convolutional blind denoising network with skip connection (DCBDNet) to achieve effective denoising performance. The proposed DCBDNet model contains a noise estimation network and a dual convolutional neural network (CNN) with two sub-networks. In the proposed dual convolutional network, one sub-network is a u-shaped convolutional network, where the downsampling and upsampling are used. Another sub-network contains dilated convolutional layers to enlarge the receptive field of the convolution layers. Skip connections are employed in both sub-networks to fuse the features from different convolution layers. The features from the two sub-networks are then concatenated to produce the denoised image. The proposed DCBDNet owns the following favorable characteristics:

(1) A noise estimation network with a large receptive field is designed, which can extract more useful information and estimate the noise level map accurately.

(2) The dual CNN of the proposed DCBDNet utilizes downsampling operations and dilated convolutions to enlarge the receptive field. Moreover, skip connections are utilized to preserve more image details for denoising.

(3) Experimental results verify that our proposed DCBDNet achieves more robust and efficient denoising performance than similar networks on both synthetic and real noisy images.

The remainder of this paper is organized as follows. Section \ref{Related_work} gives a brief introduction of the related image denoising techniques. Section \ref{Proposed_model} introduces our proposed model. Section \ref{Experiment} presents our experimental results. The paper is concluded by Section \ref{Conclusion}.

\section{Related work}\label{Related_work}
The dual sub-network structure is one of the methods that are used to expand the width of the network, where generally two different sub-networks are designed to achieve complementary feature learning. In \cite{Pan2018}, a DualCNN contains two sub-networks was proposed for low-level vision tasks, where the shallow branch is used for obtaining the whole structure of an image, and another deeper branch is designed for capturing image details. Tian et al. \cite{Tian2021} designed a dual denoising network (DudeNet), which can extract global and local features to enhance the network performance. A batch-renormalization denoising network (BRDNet) was proposed in \cite{Tian2020}, which includes an upper and a lower sub-networks to increase the network width.

It is worth noting that the BRDNet \cite{Tian2020} has a large difference in the sizes of the receptive fields in its upper and lower sub-networks, which may degrade its learning ability and denoising performance. In addition, the dilated rate of the dilated convolutions in BRDNet is fixed, which may lead to the loss of image details, generating the gridding phenomenon. The BRDNet with two 17-layer sub-networks has a deeper network structure and more network parameters. Moreover, in BRDNet, one network is trained for only one specific noise level, which leads to its inflexibility and impracticality. In this paper, we also develop a dual CNN structure, however we aim to design a more flexible denoising network to realize blind denoising. Different from the BRDNet, our dual CNN contains fewer layers. The downsampling and upsampling are used in our upper sub-network, and the hybrid dilated convolutions are utilized in the lower sub-network, which can augment the receptive field, and the sizes of the receptive fields of the two branches can keep close. More importantly, in our model, a noise estimation network is developed for obtaining a noise level map, which makes our model fully blind.

To address the problem of model flexibility, Zhang et al. \cite{Zhang2018} proposed a fast and flexible denoising convolutional neural network (FFDNet), and a tunable or non-uniform noise level map helps FFDNet can handle a wide range of noise levels including spatially variant and invariant noise, the DRUNet \cite{ZhangL2021} also takes this type of noise level map as its input. DRUNet achieves the state-of-the-art denoising performance, however it has a large number of network parameters and a complex network structure. Moreover, since the noise distributions in a real noisy image are unknown, the manually defined noise level map used in the FFDNet and the DRUNet makes them hard to be applied in real denoising scenes.

In order to make a denoising model fully blind, many different techniques have been developed. In DeamNet \cite{Ren2021}, a dual element-wise attention mechanism (DEAM) module, and a new adaptive consistency prior (ACP) are introduced for blind image denoising. The AirNet \cite{Li2022} adopts the concept of contrast learning \cite{He2020} to recover images from multiple types of degradation. The VDIR \cite{SohC2022} utilizes variational framework for image restoration tasks. One popular and effective way to realize blind denoising is to equip a denoising model with a noise estimation sub-network, which usually is a convolution neural network has a simple structure, aiming to obtain the noise distributions in an image automatically. The noise estimation network of the CBDNet \cite{Guo2019} and VDN \cite{YueYZM2019} adopts a 5-layer fully convolutional network, the convolution kernel sizes are $3\times3\times32$ and $3\times3\times64$, respectively. Neither the pooling nor batch normalization operations are used in their noise estimation network. The noise estimator of the BUIFD \cite{Helou2020} consists of a 7-layer fully convolutional network with a convolution kernel of $5\times5\times64$, where the pooling is also not used, while the batch normalization is applied.

U-Net \cite{Ronneberger2015} was originally designed for semantic segmentation of multi-scale features, and later was utilized in the field of image denoising. A residual dense U-Net neural network (RDUNet) was presented in \cite{Gurrola-Ramos2021}, which employs densely connected convolutional layers to reuse the feature maps, the local and global residual learning is used to avoid the gradient vanishing and accelerate the network training. Jia et al. \cite{Jia2021} proposed a multi-scale cascaded U-Nets architecture called the Dense U-Net (DDUNet) for image denoising. The multi-scale dense skip connections are used for feature superposing across the cascading U-Nets, which promotes feature recycling and avoids the gradient vanishing. The DRUNet \cite{ZhangL2021} utilizes the combination of the U-Net \cite{Ronneberger2015} and ResNet \cite{He2016} as its model structure, and the state-of-the-art denoising performance was reported. The success of these U-Net based denoising models has demonstrated that U-Net can be an effective tool for image denoising.

Dilated convolution was first employed for wavelet decomposition \cite{Shensa1992}. Later, Yu et. al. \cite{Yu2016} proposed a multi-scale context aggregation network for image segmentation based on dilated convolutions, and promising results have been obtained. The dilated convolution has the ability to enlarge the receptive field, a bigger receptive field can let the model extract more context information, therefore to promote the learning ability. Many denoising models have incorporated dilated convolutions in their networks. For instance, Zhang et al. \cite{ZhangZ2017} designed an image restoration CNN (IRCNN) for image blind denoising, image deblurring, and single image super-resolution. The hybrid dilated filters \cite{Yu2017, Wang2018} are utilized in IRCNN to enlarge the receptive field and address the gridding phenomenon. Following IRCNN, a dilated residual network (DSNet) \cite{Peng2019} uses symmetric skip connections to further improve the denoising performance. The lower branch of the BRDNet \cite{Tian2020} also adopts dilated convolutions to capture more context information.

Skip connection \cite{He2016, Huang2017} is generally used for tackling the problem of gradient vanishing or exploding during back-propagating. DnCNN \cite{Zhang2017} only uses one skip connection to obtain the denoised image. Mao et al. \cite{Mao2016} proposed a very deep residual encoder-decoder network called the RED-Net for image restoration, where the skip connections are used to facilitate network training. A lightweight dense dilated fusion network (DDFN) was presented in \cite{Chen2018, Chen2020} for real-world image denoising, which tackles the vanishing or exploding of gradients by skip connections during network training. A single-stage blind real image denoising network (RIDNet) was designed in \cite{Anwar2019}, which contains the enhancement attention modules (EAM) with short and long skip connections to improve denoising performance. Anwar et al. \cite{Anwar2020} developed an identity enhanced residual denoising (IERD) network with short and long skip connections for image denoising. It can be seen that using skip connections can be helpful for model training. More importantly, the features from different layers can be fused through skip connections, therefore much more semantic and structural features in images can be preserved.

\section{The proposed model}\label{Proposed_model}

\subsection{Network architecture}
\begin{figure*}[htbp]
	\begin{center}
		\includegraphics[width=\textwidth]{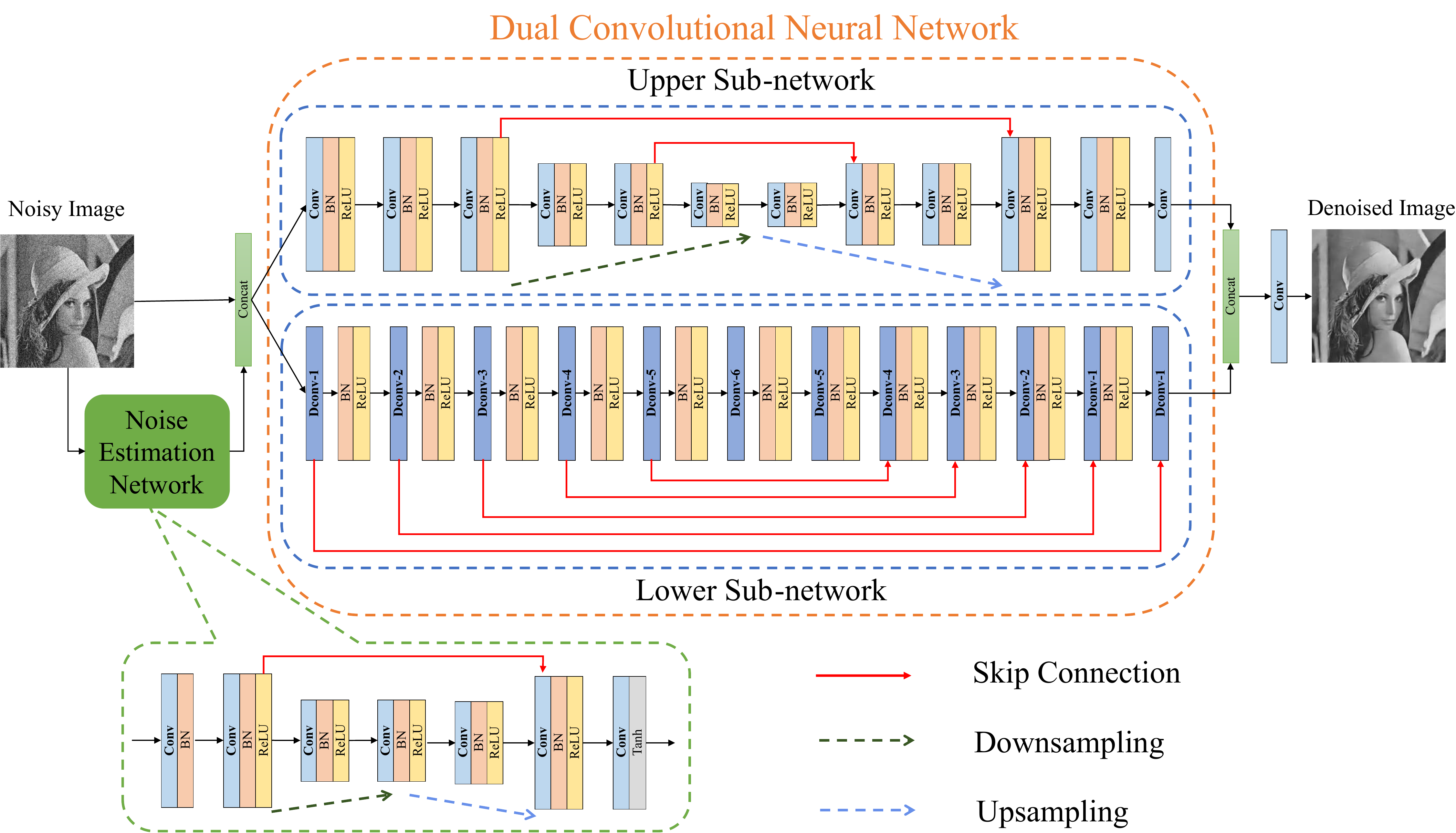}
		\caption{The network architecture of the proposed model for image denoising.}
		\label{fig:DCBDNet}
	\end{center}
\end{figure*}

Our proposed DCBDNet consists of a noise estimation network and a dual convolutional denoising network (CNN). The architecture of the proposed DCBDNet is shown in Fig. \ref{fig:DCBDNet}. The model contains two main sub-networks: a noise estimation network (the green dashed box in Fig. \ref{fig:DCBDNet}) and a dual CNN denoising network (the orange dashed box in Fig. \ref{fig:DCBDNet}). The noise level map in a noisy image is first obtained by the noise estimation network, and it will be fed into the dual CNN with the noisy image together to produce the denoised result. The proposed dual CNN contains a U-shaped upper branch, and a lower branch with dilated convolutions, skip connections are applied in both branches.

\subsubsection{Noise Estimation Network}
To achieve image blind denoising, we use a noise estimation network to estimate the noise level map. In CBDNet \cite{Guo2019} and VDN \cite{YueYZM2019}, an estimation network with 5 convolutional layers is used. In order to obtain a larger receptive field, we design a noise estimation network with 7 convolutional layers, the convolutional kernel size is $3 \times 3 \times 64$.

Our noise estimation network consists of Conv, max-pooling (Downsampling) \cite{Ranzato2007}, bilinear interpolation (Upsampling) \cite{Huang2020}, batch normalization (BN) \cite{Ioffe2015}, rectified linear unit (ReLU) \cite{Krizhevsky2012} and Tanh \cite{Malfliet1996}. The downsampling and upsampling operations are utilized to enlarge its receptive field, which can extract more complex noise information and estimate the noise level map more accurately. As the downsampling operation may lead to a loss of image information, a skip connection is used in the network to reduce the loss.

It is apparent if the estimated noise level or noise distribution does not match the noise in the input image, the denoising performance will decline. In practice, people tend to estimate a high noise level to remove more noise, however if the estimated noise level is much higher than the real one, unwanted visual artifacts will be brought. As has been verified in \cite{Zhang2018, ZhangL2021}, the orthogonal initialization of the convolutional filters \cite{Mishkin2015, Xie2017, Jia2017} is an effective way to solve this problem. Therefore the orthogonal initialization is also utilized in our network to obtain a balance between noise removal and image detail preservation.

\subsubsection{Dual Convolutional Neural Network}
Much previous work has verified that expanding the width of the network is also an effective way for improving network performance \cite{Szegedy2015}. Therefore, we propose a dual CNN contains two parallel sub-networks to expand the width of the model, which contains an upper and a lower sub-networks. The upper sub-network contains standard convolution (Conv), max-pooling (Downsampling) \cite{Ranzato2007}, bilinear interpolation (Upsampling) \cite{Huang2020}, skip connections \cite{He2016, Huang2017}, batch normalization (BN) \cite{Ioffe2015}, and rectified linear unit (ReLU) \cite{Krizhevsky2012}. The lower sub-network comprises dilated convolution (DConv) \cite{Yu2016}, skip connection, BN, and ReLU. Different types of convolution filters allow the two sub-networks can extract complementary image features \cite{ZhangT2018} and therefore enhance the generalization ability of the proposed DCBDNet.

Aiming for a desirable balance between network complexity and denoising performance, we set 12 convolutional layers for the upper and lower sub-networks respectively, and one convolutional layer after their concatenation. The convolutional filter size is set to $3 \times 3 \times 64$ for both grayscale and color images. Moreover, in order to obtain similar sizes of the receptive fields in the two sub-networks and reduce the computational cost of the upper sub-network, different from the standard U-Net, in our dual CNN only two downsampling and upsampling operations are used. Since the downsampling operations will lead to the loss of image information, two skip connections are applied for feature superposing in the upper sub-network. Furthermore, symmetric skip connections are employed in the lower sub-network to accelerate its training and improve its detail preservation. The skip connections can also avoid the gradient vanishing or exploding for back-propagating during the network training.

The downsampling in the upper sub-network and the dilated convolutions in the lower sub-network aim to widen the receptive field and keep the sizes of the receptive fields in both two sub-networks close, which can obtain more image context and detail features. Inspired by the hybrid dilated convolution (HDC) \cite{Yu2017, Wang2018}, we employ different dilated rates for each dilated convolution layer, and the rate in each layer is set to 1, 2, 3, 4, 5, 6, 5, 4, 3, 2, 1, 1, which can eliminate the gridding phenomenon and enhance the denoising performance \cite{Wang2018}. The receptive fields of upper and lower sub-networks in different layers are listed in Table \ref{tab:receptive_field}. It can be found that the two sub-networks own approximately the same size of receptive fields in the last layer. Specifically, if the receptive fields of two sub-networks have a large difference, it will lead to an increase in the computational cost of one sub-network, simultaneously the other sub-network may not extract enough image information due to the size limitation of its receptive fields, which may result in degradation of denoising performance.

\begin{table}[htbp]
\centering
\caption{The receptive fields of the upper and lower sub-networks in different layers.}
\label{tab:receptive_field}
\begin{tabular}{ccccccccccccc}
\hline
Layer & 1 & 2 & 3 & 4 & 5 & 6 & 7 & 8 & 9 & 10 & 11 & 12\\
\hline
Upper sub-network  & 30 & 34 & 38 & 48 & 56 & 74 & 90 & 106 & 122 & 138 & 154 & 170\\
\hline
Lower sub-network  & 30 & 38 & 50 & 66 & 86 & 110 & 130 & 146 & 158 & 166 & 170 & 174\\
\hline
\end{tabular}
\end{table}

\subsection{Loss function}
In this subsection, we discuss the loss function of the proposed DCBDNet. To train the network parameters of the proposed DCBDNet for the AWGN removal, we select the average mean-square error (MSE) as the optimization target, which is the most widely used optimization function. The loss function is defined in Eqn. (\ref{loss}).
\begin{equation}
\label{loss}
\begin{aligned}
    \mathcal{L}(\theta) &= \frac{1}{2K} \sum_{j=1}^K ||\mathcal{F}(y_j; \theta) - x_j||^2 \\
                        &= \frac{1}{2K} \sum_{j=1}^K ||\hat{x} - x_j||^2,
\end{aligned}
\end{equation}
\noindent where $x_j$, $\hat{x}$, and $y_j$ represent the clean, predicted, and noisy images, respectively. $\theta$ is the trainable network parameters, and $K$ is the number of clean-noisy image patches.

For the real noise with spatial variation, using MSE as a loss function will produce a blurry and over-smoothed visual effect, and the high-frequency textures may be lost due to the square penalty. Therefore, the Charbonnier loss \cite{Lai2017} is chosen as the reconstruction loss to optimize our DCBDNet. Moreover, to further enhance the fidelity and authenticity of high-frequency details when removing noise, following \cite{Jiang2020}, we employ an edge loss to constrain the high-frequency components between the ground-truth image $x$ and the denoised image $\hat{x}$. As our proposed DCBDNet includes a noise estimation network to estimate the noise level map $\sigma(y)$ in the noisy image $y$, referring to \cite{Guo2019}, we adopt a total variation (TV) regularizer to constrain the smoothness of the estimated noise level. In summary, the overall loss function of our DCBDNet is defined as:
\begin{equation}
\mathcal{L} = \mathcal{L}_{char}(\hat{x}, x) + \lambda_{edge}\mathcal{L}_{edge}(\hat{x}, x) + \lambda_{TV}\mathcal{L}_{TV}(\sigma(y)),
\label{eq.2}
\end{equation}
where $\lambda_{edge}$ and $\lambda_{TV}$ empirically are set to 0.1 and 0.05, respectively. $\mathcal{L}_{char}$ stands for the Charbonnier loss, which is defined as:
\begin{equation}
\mathcal{L}_{char} = \sqrt{\left\|\hat{x} - x\right\|^2 + \epsilon^2},
\end{equation}
where the constant $\epsilon$ is set as $10^{-3}$. The edge loss $\mathcal{L}_{edge}$ is designed as:
\begin{equation}
\mathcal{L}_{edge} = \sqrt{\left\|\bigtriangleup{(\hat{x})} - \bigtriangleup{(x)}\right\|^2 + \epsilon^2},
\end{equation}
where $\bigtriangleup$ denotes the Laplacian operator \cite{Kamgar1999}. $\mathcal{L}_{TV}$ is defined as:
\begin{equation}
\mathcal{L}_{TV} = \left\|\bigtriangledown_{h}{\sigma(y)}\right\|_{2}^{2} + \left\|\bigtriangledown_{v}{\sigma(y)}\right\|_{2}^{2},
\end{equation}
where $\bigtriangledown_{h} (\bigtriangledown_{v})$ is the gradient operator along the horizontal (vertical) direction.

\section{Experiments and results}\label{Experiment}
\subsection{Datasets}
For the AWGN removal, the DIV2K dataset \cite{Agustsson2017} is used for our DCBDNet training, the dataset contains 800 high-resolution color images as training data, and 100 high-resolution color images for validation. The size of the training images in DIV2K was re-scaled to $512\times512$, and the images were grayscaled for training the grayscale image denoising model. The training images were randomly cropped into image patches. As the patch size is extremely important for the network training, based on the architecture of our proposed DCBDNet, the size of the image patches used in our experiments is set to $180\times180$. Since a noise estimation network and two sub-networks are contained in our proposed DCBDNet, the size of the receptive field of the upper sub-network is 170, and 174 for the lower sub-network.

In general, the size of image patches should be larger than the receptive fields of convolution layers, therefore the patch size of $180\times180$ is appropriate for our DCBDNet training, for both grayscale images and color images. The patch size used here is larger than the patch sizes used in many other denoising models, which are usually set as $50\times 50$, however our experimental results illustrate that due to the larger patch size, the proposed DCBDNet can capture more image context features for improving denoising performance, especially for the images contain high noise levels. The AWGN in the noise level range of $[0, 75]$ was added to each clean image patch to obtain noisy image patches.

The Set12 and BSD68 datasets \cite{Roth2005} were used for our grayscale image denoising evaluation. The CBSD68 \cite{Roth2005}, Kodak24 \cite{Kodak24} and McMaster \cite{Zhang2011} datasets were selected for the evaluation of color image denoising. For real image denoising, the training data of the SIDD\cite{Abdelhamed2018} and RENOIR \cite{Anaya2018} datasets are selected for our model training. The SIDD training data contains 320 pairs of noisy images and the near noise-free counterparts. The RENOIR dataset is composed of 240 pairs of noisy images and the near noise-free counterparts. In order to facilitate network training, the images in these two datasets were randomly cropped into image patches of size $180 \times 180$. Moreover, the rotation and flipping operations were used for data augmentation. The SIDD validation set, DND sRGB images \cite{Plotz2017}, RNI15 \cite{Lebrun2015} and Nam \cite{Nam2016} datasets were used for evaluation. We also validated the effectiveness of our model on the Set5 dataset \cite{Bevilacqua2012} for spatially variant noise.

\subsection{Experimental settings}
All our experiments were implemented on a computer with a sixteen-core Intel(R) Core(TM) i7-11700KF CPU @ 2.50GHz, 32 GB of RAM, and an NVIDIA GeForce RTX 3080Ti GPU. The proposed DCBDNet for grayscale and color images were trained respectively. It costs about 42 hours to complete DCBDNet training. For the DCBDNet model of real denoising, it takes about 50 hours to train our model.

The model parameters are optimized by the Adam optimizer \cite{Kingma2014}. For the AWGN removal, the DCBDNet was trained for 700,000 iterations, during which the initial learning rate is $10^{-4}$ and then decreases by half every 100,000 iterations. For real image denoising, we applied 120 epochs to train the DCBDNet model, during which the initial learning rate is $2\times10^{-4}$ and is steadily reduced to $10^{-6}$ using cosine annealing strategy \cite{Loshchilov2017}. The batch size was set to 16. For other hyper-parameters of the Adam algorithm, we used the default settings.

\subsection{Ablation study}
\begin{figure*}[htbp]
	\centering
	\begin{subfigure}{0.495\linewidth}
		\centering
		\includegraphics[width=0.98\linewidth]{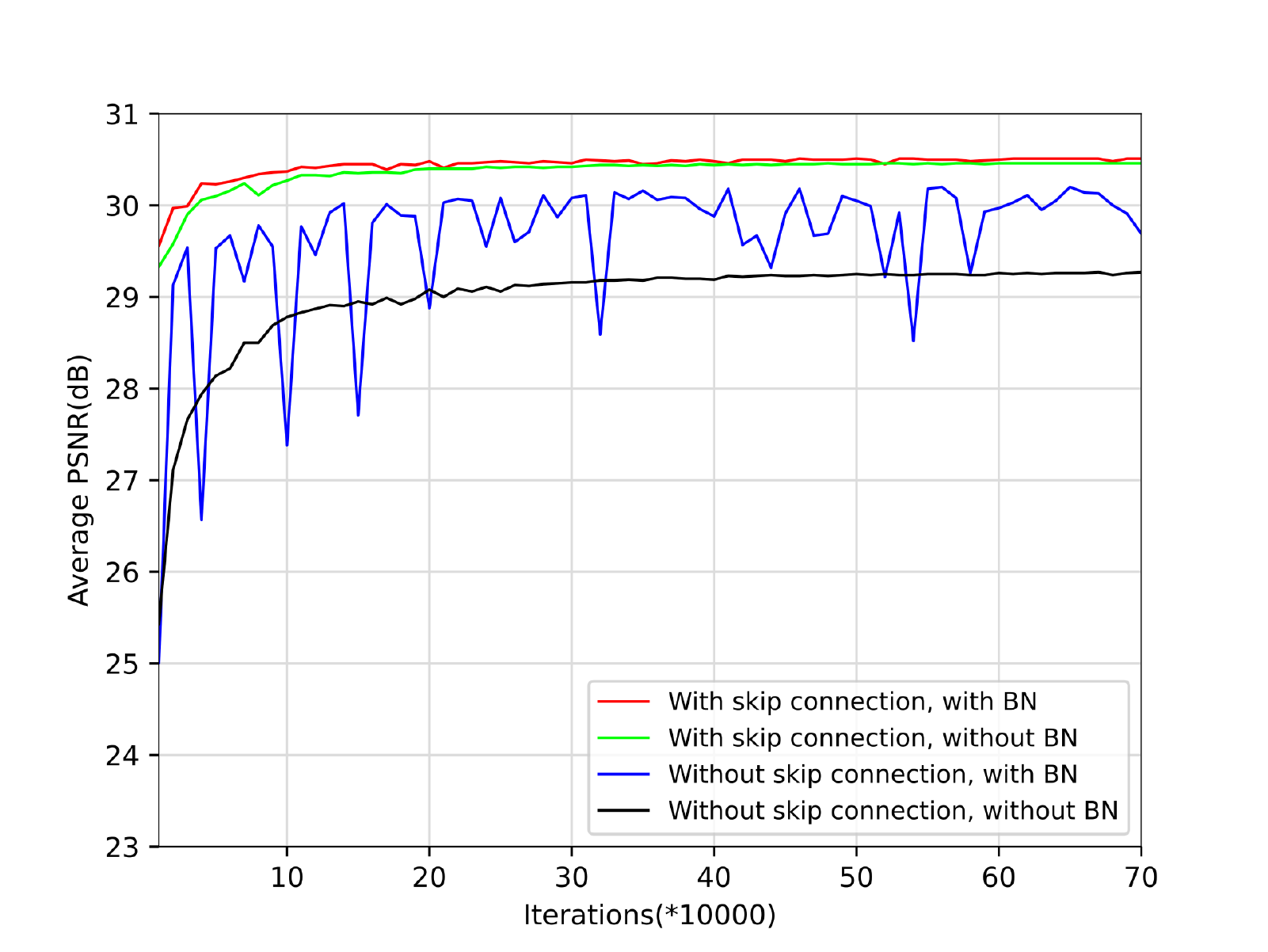}
		\caption{Grayscale image}
		\label{fig:Set12_line}
	\end{subfigure}
	\centering
	\begin{subfigure}{0.495\linewidth}
		\centering
		\includegraphics[width=0.98\linewidth]{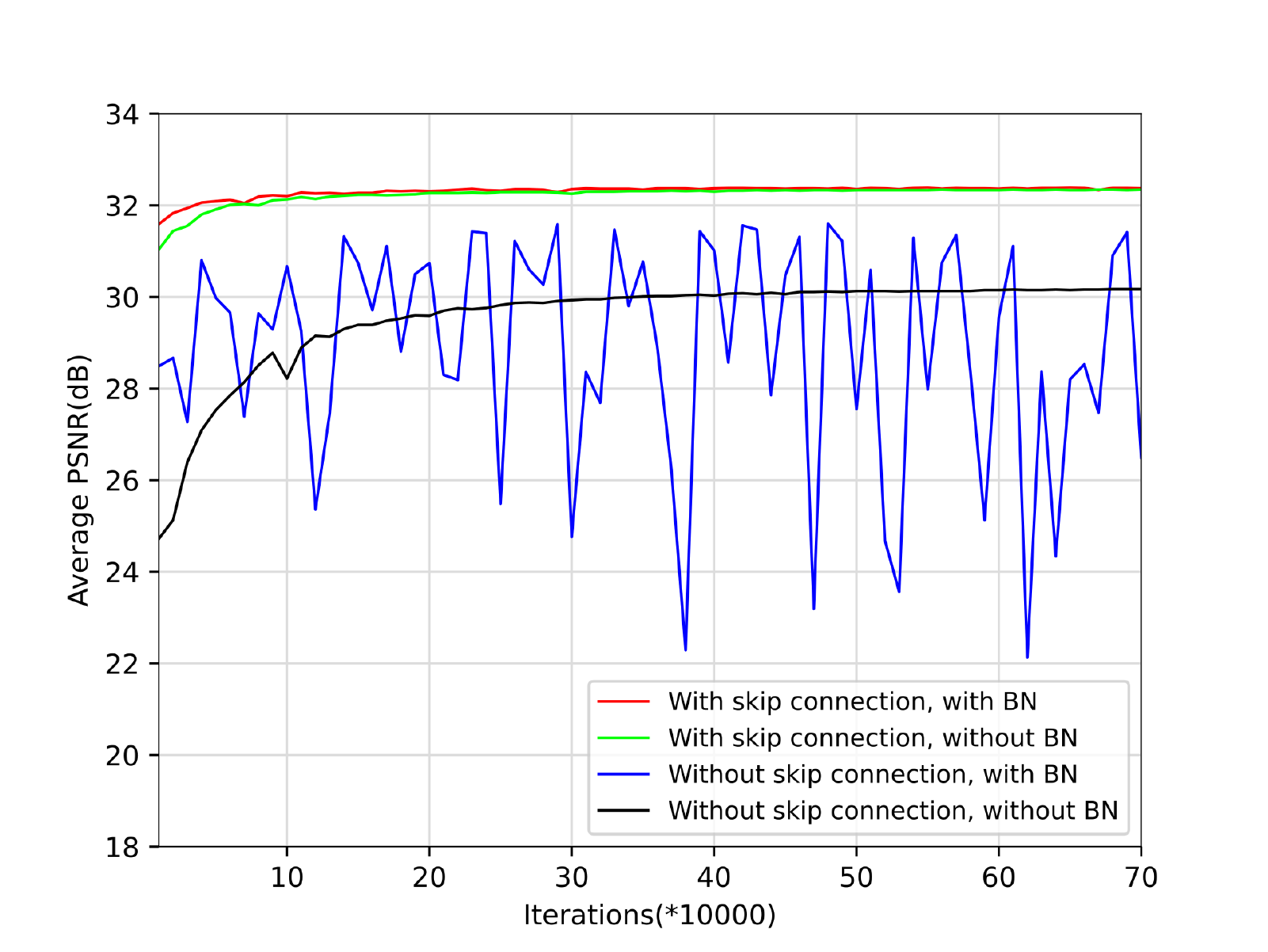}
		\caption{Color image}
		\label{fig:Kodak24_line}
	\end{subfigure}
\caption{Ablation study results of the proposed model. The average PSNR(dB) of four various network architectures on Set12 and Kodak24 datasets were recorded at noise level 25 during different iterations.}
\label{fig:line}
\end{figure*}

In order to validate the effectiveness of our proposed network architecture, especially the effects of the skip connection, we trained four different networks for grayscale and color image denoising, respectively. The models under the combinations of with/without skip connection and with/without batch normalization (BN) were trained and tested. The four different models and their corresponding performances can be seen in Fig. \ref{fig:line}. The Set12 \cite{Roth2005} and Kodak24 \cite{Kodak24} datasets were used for grayscale and color image denoising evaluation, respectively. The noise level was set to 25, and the averaged PSNR on each 10,000 iterations was recorded.

In Fig. \ref{fig:line}, it can be seen that the denoising network with skip connection and BN achieves a better denoising performance than the other three networks on both grayscale images and color images. The results have verified that using skip connection and BN in the denoising network can enhance its performance.

\subsection{Denoising Evaluation}
We evaluated the denoising performance of our DCBDNet model quantitatively and qualitatively. For quantitative analysis, we calculate the peak signal-to-noise ratio (PSNR), structural similarity index measure (SSIM) \cite{Wang2004}, feature similarity index measure (FSIM) \cite{ZhangZ2011}, the learned perceptual image patch similarity (LPIPS) as well as the DeepFeatures \cite{ZhangIESW2018} and Inception-Score \cite{Salimans2016}. For qualitative evaluation, we compared the visual effect of the denoised images from different methods.

\subsection{Evaluation of spatially invariant AWGN removal}
We first present the results of the spatially invariant AWGN removal on grayscale images. Table \ref{tab:Set12_PSNR} reports the PSNR values at different noise levels for the compared denoising methods on the Set12 dataset. It can be seen that the denoising performance of our proposed DCBDNet is slightly inferior to the DeamNet and DRUNet. However, the DeamNet and DRUNet are both equipped with complex network structures. The performance of the proposed DCBDNet is lower than the denoising results of the BRDNet on the noise levels of 15 and 25, but higher than BRDNet when the noise level is 50, which indicates that our DCBDNet is more robust for the images with high noise levels. More importantly, our proposed DCBDNet has a more compact network than BRDNet.

It should be noted that the WNNM obtained excellent denoising performances under different noise levels on the image ``Barbara''. There are rich repetitive structures exist in this image, and they can be effectively learned by the methods like WNNM which is based on non-local self-similarity learning. One can also see that the performance of TID is unsatisfied under all tested noise levels, as TID works well only on the specific datasets which contain images have similar image patches, however the Set12 dataset does not meet the above condition.

\begin{table*}[htbp]\scriptsize
\centering
\caption{The PSNR (dB) results of the compared methods on the Set12 dataset with different noise levels. The best, second best and third best are emphasized in red, blue and green respectively.}
\label{tab:Set12_PSNR}
\begin{tabular}{|c||c|c|c|c|c|c||c|c|c|c|c|c|c|}
\hline
Images & C.man & House & Peppers & Starfish &  Monar. &  Airpl. & Parrot &  Lena &  Barbara &  Boat &  Man & Couple & Average\\
\hline
\hline
Noise level & \multicolumn{13}{c|}{$\sigma$=15}\\
\hline
BM3D \cite{Dabov2007} & 31.91 & 34.93 & 32.69 & 31.14	& 31.85	& 31.07	& 31.37	& 34.26	& 33.10	& 32.13	& 31.92	& 31.10	& 32.37\\
\hline
WNNM \cite{Gu2014} & 32.17	& 35.13	& 32.99	& 31.82 & 32.71	& 31.39	& 31.62	& 34.27	& \textcolor{red}{33.60}	& 32.27	& 32.11	& 32.17	& 32.70\\
\hline
GNLM \cite{LiS2016} & - & 35.01 & 32.98 & - & - & - & - & 33.89 & - & 31.69 & 31.94 & - & - \\
\hline
TID \cite{LuoCN2015} & 24.32 & 29.29 & 24.93 & 23.88 &  24.96 &  24.30 & 23.75 &  28.92 &  23.62 &  26.20 &  27.56 & 26.52 & 25.69\\
\hline
TNRD \cite{Chen2017} & 32.19	& 34.53	& 33.04	& 31.75	& 32.56	& 31.46	& 31.63	& 34.24	& 32.13	& 32.14	& 32.23	& 32.11	& 32.50\\
\hline
DnCNN-S \cite{Zhang2017} & 32.61 & 34.97 & 33.30 & 32.20 & 33.09 & 31.70 & 31.83 & 34.62 & 32.64 & 32.42 & 32.46 & 32.47 & 32.86\\
\hline
BUIFD \cite{Helou2020} & 31.74	& 34.78	& 32.80	& 31.92	& 32.77	& 31.34 & 31.39	& 34.38	& 31.68	& 32.18	& 32.25	& 32.22	& 32.46\\
\hline
IRCNN \cite{ZhangZ2017} & 32.55 & 34.89	& 33.31	& 32.02	& 32.82	& 31.70	& 31.84	& 34.53	& 32.43	& 32.34	& 32.40	& 32.40	& 32.77\\
\hline
FFDNet \cite{Zhang2018} & 32.43	& 35.07	& 33.25	& 31.99	& 32.66	& 31.57	& 31.81	& 34.62	& 32.54	& 32.38	& 32.41	& 32.46	& 32.77\\
\hline
BRDNet \cite{Tian2020} & 32.80 & \textcolor{green}{35.27} & \textcolor{green}{33.47} & 32.24 & \textcolor{green}{33.35}	& 31.85	& \textcolor{green}{32.00} & \textcolor{green}{34.75} & 32.93 & 32.55	& \textcolor{green}{32.50}	& \textcolor{green}{32.62}	& \textcolor{green}{33.03}\\
\hline
ADNet  \cite{TianX2020} & \textcolor{green}{32.81} & 35.22 & \textcolor{blue}{33.49} & 32.17 & 33.17 & \textcolor{green}{31.86} & 31.96 & 34.71 & 32.80 & \textcolor{green}{32.57} & 32.47 & 32.58 & 32.98\\
\hline
DudeNet \cite{Tian2021} & 32.71 & 35.13 & 33.38 & \textcolor{green}{32.29} & 33.28 & 31.78 & 31.93 & 34.66 & 32.73 & 32.46 & 32.46 & 32.49 & 32.94\\
\hline
DeamNet \cite{Ren2021} & \textcolor{blue}{32.82} & \textcolor{blue}{35.77} & \textcolor{green}{33.47} & \textcolor{red}{32.55} & \textcolor{blue}{33.57} & \textcolor{blue}{31.94} & \textcolor{blue}{32.05} & \textcolor{red}{34.94} & \textcolor{green}{33.21} & \textcolor{blue}{32.67} & \textcolor{blue}{32.56} & \textcolor{blue}{32.73} & \textcolor{blue}{33.19}\\
\hline
DRUNet \cite{ZhangL2021} & \textcolor{red}{32.91} & \textcolor{red}{35.83} & \textcolor{red}{33.56} & \textcolor{blue}{32.44} & \textcolor{red}{33.61} & \textcolor{red}{31.99} & \textcolor{red}{32.13} & \textcolor{blue}{34.93} & \textcolor{blue}{33.44} & \textcolor{red}{32.71} & \textcolor{red}{32.61} & \textcolor{red}{32.78} & \textcolor{red}{33.25}\\
\hline
DCBDNet & 32.14 & 35.05 & 33.01 & 32.06 &  33.09 &  31.50 & 31.66 &  34.68 &  32.57 &  32.41 &  32.39 & 32.43 & 32.75\\
\hline
\hline
Noise level & \multicolumn{13}{c|}{$\sigma$=25}\\
\hline
BM3D \cite{Dabov2007} & 29.45 & 32.85 & 30.16 & 28.56 & 29.25 & 28.42 & 28.93 & 32.07 & 30.71 & 29.90 & 29.61 & 29.71 & 29.97 \\
\hline
WNNM \cite{Gu2014} & 29.64 & 33.22 & 30.42 & 29.03 & 29.84 & 28.69 & 29.15 & 32.24 & \textcolor{red}{31.24} & 30.03 & 29.76 & 29.82 & 30.26 \\
\hline
GNLM  \cite{LiS2016} & - & 32.91 & 30.19 & - & - & - & - & 31.67 & - & 29.71 & 29.63 & - & -\\
\hline
TID \cite{LuoCN2015} & 24.17 & 28.65 & 24.64 & 23.53 & 24.74 &  23.92 & 23.56 &  28.30 &  23.37 &  25.83 &  27.04 & 26.03 & 25.32\\
\hline
TNRD \cite{Chen2017} & 29.72 & 32.53 & 30.57 & 29.02 & 29.85 & 28.88 & 29.18 & 32.00 & 29.41 & 29.91 & 29.87 & 29.71 & 30.06\\
\hline
DnCNN-S \cite{Zhang2017} & 30.18 & 33.06 & 30.87 & 29.41 & 30.28 & 29.13 & 29.43 & 32.44 & 30.00 & 30.21 & 30.10 & 30.12 & 30.43 \\
\hline
BUIFD \cite{Helou2020} & 29.42	& 33.03	& 30.48	& 29.21	& 30.20	& 28.99	& 28.94	& 32.20	& 29.18	& 29.97	& 29.88	& 29.90	& 30.12\\
\hline
IRCNN \cite{ZhangZ2017} & 30.08 & 33.06 & 30.88 & 29.27 & 30.09 & 29.12 & 29.47 & 32.43 & 29.92 & 30.17 & 30.04 & 30.08 & 30.38 \\
\hline
FFDNet \cite{Zhang2018} & 30.10 & 33.28 & 30.93 & 29.32 & 30.08 & 29.04 & 29.44 & 32.57 & 30.01 & 30.25 & 30.11 & 30.20 & 30.44\\
\hline
BRDNet \cite{Tian2020} & \textcolor{red}{31.39}	& \textcolor{green}{33.41}	& 31.04	& 29.46	& \textcolor{green}{30.50}	& \textcolor{green}{29.20}	& \textcolor{green}{29.55} & 32.65	& 30.34	& 30.33	& \textcolor{green}{30.14}	& \textcolor{green}{30.28}	& \textcolor{green}{30.61}\\
\hline
ADNet  \cite{TianX2020} & 30.34 & \textcolor{green}{33.41} & \textcolor{blue}{31.14} & 29.41 & 30.39 & 29.17 & 29.49 & 32.61 & 30.25 & \textcolor{green}{30.37} & 30.08 & 30.24 & 30.58\\
\hline
DudeNet \cite{Tian2021} & 30.23 & 33.24 & 30.98 & \textcolor{green}{29.53} & 30.44 & 29.14 & 29.48 & 32.52 & 30.15 & 30.24 & 30.08 & 30.15 & 30.52\\
\hline
DeamNet \cite{Ren2021} & \textcolor{green}{30.41} & \textcolor{blue}{33.79} & \textcolor{green}{31.13} & \textcolor{red}{29.92} & \textcolor{blue}{30.75} & \textcolor{blue}{29.32} & \textcolor{blue}{29.62} & \textcolor{blue}{32.90} & \textcolor{green}{30.76} & \textcolor{blue}{30.51} & \textcolor{blue}{30.20} & \textcolor{blue}{30.46} & \textcolor{blue}{30.81}\\
\hline
DRUNet \cite{ZhangL2021} & \textcolor{blue}{30.61} & \textcolor{red}{33.92} & \textcolor{red}{31.22} & \textcolor{blue}{29.88} & \textcolor{red}{30.89} & \textcolor{red}{29.35} & \textcolor{red}{29.72} & \textcolor{red}{32.97} & \textcolor{blue}{31.23} & \textcolor{red}{30.58} & \textcolor{red}{30.30} & \textcolor{red}{30.56} & \textcolor{red}{30.94}\\
\hline
DCBDNet & 30.07 & 33.27 & 30.74 & 29.46 &  30.44 &  29.05 & 29.42 &  \textcolor{green}{32.70} &  30.29 &  30.31 &  30.12 & 30.21 & 30.51\\
\hline
\hline
Noise level & \multicolumn{13}{c|}{$\sigma$=50}\\
\hline
BM3D \cite{Dabov2007} & 26.13 & 29.69 & 26.68 & 25.04 & 25.82 & 25.10 & 25.90 & 29.05 & 27.22 & 26.78 & 26.81 & 26.46 & 26.72\\
\hline
WNNM \cite{Gu2014} & 26.45 & 30.33 & 26.95 & 25.44 & 26.32 & 25.42 & 26.14 & 29.25 & \textcolor{blue}{27.79} & 26.97 & 26.94 & 26.64 & 27.05\\
\hline
GNLM \cite{LiS2016} & - & 28.99 & 26.96 & - & - & - & - & 28.49 & - & 26.63 & 26.78 & - & -\\
\hline
TID \cite{LuoCN2015} & 23.39 & 27.00 & 23.73 & 22.49 &  23.43 &  23.03 & 22.75 &  26.89 &  22.80 &  24.76 &  25.72 & 24.87 & 24.24\\
\hline
TNRD \cite{Chen2017} & 	26.62 & 29.48 & 27.10 & 25.42 & 26.31 & 25.59 & 26.16 & 28.93 & 25.70 & 26.94 & 26.98 & 26.50 & 26.81\\
\hline
DnCNN-S \cite{Zhang2017} & 27.03 & 30.00 & 27.32 & 25.70 & 26.78 & 25.87 & 26.48 & 29.39 & 26.22 & 27.20 & 27.24 & 26.90 & 27.18\\
\hline
BUIFD \cite{Helou2020} & 25.44	& 29.76	& 26.50	& 24.87	& 26.49	& 25.34	& 25.07	& 28.81	& 25.49	& 26.59	& 26.87	& 26.34	& 26.46\\
\hline
IRCNN \cite{ZhangZ2017} & 26.88 & 29.96 & 27.33 & 25.57 & 26.61 & 25.89 & 26.55 & 29.40 & 26.24 & 27.17 & 27.17 & 26.88 & 27.14 \\
\hline
FFDNet \cite{Zhang2018} & 27.05 & 30.37 & 27.54 & 25.75 & 26.81 & 25.89 & 26.57 & 29.66 & 26.45 & 27.33 & 27.29 & 27.08 & 27.32 \\
\hline
BRDNet \cite{Tian2020} & \textcolor{blue}{27.44} & 30.53	& 27.67	& 25.77	& 26.97	& \textcolor{green}{25.93}	& \textcolor{green}{26.66}	& 25.93	& 26.66	& 27.38	& 27.27	& \textcolor{green}{27.17}	& 27.45\\
\hline
ADNet \cite{TianX2020} & 27.31 & \textcolor{green}{30.59} & \textcolor{green}{27.69} & 25.70 & 26.90 & 25.88 & 26.56 & 29.59 & 26.64 & 27.35 & 27.17 & 27.07 & 27.37\\
\hline
DudeNet \cite{Tian2021} & 27.22 & 30.27 & 27.51 & 25.88 & 26.93 & 25.88 & 26.50 & 29.45 & 26.49 & 27.26 & 27.19 & 26.97 & 27.30\\
\hline
DeamNet \cite{Ren2021} & \textcolor{blue}{27.44} & \textcolor{blue}{31.19} & \textcolor{blue}{27.78} & \textcolor{blue}{26.48} & \textcolor{blue}{27.19} & \textcolor{red}{26.08} & \textcolor{blue}{26.72} & \textcolor{blue}{29.96} & \textcolor{green}{27.62} & \textcolor{blue}{27.61} & \textcolor{blue}{27.38} & \textcolor{blue}{27.45} & \textcolor{blue}{27.74}\\
\hline
DRUNet \cite{ZhangL2021} & \textcolor{red}{27.80} & \textcolor{red}{31.26} & \textcolor{red}{27.87} & \textcolor{red}{26.49} & \textcolor{red}{27.31} & \textcolor{red}{26.08} & \textcolor{red}{26.92} & \textcolor{red}{30.15} & \textcolor{red}{28.16} & \textcolor{red}{27.66} & \textcolor{red}{27.46} & \textcolor{red}{27.59} & \textcolor{red}{27.90}\\
\hline
DCBDNet & \textcolor{green}{27.35} & 30.45 & 27.50 & \textcolor{green}{25.91} &  \textcolor{green}{27.01} &  \textcolor{blue}{25.94} & 26.63 &  \textcolor{green}{29.78} &  27.10 &  \textcolor{green}{27.39} &  \textcolor{green}{27.30} & 27.14 & \textcolor{green}{27.46}\\
\hline
\end{tabular}
\end{table*}

Table \ref{tab:Set12_SSIM} shows the averaged SSIM on the Set12 dataset for the compared denoising methods. It can be seen that our DCBDNet obtains competitive performance, the DeamNet and DRUNet obtain the leading results, however the DRUNet uses a manually pre-defined noise level map to obtain the results, and both the DeamNet and DRUNet have much deeper network structure than our model.

\begin{table}[htbp]
\centering
\caption{The averaged SSIM results of the compared methods on Set12 dataset with different noise levels. The best, second best and third best is highlighted in red, blue and green, respectively.}
\label{tab:Set12_SSIM}
\begin{tabular}{cccc}
\hline
Noise level & $\sigma$=15 & $\sigma$=25 & $\sigma$=50 \\
\hline
BM3D \cite{Dabov2007} & 0.896 & 0.851 & 0.766\\
\hline
WNNM \cite{Gu2014} & 0.894 & 0.846 & 0.756\\
\hline
TNRD \cite{Chen2017} & 0.896 &  0.851 & 0.768 \\
\hline
DnCNN-S \cite{Zhang2017} & 0.903 & 0.862 & 0.783 \\
\hline
BUIFD \cite{Helou2020} &  0.899 & 0.855 & 0.755 \\
\hline
IRCNN  \cite{ZhangZ2017} & 0.901 & 0.860 & 0.780 \\
\hline
FFDNet \cite{Zhang2018} & 0.903 & 0.864 & 0.791 \\
\hline
BRDNet \cite{Tian2020} & \textcolor{blue}{0.906} & 0.866 & \textcolor{green}{0.794} \\
\hline
ADNet \cite{TianX2020} & \textcolor{green}{0.905} & 0.865 & 0.791 \\
\hline
RIDNet \cite{Anwar2019} & \textcolor{blue}{0.906} & \textcolor{green}{0.867} & 0.793 \\
\hline
CDNet \cite{Quan2021} & 0.903 & 0.865 & 0.792 \\
\hline
DeamNet \cite{Ren2021} & \textcolor{red}{0.910} & \textcolor{blue}{0.872} & \textcolor{blue}{0.806} \\
\hline
DRUNet \cite{ZhangL2021} & \textcolor{red}{0.910} & \textcolor{red}{0.873} & \textcolor{red}{0.810} \\
\hline
DCBDNet & 0.902 & 0.865 & \textcolor{green}{0.794} \\
\hline
\end{tabular}
\end{table}

Table \ref{tab:BSD68_PSNR} and Table \ref{tab:BSD68_SSIM} list the averaged PSNR and SSIM results on the BSD68 dataset for the compared denoising methods, respectively. Table \ref{tab:BSD68_FSIM} lists the average FSIM on the Set12 and BSD68 datasets of the compared denoising methods. One can see that our DCBDNet model obtains competitive denoising performance.

\begin{table}[htbp]
\centering
\caption{The averaged PSNR (dB) results of the compared methods on BSD68 dataset with different noise levels. The best, second best and third best is highlighted in red, blue and green, respectively.}
\label{tab:BSD68_PSNR}
\begin{tabular}{|c|c|c|c|c|c|}
\hline
Methods & $\sigma$=15 & $\sigma$=25 & $\sigma$=35 & $\sigma$=50 & $\sigma$=75\\
\hline
BM3D \cite{Dabov2007} & 31.07 & 28.57 & 27.08 & 25.62 & 24.21\\
\hline
WNNM \cite{Gu2014} & 31.37 	& 28.83 & 27.30 & 25.87 & 24.40\\
\hline
TNRD \cite{Chen2017} & 31.42 & 28.92 &   -  & 25.97 &  -\\
\hline
DnCNN-S \cite{Zhang2017} & 31.72 & 29.23 & 27.69 & 26.23 & 24.64\\
\hline
BUIFD \cite{Helou2020} &  31.35 & 28.75	& 27.03 & 25.11 & 22.68\\
\hline
IRCNN  \cite{ZhangZ2017} & 31.63 & 29.15 & 27.66 & 26.19 &  -\\
\hline
FFDNet \cite{Zhang2018} & 31.63	& 29.19 & 27.73 & 26.29 & 24.79\\
\hline
DSNetB \cite{Peng2019} & 31.69 & 29.22 & - & 26.29 & -\\
\hline
RIDNet \cite{Anwar2019} & \textcolor{blue}{31.81} & \textcolor{green}{29.34} & - & \textcolor{green}{26.40} & -\\
\hline
BRDNet \cite{Tian2020} & \textcolor{green}{31.79} & 29.29 & - & 26.36 & -\\
\hline
ADNet  \cite{TianX2020} & 31.74 & 29.25 & - & 26.29 & -\\
\hline
DudeNet \cite{Tian2021} & 31.78 & 29.29 & - & 26.31 & -\\
\hline
CDNet  \cite{Quan2021} & 31.74 & 29.28 & \textcolor{green}{27.77} & 26.36 & \textcolor{green}{24.85}\\
\hline
DeamNet \cite{Ren2021} & \textcolor{red}{31.91} & \textcolor{blue}{29.44} & - & \textcolor{blue}{26.54} & - \\
\hline
DRUNet \cite{ZhangL2021} & \textcolor{red}{31.91} & \textcolor{red}{29.48} &  \textcolor{red}{28.02}  & \textcolor{red}{26.59} & \textcolor{red}{25.10}\\
\hline
DCBDNet & 31.65 & 29.24 & \textcolor{blue}{27.80} & 26.37 & \textcolor{blue}{24.86}\\
\hline
\end{tabular}
\end{table}

\begin{table}[htbp]
\centering
\caption{The averaged SSIM results of the compared methods on BSD68 dataset with different noise levels. The best, second best and third best is highlighted in red, blue and green respectively.}
\label{tab:BSD68_SSIM}
\begin{tabular}{cccc}
\hline
Noise level & $\sigma$=15 & $\sigma$=25 & $\sigma$=50 \\
\hline
BM3D \cite{Dabov2007} & 0.872 & 0.802 & 0.687\\
\hline
WNNM \cite{Gu2014} & 0.878 & 0.810 & 0.698\\
\hline
TNRD \cite{Chen2017} & 0.883 & 0.816 & 0.703\\
\hline
DnCNN-S \cite{Zhang2017} & 0.891 & 0.828 & 0.719 \\
\hline
BUIFD \cite{Helou2020} &  0.886 & 0.819 & 0.682 \\
\hline
IRCNN  \cite{ZhangZ2017} & 0.888 & 0.825 & 0.717 \\
\hline
FFDNet \cite{Zhang2018} & 0.890 & 0.830 & 0.726 \\
\hline
BRDNet \cite{Tian2020} & \textcolor{green}{0.893} & \textcolor{green}{0.831} & \textcolor{green}{0.727} \\
\hline
ADNet  \cite{TianX2020} & 0.892 & 0.829 &  0.722 \\
\hline
RIDNet  \cite{Anwar2019} &  \textcolor{green}{0.893} & \textcolor{blue}{0.833} & \textcolor{green}{0.727} \\
\hline
CDNet \cite{Quan2021} & 0.892 & \textcolor{green}{0.831} & \textcolor{green}{0.727} \\
\hline
DeamNet \cite{Ren2021} & \textcolor{red}{0.896} & \textcolor{red}{0.837} & \textcolor{blue}{0.737} \\
\hline
DRUNet \cite{ZhangL2021} & \textcolor{blue}{0.895} & \textcolor{red}{0.837} & \textcolor{red}{0.738} \\
\hline
DCBDNet & 0.889 & 0.829 & \textcolor{green}{0.727} \\
\hline
\end{tabular}
\end{table}

\begin{table}[htbp]
\centering
\caption{The averaged FSIM results of the compared methods on Set12 and BSD68 dataset with different noise levels.}
\label{tab:BSD68_FSIM}
\begin{tabular}{ccccc}
\cline{1-5}
Datasets & Methods & $\sigma$=15 & $\sigma$=25 & $\sigma$=50 \\
\cline{1-5}
\multirow{6}*{Set12} & DnCNN-S \cite{Zhang2017} & 0.761 & 0.719 & 0.655\\
\cline{2-5}
    & DnCNN-B \cite{Zhang2017} & 0.759 & 0.720 & 0.656\\
\cline{2-5}
    & IRCNN  \cite{ZhangZ2017} & 0.759 & 0.718 & 0.656\\
\cline{2-5}
    & FFDNet \cite{Zhang2018} & 0.760 & 0.719 & 0.656\\
\cline{2-5}
    & DRUNet \cite{ZhangL2021} & 0.768 & 0.727 & 0.666\\
\cline{2-5}
    & DCBDNet & 0.760 & 0.722 & 0.662\\
\hline
\multirow{6}*{BSD68} & DnCNN-S \cite{Zhang2017} & 0.746 & 0.689 & 0.602\\
\cline{2-5}
    & DnCNN-B \cite{Zhang2017} & 0.744 & 0.688 & 0.601\\
\cline{2-5}
    & IRCNN  \cite{ZhangZ2017} & 0.742 & 0.686 & 0.602\\
\cline{2-5}
    & FFDNet \cite{Zhang2018} & 0.745 & 0.690 & 0.602\\
\cline{2-5}
    & DRUNet \cite{ZhangL2021} & 0.750 & 0.696 & 0.612\\
\cline{2-5}
    & DCBDNet & 0.746 & 0.692 & 0.609\\
\hline
\end{tabular}
\end{table}

\begin{figure*}[htbp]
	\centering
	\begin{subfigure}{0.194\linewidth}
		\centering
		\includegraphics[width=0.98\linewidth]{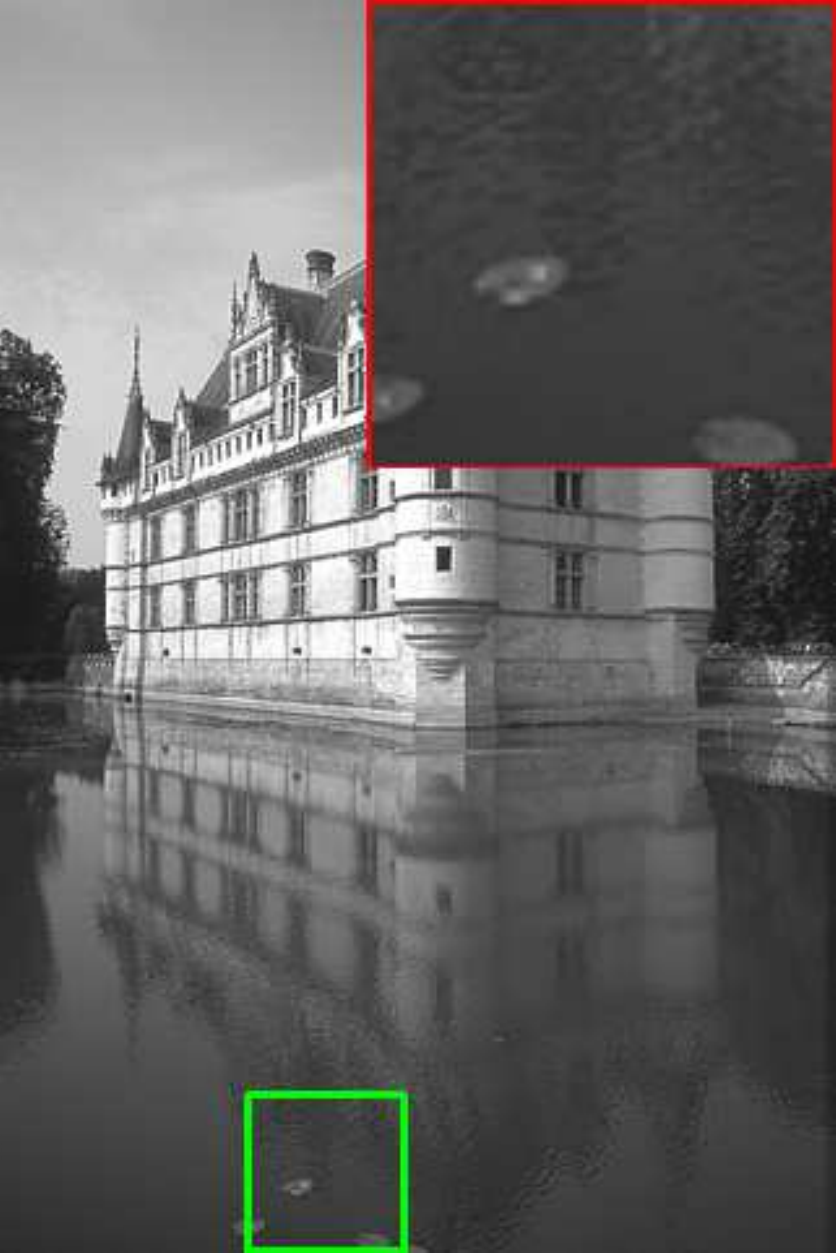}
		\caption{\tiny Ground truth}
	\end{subfigure}
    \centering
	\begin{subfigure}{0.194\linewidth}
		\centering
		\includegraphics[width=0.98\linewidth]{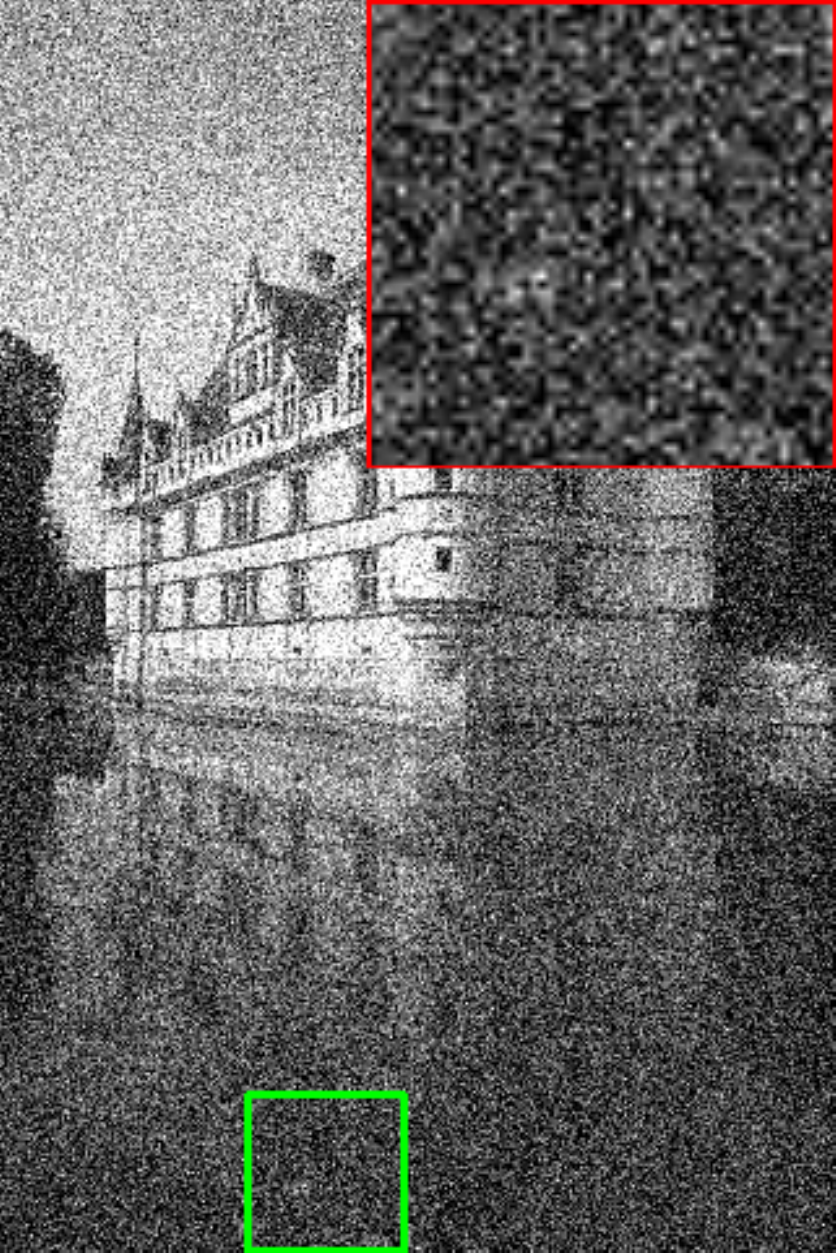}
		\caption{\tiny Noisy/14.15dB}
	\end{subfigure}
    \centering
	\begin{subfigure}{0.194\linewidth}
		\centering
		\includegraphics[width=0.98\linewidth]{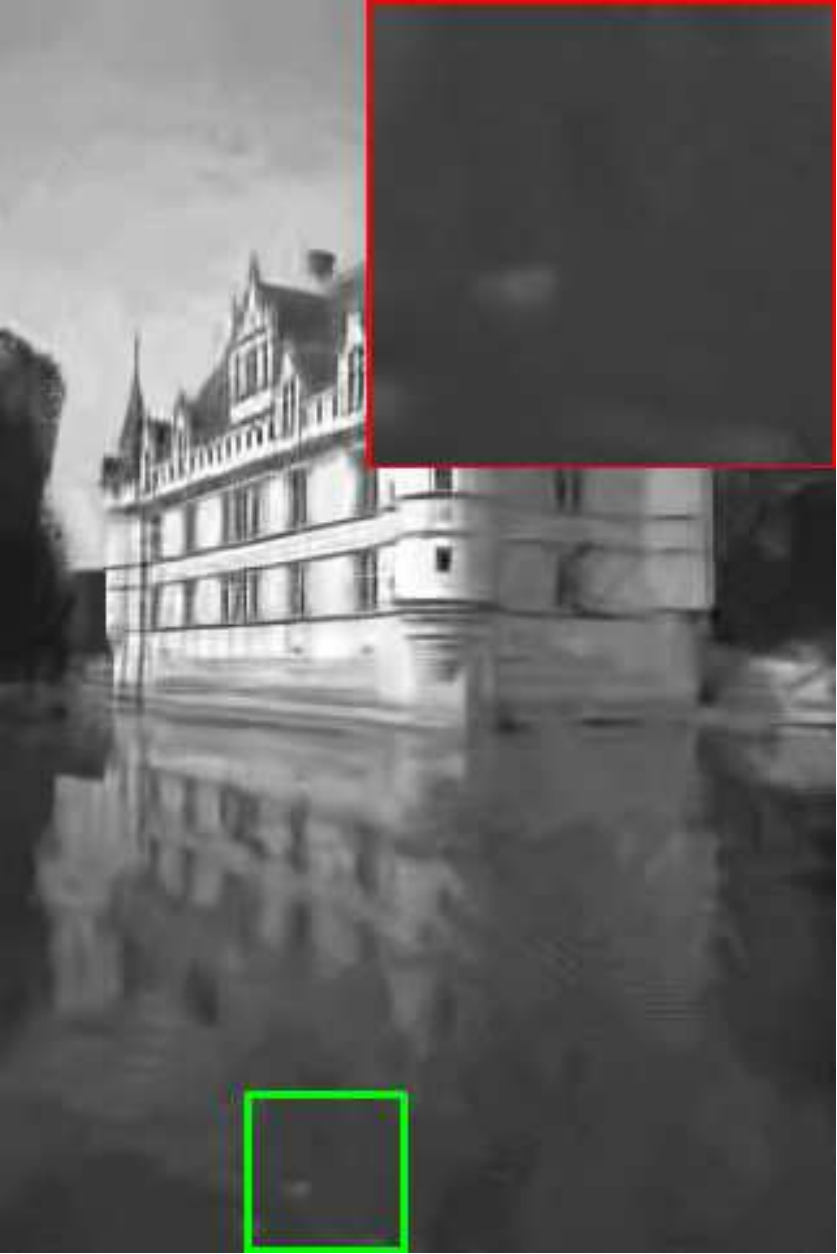}
		\caption{\tiny BM3D \cite{Dabov2007} /26.21dB}
	\end{subfigure}
    \centering
	\begin{subfigure}{0.194\linewidth}
		\centering
		\includegraphics[width=0.98\linewidth]{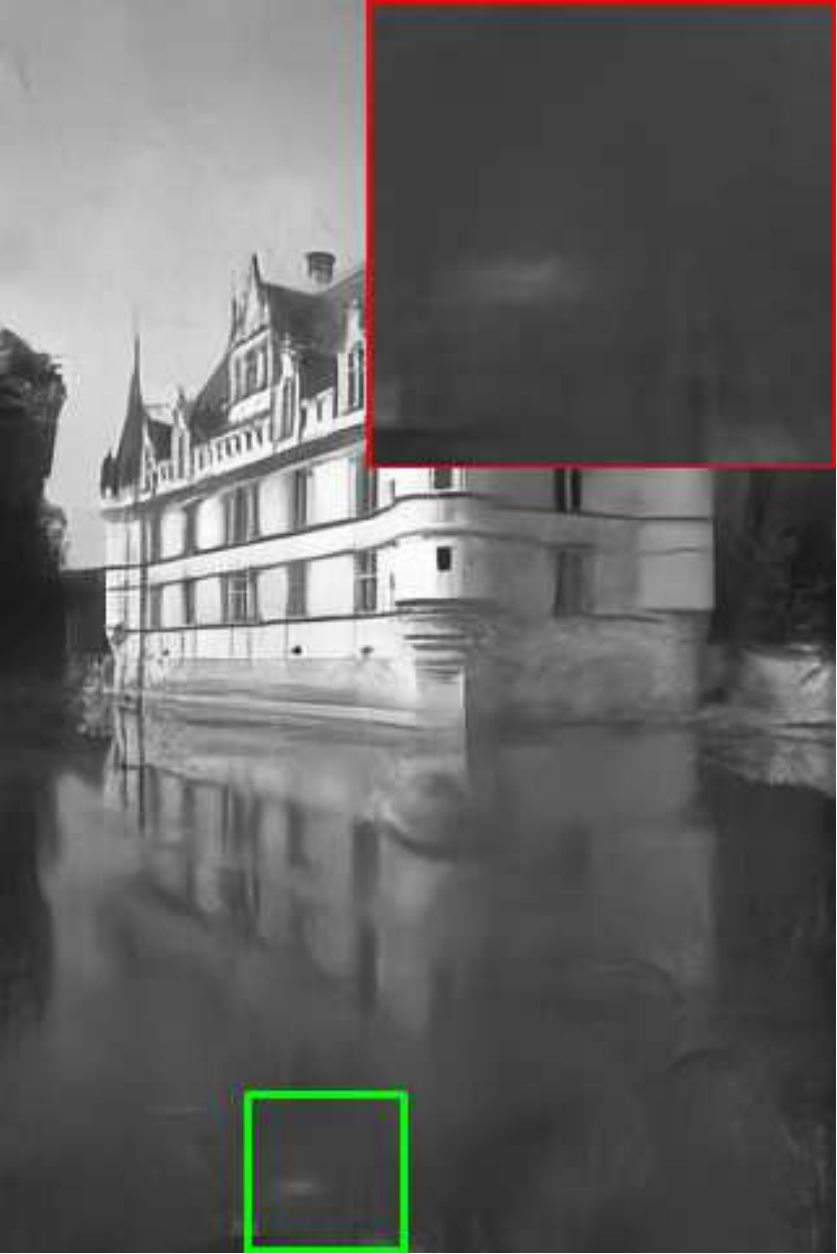}
		\caption{\tiny DnCNN-B \cite{Zhang2017} /26.89dB}
	\end{subfigure}
    \centering
	\begin{subfigure}{0.194\linewidth}
		\centering
		\includegraphics[width=0.98\linewidth]{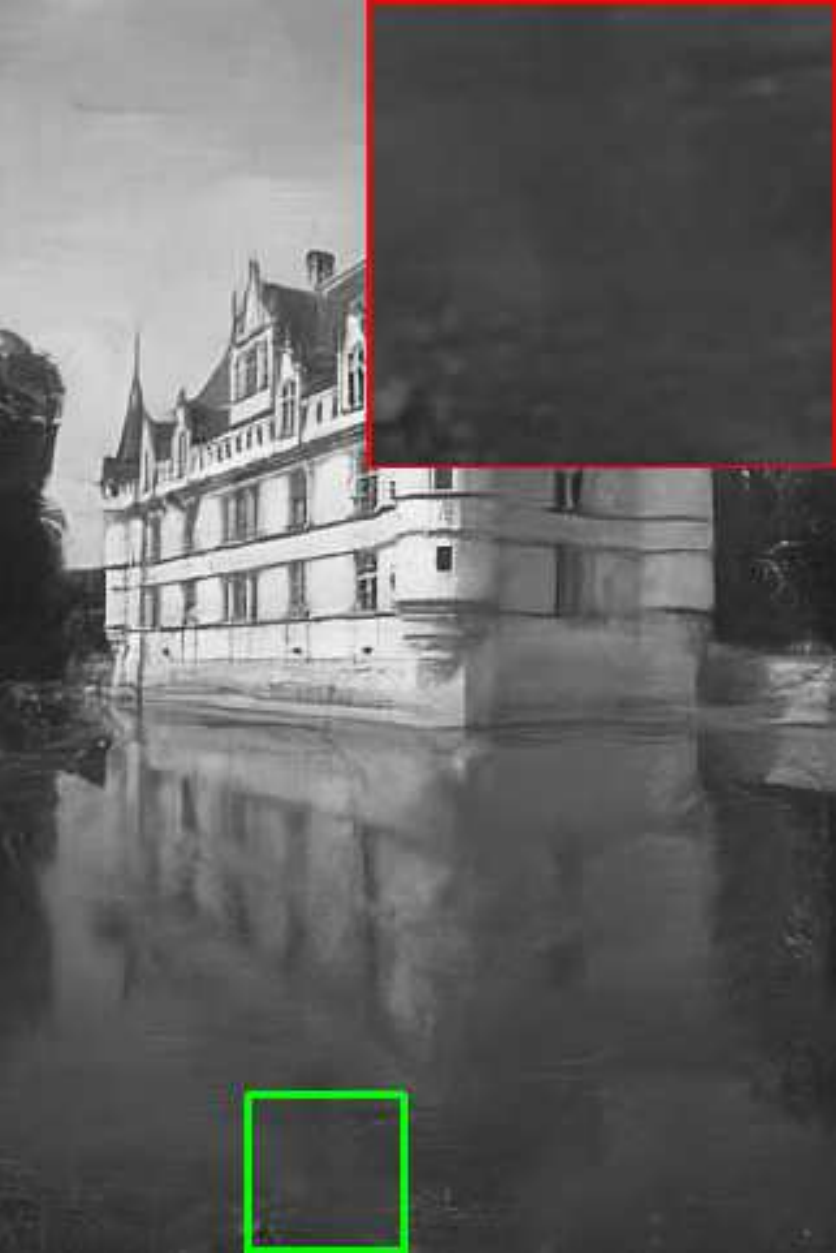}
		\caption{\tiny BUIFD \cite{Helou2020} /26.32dB}
	\end{subfigure}
    \centering
	\begin{subfigure}{0.194\linewidth}
		\centering
		\includegraphics[width=0.98\linewidth]{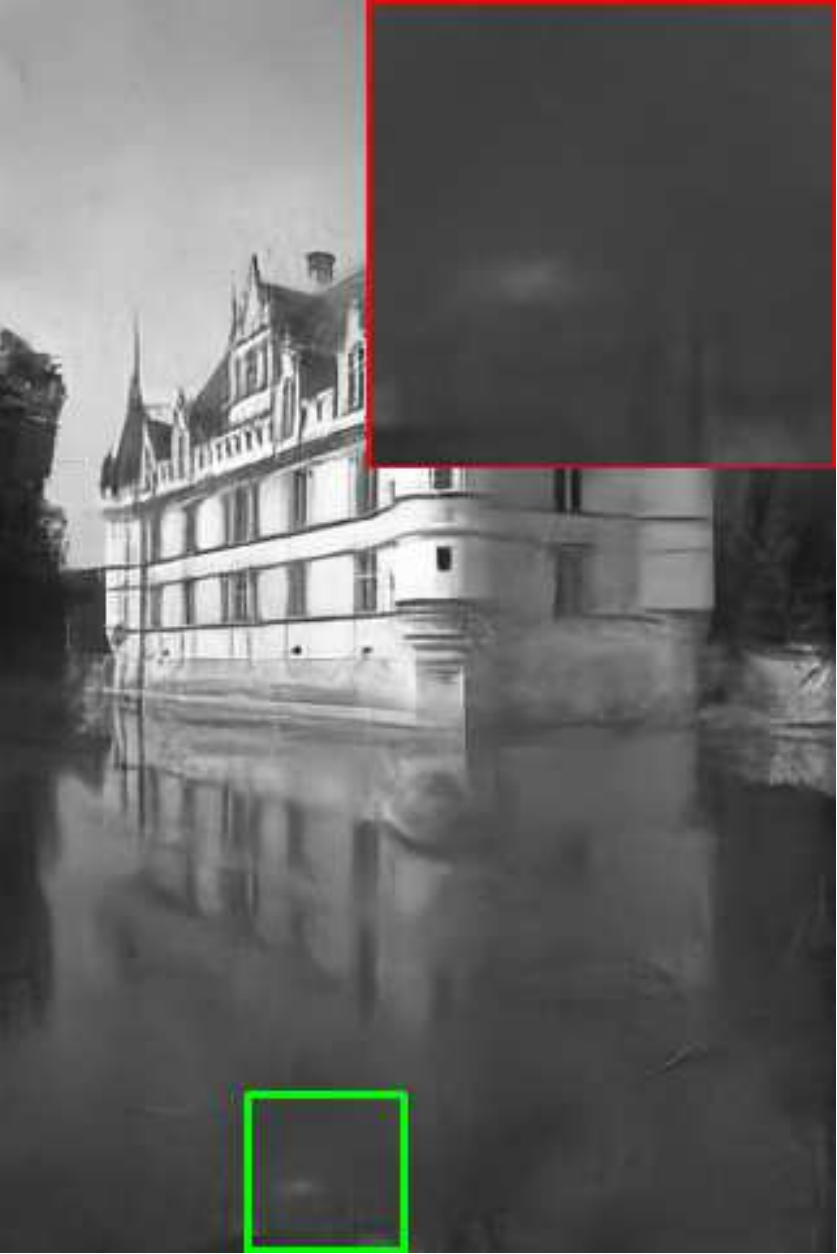}
		\caption{\tiny IRCNN \cite{ZhangZ2017} /26.85dB}
	\end{subfigure}
    \centering
	\begin{subfigure}{0.194\linewidth}
		\centering
		\includegraphics[width=0.98\linewidth]{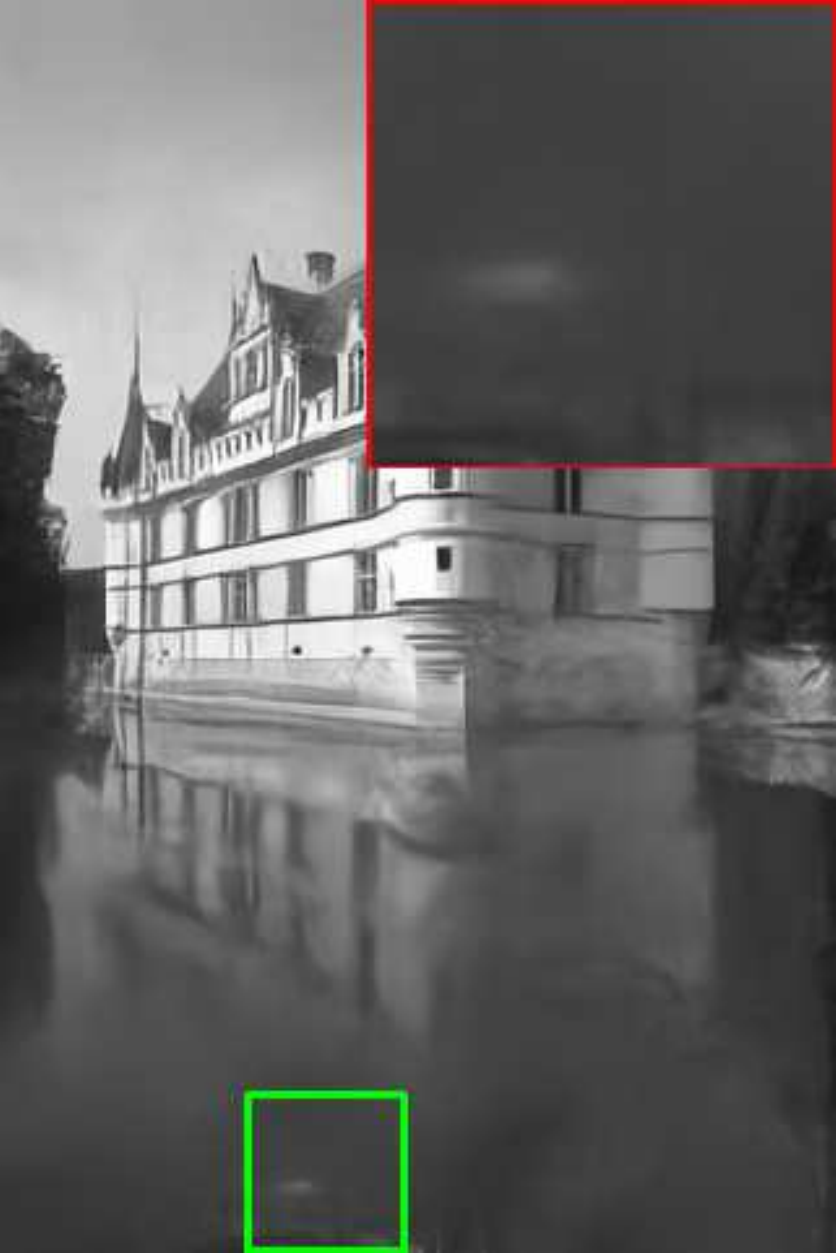}
		\caption{\tiny FFDNet \cite{Zhang2018} /26.92dB}
	\end{subfigure}
    \centering
	\begin{subfigure}{0.194\linewidth}
		\centering
		\includegraphics[width=0.98\linewidth]{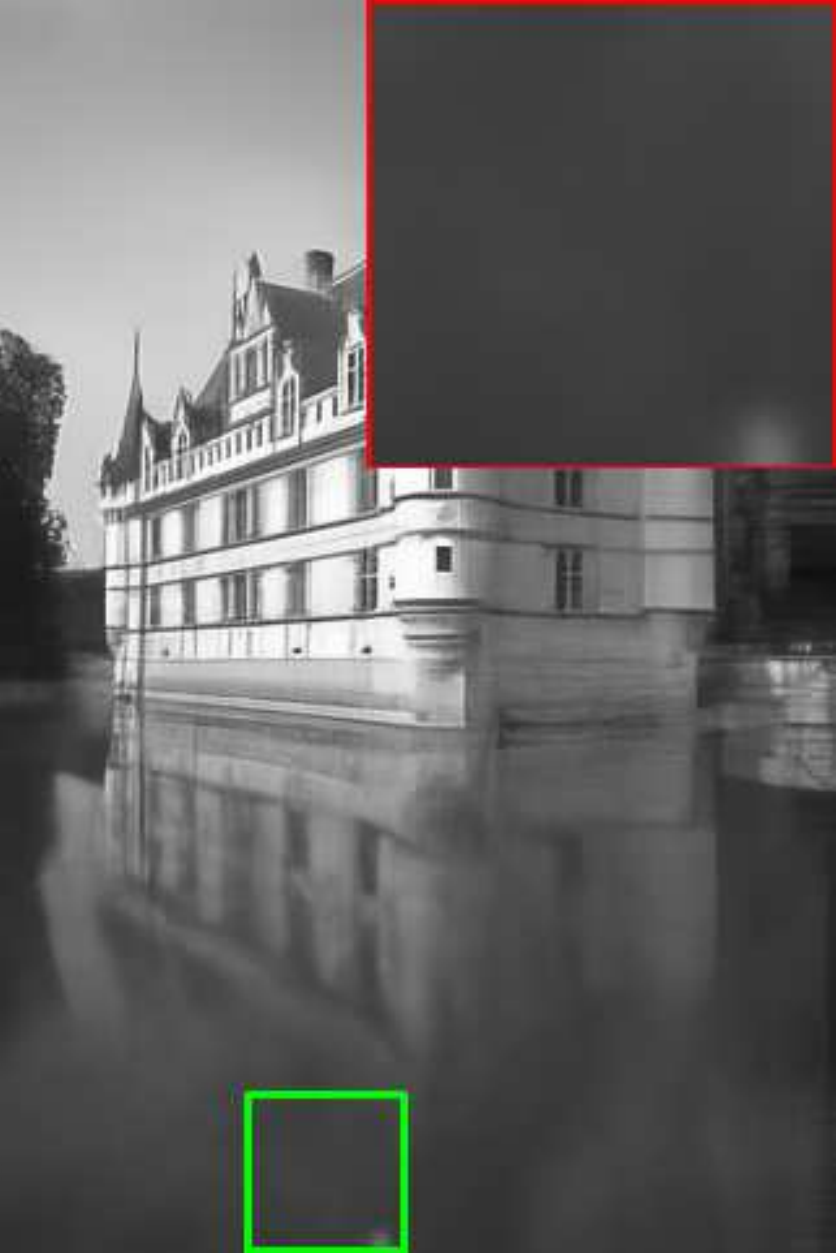}
		\caption{\tiny DeamNet \cite{Ren2021} /27.41dB}
	\end{subfigure}
    \centering
	\begin{subfigure}{0.194\linewidth}
		\centering
		\includegraphics[width=0.98\linewidth]{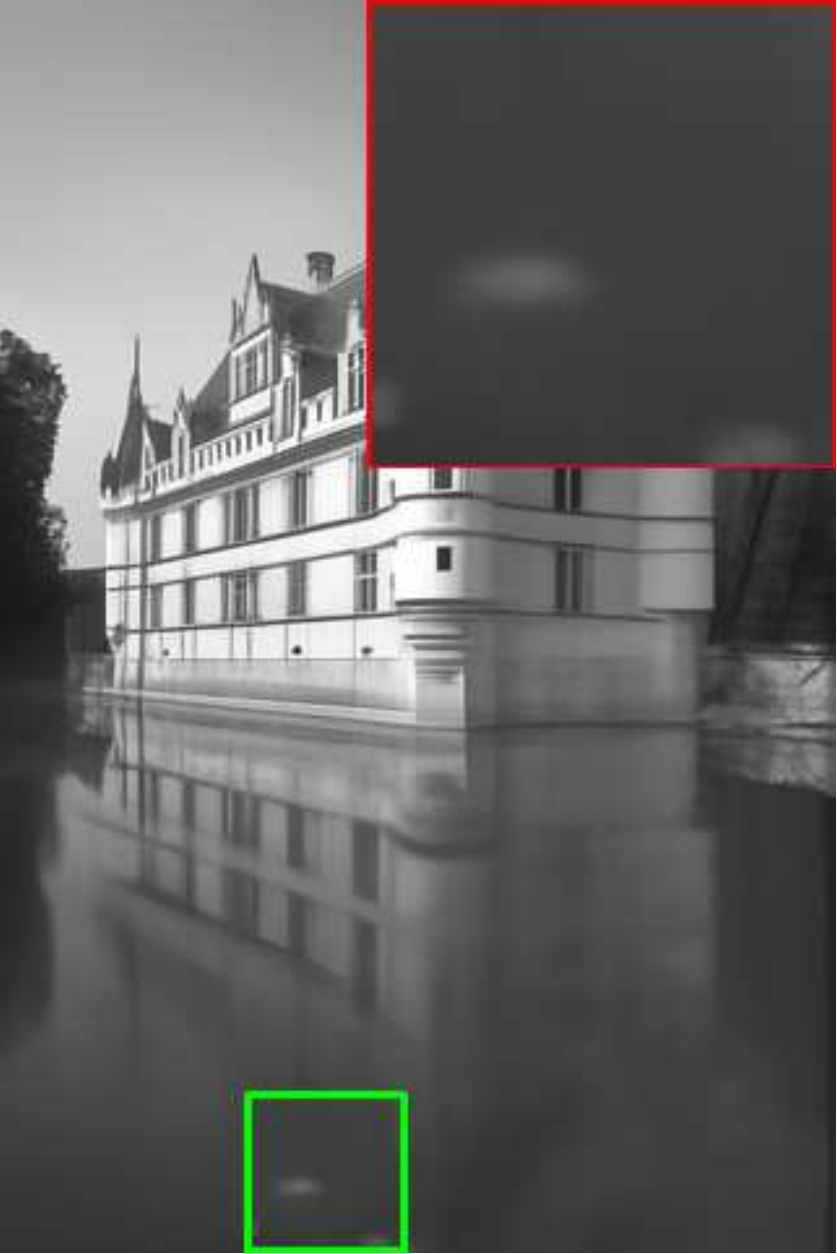}
		\caption{\tiny DRUNet \cite{ZhangL2021} /27.47dB}
	\end{subfigure}
    \centering
	\begin{subfigure}{0.194\linewidth}
		\centering
		\includegraphics[width=0.98\linewidth]{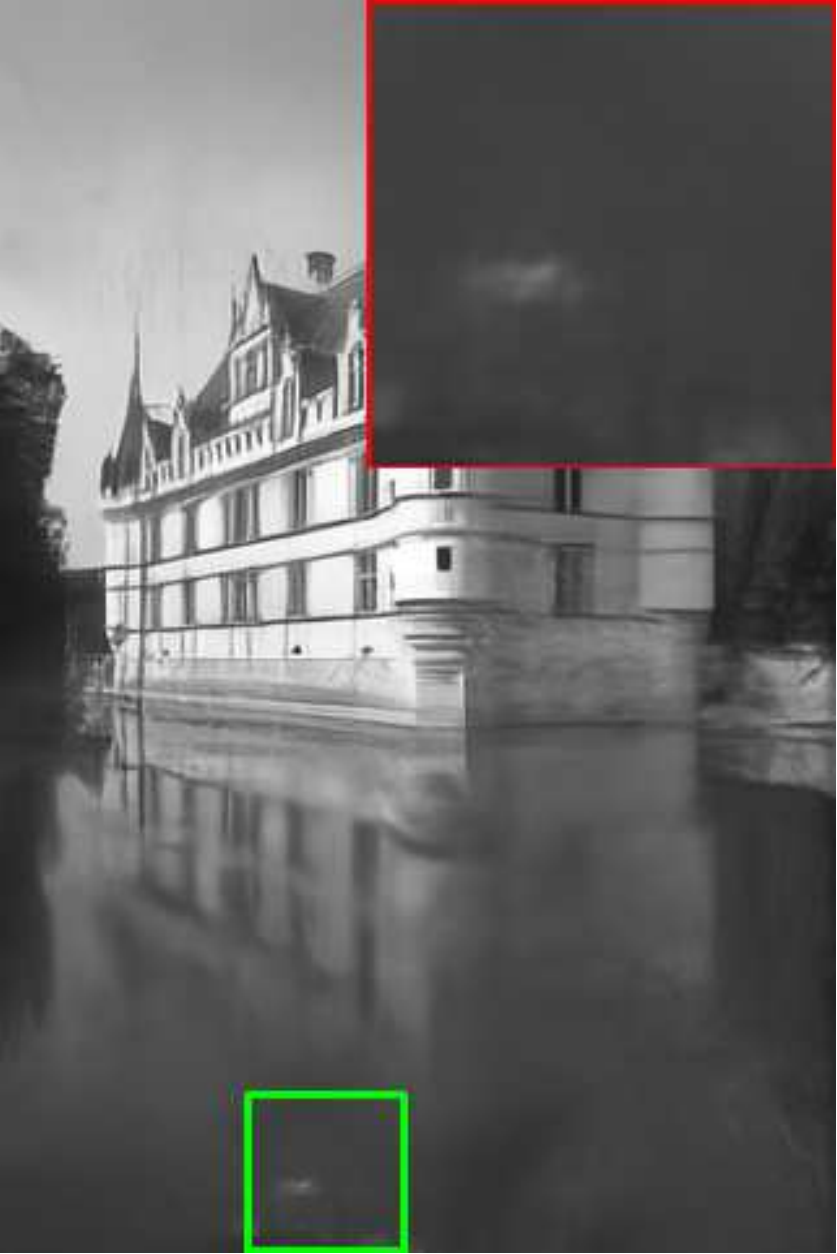}
		\caption{\tiny DCBDNet/27.05dB}
	\end{subfigure}
\caption{Visual results on the image ``Castle'' of different denoising methods.}
\label{fig:Castle}
\end{figure*}

Fig. \ref{fig:Castle} shows the denoised results on the ``Castle'' image from the BSD68 dataset at the noise level 50 of different methods. In Fig. \ref{fig:Castle}, we zoom in a region (green box) for detail comparison (red box). It can be seen that though the DeamNet and DRUNet achieve high PSNRs, the `over-smooth' effects are also brought. The DeamNet completely replaces the white small region (not noise) with the image context (Fig. \ref{fig:Castle} (h)), and the DRUNet changes the original irregular white region into a regular one (Fig. \ref{fig:Castle} (i)), while our proposed method can maintain a balance between noise removal and detail preservation (Fig. \ref{fig:Castle} (j)).

To evaluate the denoising performance of AWGN removal on color images, the CBSD68, Kodak24, and McMaster datasets were used. The PSNR values obtained by the compared methods are listed in Table \ref{tab:CBSD68_Kodak24_McMaster}. It can be seen that our proposed DCBDNet outperforms all the compared methods except the DRUNet on CBSD68, Kodak24, and McMaster datasets at noise levels 50 and 75. The proposed DCBDNet outperforms the BRDNet on the three datasets when the noise level is greater than 35. As the BRDNet also employs a dual CNN structure, we think the reason is a deeper network structure like the one used in the BRDNet may have more learning ability when the noise power is not strong.

\begin{table}[htbp]
\centering
\caption{The averaged PSNR (dB) of the compared methods on the CBSD68, Kodak24 and McMaster datasets. The best, second best and third best is marked in red, blue and green respectively.}
\label{tab:CBSD68_Kodak24_McMaster}
\begin{tabular}{|c|c|c|c|c|c|c|c|c|c|c|c|c|}
\hline
Datasets &  Methods & $\sigma$=15 & $\sigma$=25 & $\sigma$=35 & $\sigma$=50 & $\sigma$=75\\
\hline
\multirow{17}*{CBSD68} & CBM3D \cite{Dabov2007} & 33.52 & 30.71 & 28.89 & 27.38 & 25.74\\
\cline{3-7}
    & MCWNNM \cite{Xu2017} &  30.91 & 27.61 & 25.88 & 23.18 & 21.21\\
\cline{3-7}
    & CDnCNN-S \cite{Zhang2017} & 33.89 & 31.23 & 29.58 & 27.92 & 24.47\\
\cline{3-7}
    & BUIFD \cite{Helou2020} & 33.65 & 30.76 & 28.82 & 26.61 & 23.64\\
\cline{3-7}
    & IRCNN \cite{ZhangZ2017} & 33.86 & 31.16 & 29.50 & 27.86 & -\\
\cline{3-7}
    & FFDNet \cite{Zhang2018} & 33.87 & 31.21 & 29.58 & 27.96 & 26.24\\
\cline{3-7}
    & DSNetB \cite{Peng2019} & 33.91 & 31.28 & - & 28.05 & -\\
\cline{3-7}
    & RIDNet \cite{Anwar2019} & 34.01 & 31.37 & - & 28.14 & -\\
\cline{3-7}
    & VDN \cite{YueYZM2019} & 33.90 & 31.35 & - & 28.19 & -\\
\cline{3-7}
    & BRDNet \cite{Tian2020} & \textcolor{blue}{34.10} & \textcolor{blue}{31.43} & 29.77 & \textcolor{green}{28.16} & \textcolor{green}{26.43}\\
\cline{3-7}
    & ADNet  \cite{TianX2020} & 33.99 & 31.31 & 29.66 & 28.04 & 26.33\\
\cline{3-7}
    & DudeNet \cite{Tian2021} & 34.01 & 31.34 & 29.71 & 28.09 & 26.40\\
\cline{3-7}
    & GradNet \cite{LiuAZT2020} & \textcolor{green}{34.07} & 31.39 & - & 28.12 & -\\
\cline{3-7}
    & CDNet \cite{Quan2021} & - & 31.34 & \textcolor{blue}{29.84} & 28.14 & -\\
\cline{3-7}
    & AirNet \cite{Li2022} & 33.92 & 31.26 & - & 28.01 & -\\
\cline{3-7}
    & DRUNet \cite{ZhangL2021} & \textcolor{red}{34.30} & \textcolor{red}{31.69} & \textcolor{red}{30.09} & \textcolor{red}{28.51} & \textcolor{red}{26.84}\\
\cline{3-7}
    & DCBDNet & 34.01 & \textcolor{green}{31.41} & \textcolor{green}{29.81} & \textcolor{blue}{28.22} & \textcolor{blue}{26.54}\\
\hline
\multirow{14}*{Kodak24} & CBM3D \cite{Dabov2007} & 34.28 & 31.68 & 29.90 & 28.46 & 26.82\\
\cline{3-7}
    & MCWNNM \cite{Xu2017} & 32.00 & 28.76 & 27.02 & 21.18 & 18.06\\
\cline{3-7}
    & CDnCNN-S \cite{Zhang2017} & 34.48 & 32.03 & 30.46 & 28.85 & 25.04\\
\cline{3-7}
    & BUIFD \cite{Helou2020} & 34.41 & 31.77 & 29.94 & 27.74 & 24.67\\
\cline{3-7}
    & IRCNN  \cite{ZhangZ2017} & 34.56 & 32.03 & 30.43 & 28.81 & -\\
\cline{3-7}
    & FFDNet \cite{Zhang2018} & 34.63 & 32.13 & 30.57 & 28.98 & 27.27\\
\cline{3-7}
    & DSNetB \cite{Peng2019} & 34.63 & 32.16 & - & 29.05 & -\\
\cline{3-7}
    & BRDNet \cite{Tian2020} & \textcolor{blue}{34.88} & \textcolor{blue}{32.41} & \textcolor{green}{30.80} & 29.22 & \textcolor{green}{27.49}\\
\cline{3-7}
    & ADNet  \cite{TianX2020} & 34.76 & 32.26 & 30.68 & 29.10 & 27.40\\
\cline{3-7}
    & DudeNet \cite{Tian2021} & 34.81 & 32.26 & 30.69 & 29.10 & 27.39\\
\cline{3-7}
    & GradNet \cite{LiuAZT2020} & \textcolor{green}{34.85} & 32.35 & - & \textcolor{green}{29.23} & -\\
\cline{3-7}
    & AirNet \cite{Li2022} & 34.68 & 32.21 & - & 29.06 & -\\
\cline{3-7}
    & DRUNet \cite{ZhangL2021} & \textcolor{red}{35.31} & \textcolor{red}{32.89} & \textcolor{red}{31.26} & \textcolor{red}{29.86} & \textcolor{red}{28.06}\\
\cline{3-7}
    & DCBDNet & 34.80 & \textcolor{green}{32.38} & \textcolor{blue}{30.84} & \textcolor{blue}{29.29} & \textcolor{blue}{27.60}\\
\hline
\multirow{13}*{McMaster} & CBM3D \cite{Dabov2007} & 34.06 & 31.66 & 29.92 & 28.51 & 26.79\\
\cline{3-7}
    & MCWNNM \cite{Xu2017} & 32.75 & 29.39 & 27.44 & 21.37 & 18.16\\
\cline{3-7}
    & CDnCNN-S \cite{Zhang2017} & 33.44 & 31.51 & 30.14 & 28.61 & 25.10\\
\cline{3-7}
    & BUIFD \cite{Helou2020} & 33.84 & 31.06 & 28.87 & 26.20 & 22.75\\
\cline{3-7}
    & IRCNN \cite{ZhangZ2017} & 34.58 & 32.18 & 30.59 & 28.91 & -	\\
\cline{3-7}
    & FFDNet \cite{Zhang2018} & 34.66 & 32.35 & 30.81 & 29.18 & 27.33\\
\cline{3-7}
    & DSNetB \cite{Peng2019} & 34.67 & 32.40 & - & 29.28 & -\\
\cline{3-7}
    & BRDNet \cite{Tian2020} & \textcolor{blue}{35.08} & \textcolor{blue}{32.75} & \textcolor{blue}{31.15} & \textcolor{green}{29.52} & \textcolor{green}{27.72}\\
\cline{3-7}
    & ADNet \cite{TianX2020} & \textcolor{green}{34.93} & \textcolor{green}{32.56} & 31.00 & 29.36 & 27.53\\
\cline{3-7}
    & GradNet \cite{LiuAZT2020} & 34.81 & 32.45 & - & 29.39 & -\\
\cline{3-7}
    & AirNet \cite{Li2022} & 34.70 & 32.44 & - & 29.26 & -\\
\cline{3-7}
    & DRUNet \cite{ZhangL2021} & \textcolor{red}{35.40} & \textcolor{red}{33.14} & \textcolor{red}{31.66} & \textcolor{red}{30.08} & \textcolor{red}{28.29}\\
\cline{3-7}
    & DCBDNet & 34.76 & \textcolor{green}{32.56} & \textcolor{green}{31.10} & \textcolor{blue}{29.54} & \textcolor{blue}{27.75}\\
\hline
\end{tabular}
\end{table}

Table \ref{tab: Kodak24_Metrics} shows the averaged SSIM, FSIM, LPIPS, and the Inception-Score (IS) results of different methods on CBSD68, Kodak24, and McMaster datasets. Our DCBDNet model outperforms DnCNN, IRCNN, and FFDNet on the four evaluation metrics.

\begin{table*}[htbp]\scriptsize
\centering
\caption{The averaged SSIM, FSIM, LPIPS and Inception-Score (IS) results of the compared methods on CBSD68, Kodak24 and McMaster datasets with different noise levels.}
\label{tab: Kodak24_Metrics}
\begin{tabular}{c|c|cccc|cccc|cccc}
\toprule[1pt]
\multirow{2}*{Datasets} & \multirow{2}*{Methods} & \multicolumn{4}{c|}{$\sigma$=15} & \multicolumn{4}{c|}{$\sigma$=25} & \multicolumn{4}{c}{$\sigma$=50}\\
\cline{3-14}
     &  & SSIM & FSIM & LPIPS & IS & SSIM & FSIM & LPIPS & IS & SSIM & FSIM & LPIPS & IS\\
\midrule[1pt]
\multirow{5}*{CBSD68} & CDnCNN-B \cite{Zhang2017} &  0.929 & 0.784 & 0.0628 & 4.354 & 0.883 & 0.737 & 0.1090 & 4.250 & 0.790 & 0.658 & 0.2101 & 4.064\\
\cline{2-14}
& IRCNN  \cite{ZhangZ2017} &  0.929 & 0.782 & 0.0626 & 4.330 & 0.882 & 0.736 & 0.1078 & 4.243 & 0.790 & 0.660 & 0.2039 & 4.189\\
\cline{2-14}
& FFDNet \cite{Zhang2018} &  0.929 & 0.782 & 0.0655 & 4.270 & 0.882 & 0.733 & 0.1210 & 4.166 & 0.789 & 0.648 & 0.2442 & 3.946\\
\cline{2-14}
& DRUNet \cite{ZhangL2021} &  0.934 & 0.791 & 0.0556 & 4.314 & 0.893 & 0.748 & 0.0962 & 4.259 & 0.810 & 0.675 & 0.1815 & 4.097\\
\cline{2-14}
& DCBDNet & 0.931 & 0.785 & 0.0604 & 4.330 & 0.887 & 0.740 & 0.1036 & 4.273 & 0.780 & 0.665 & 0.1994 & 4.054\\
\midrule[1pt]
\multirow{5}*{Kodak24} & CDnCNN-B \cite{Zhang2017} & 0.920 & 0.765 & 0.0828 & 2.003 & 0.876 & 0.713 & 0.1292 & 1.965 & 0.791 & 0.632 & 0.2290 & 1.917\\
\cline{2-14}
& IRCNN  \cite{ZhangZ2017} & 0.920 & 0.764 & 0.0810 & 1.989 & 0.877 & 0.712 & 0.1270 & 1.977 & 0.793 & 0.634 & 0.2202 & 1.931\\
\cline{2-14}
& FFDNet \cite{Zhang2018} & 0.922 & 0.764 & 0.0846 & 1.988 & 0.878 & 0.709 & 0.1395 & 1.947 & 0.794 & 0.621 & 0.2553 & 1.933\\
\cline{2-14}
& DRUNet \cite{ZhangL2021} & 0.929 & 0.777 & 0.0688 & 2.021 & 0.891 & 0.730 & 0.1081 & 1.971 & 0.820 & 0.655 & 0.1862 & 1.874\\
\cline{2-14}
& DCBDNet & 0.924 & 0.770 & 0.0762 & 2.009 & 0.884 & 0.721 & 0.1182 & 1.961 & 0.806 & 0.644 & 0.2064 & 1.907\\
\midrule[1pt]
\multirow{5}*{McMaster} & CDnCNN-B \cite{Zhang2017} & 0.904 & 0.759 & 0.0684 & 1.488 & 0.869 & 0.723 & 0.1014 & 1.441 & 0.799 & 0.662 & 0.1725 & 1.457\\
\cline{2-14}
& IRCNN  \cite{ZhangZ2017} & 0.920 & 0.770 & 0.0608 & 1.485 & 0.882 & 0.730 & 0.0935 & 1.450 & 0.807 & 0.668 & 0.1567 & 1.458\\
\cline{2-14}
& FFDNet \cite{Zhang2018} & 0.922 & 0.769 & 0.0648 & 1.505 & 0.886 & 0.731 & 0.1026 & 1.465 & 0.815 & 0.668 & 0.1832 & 1.458\\
\cline{2-14}
& DRUNet \cite{ZhangL2021} &  0.932 & 0.781 & 0.0512 & 1.472 & 0.903 & 0.746 & 0.0812 & 1.458 & 0.846 & 0.689 & 0.1403 & 1.445\\
\cline{2-14}
& DCBDNet & 0.923 & 0.772 & 0.0599 & 1.487 & 0.891 & 0.737 & 0.0909 & 1.472 & 0.828 & 0.679 & 0.1558 & 1.435\\
\bottomrule[1pt]
\end{tabular}
\end{table*}

\begin{figure*}[htbp]
	\centering
	\begin{subfigure}{0.194\linewidth}
		\centering
		\includegraphics[width=0.98\linewidth]{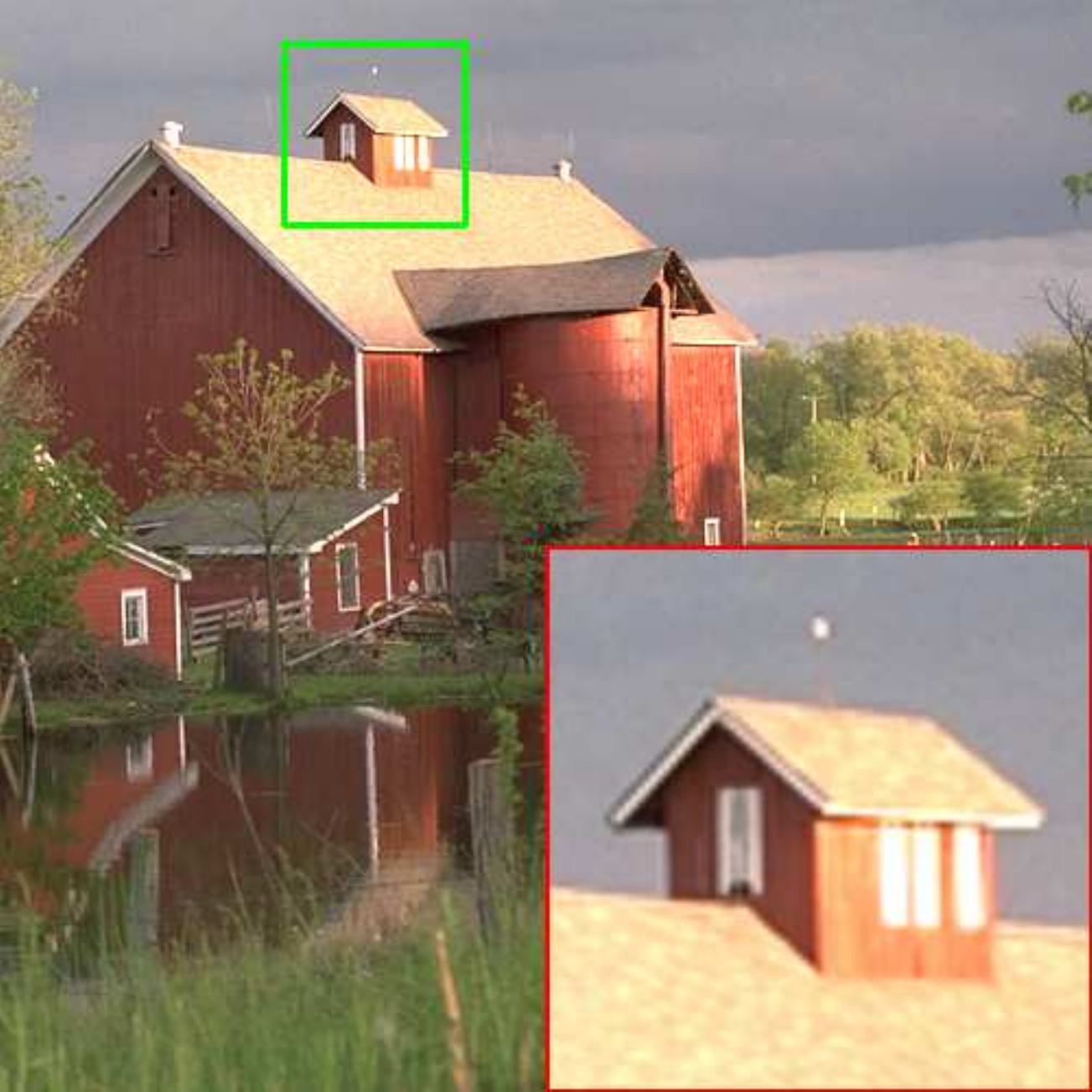}
		\caption{\tiny Ground truth}
	\end{subfigure}
    \centering
	\begin{subfigure}{0.194\linewidth}
		\centering
		\includegraphics[width=0.98\linewidth]{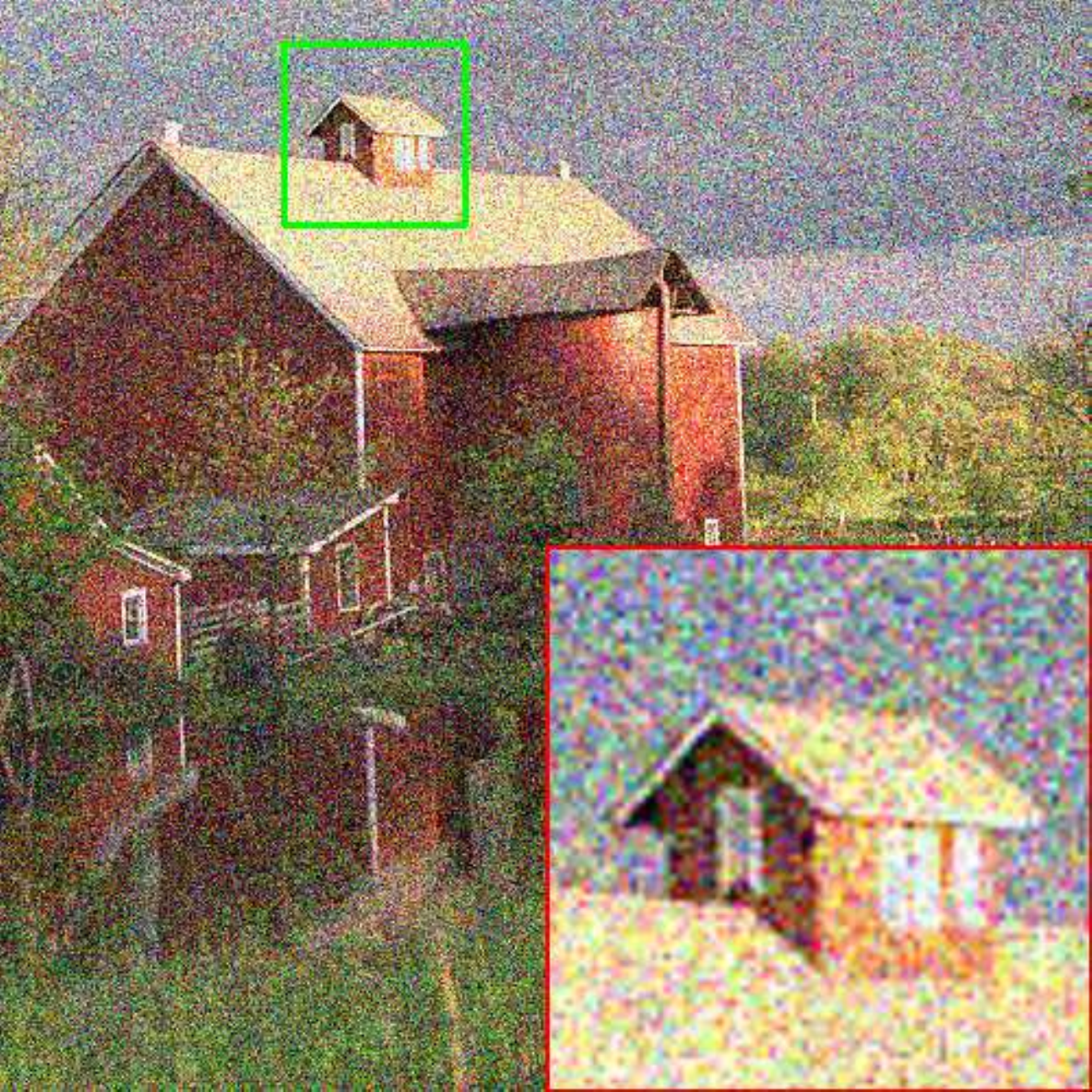}
		\caption{\tiny Noisy/14.16dB}
	\end{subfigure}
    \centering
	\begin{subfigure}{0.194\linewidth}
		\centering
		\includegraphics[width=0.98\linewidth]{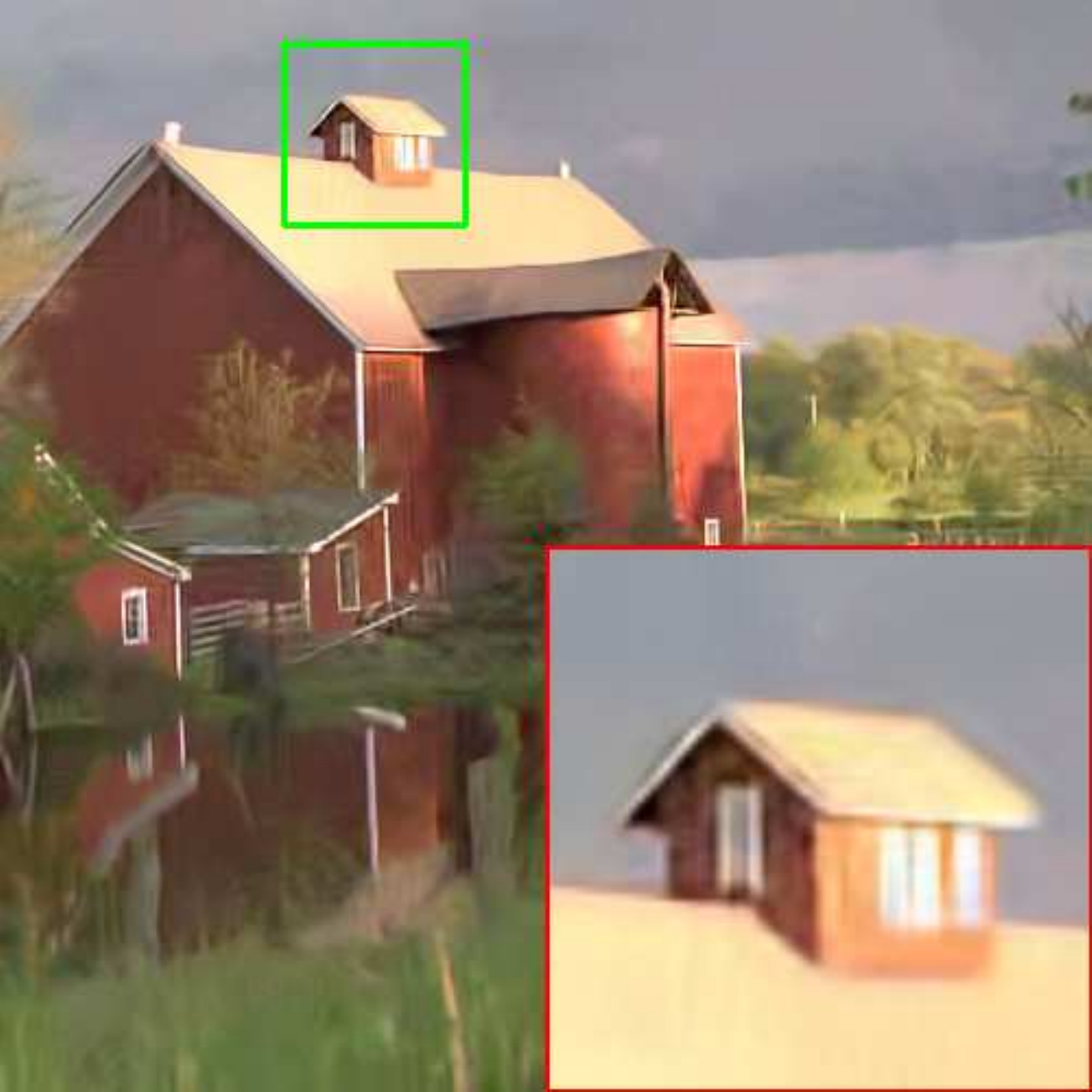}
		\caption{\tiny CBM3D \cite{Dabov2007} /28.34dB}
	\end{subfigure}
    \centering
	\begin{subfigure}{0.194\linewidth}
		\centering
		\includegraphics[width=0.98\linewidth]{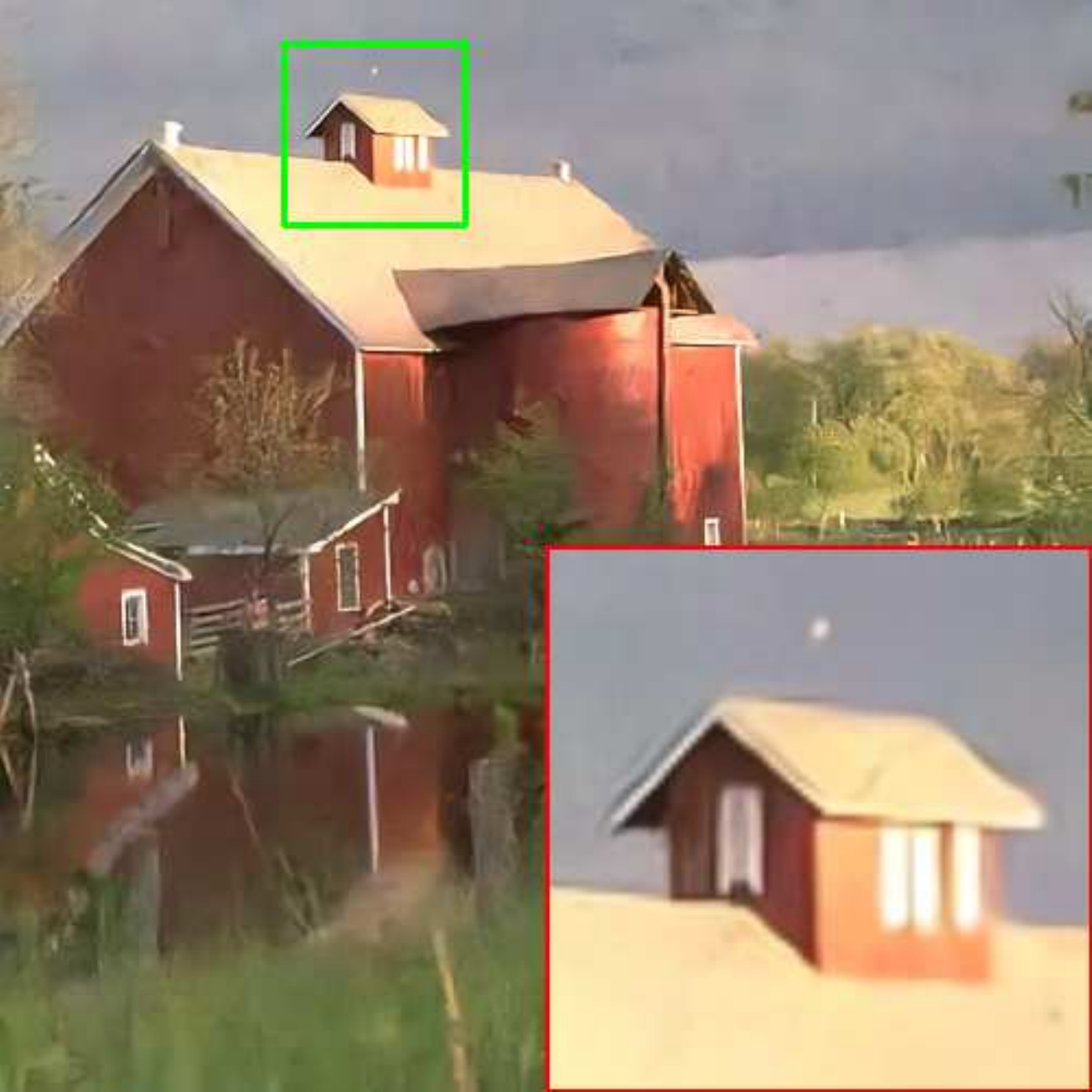}
		\caption{\tiny CDnCNN-B \cite{Zhang2017}/28.73dB}
	\end{subfigure}
    \centering
	\begin{subfigure}{0.194\linewidth}
		\centering
		\includegraphics[width=0.98\linewidth]{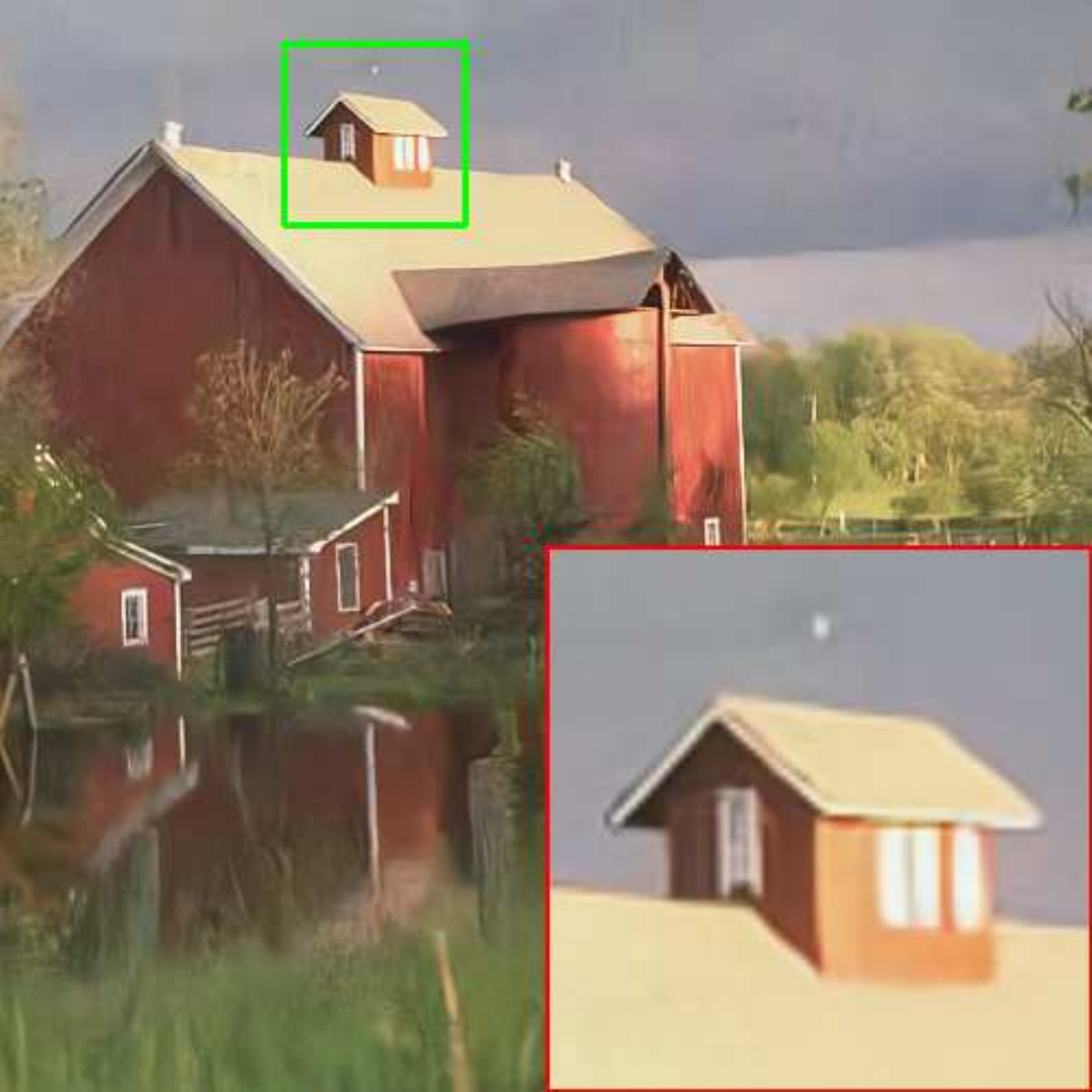}
		\caption{\tiny BUIFD \cite{Helou2020} /28.30dB}
	\end{subfigure}
    \centering
	\begin{subfigure}{0.194\linewidth}
		\centering
		\includegraphics[width=0.98\linewidth]{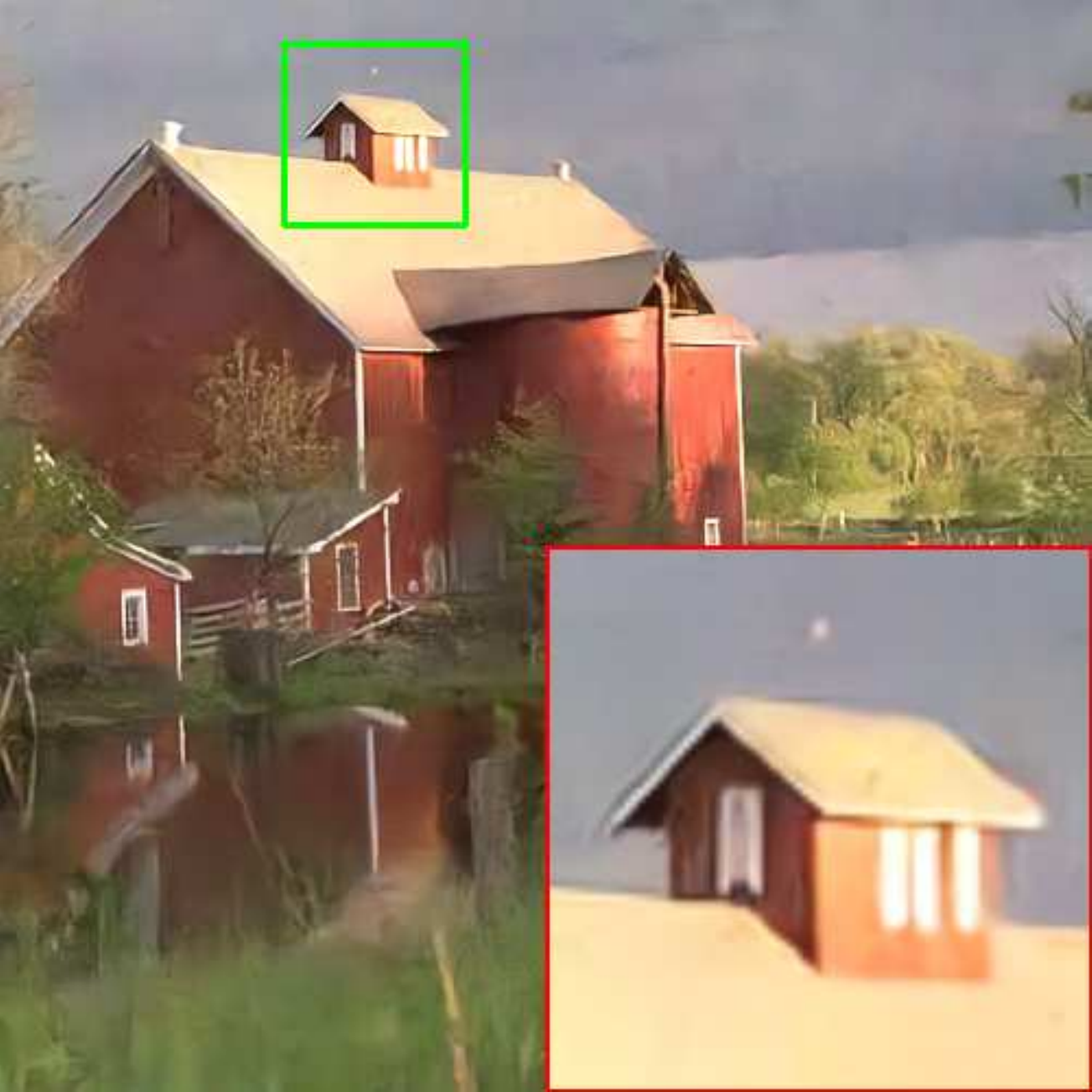}
		\caption{\tiny IRCNN \cite{ZhangZ2017} /28.71dB}
	\end{subfigure}
    \centering
	\begin{subfigure}{0.194\linewidth}
		\centering
		\includegraphics[width=0.98\linewidth]{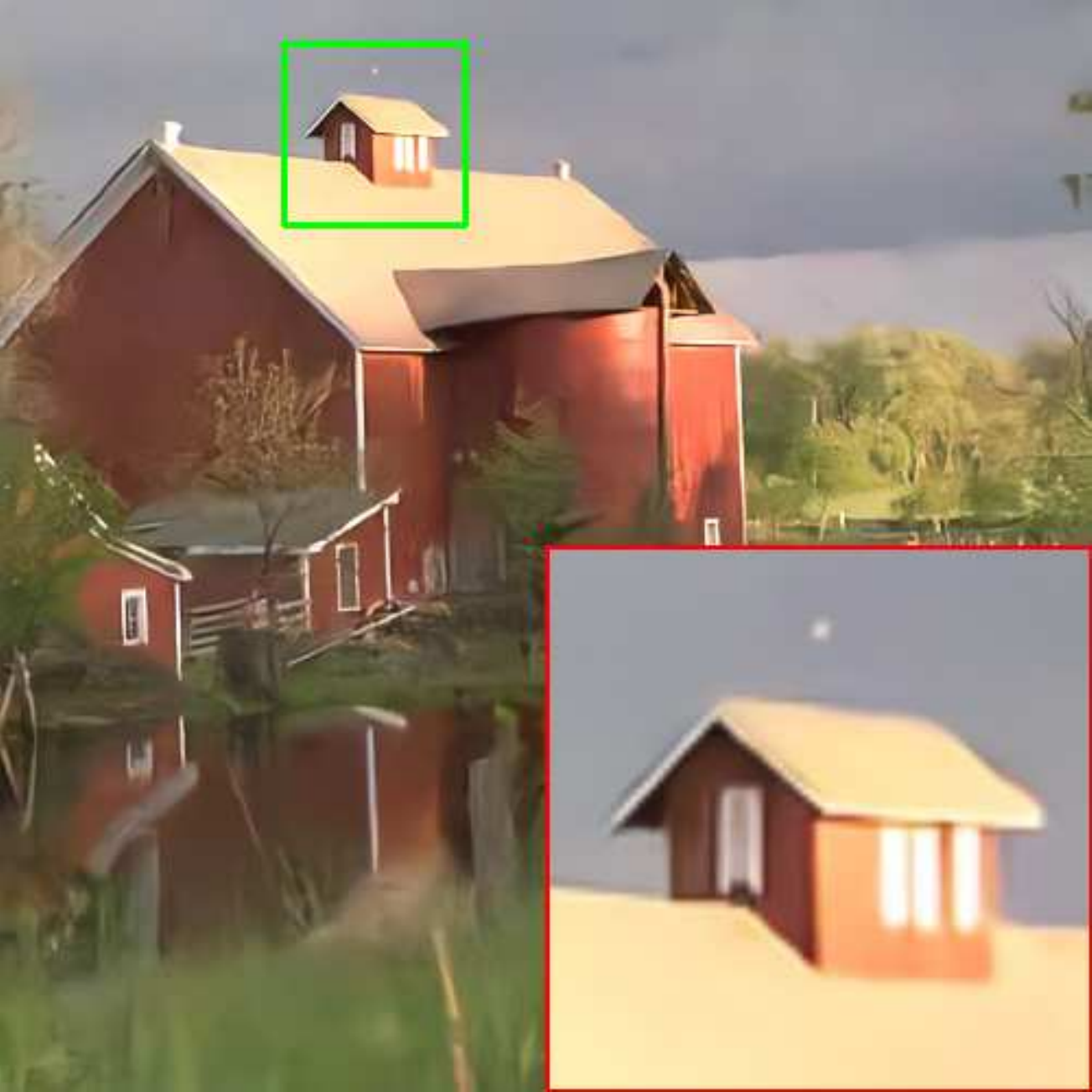}
		\caption{\tiny FFDNet \cite{Zhang2018} /28.83dB}
	\end{subfigure}
    \centering
	\begin{subfigure}{0.194\linewidth}
		\centering
		\includegraphics[width=0.98\linewidth]{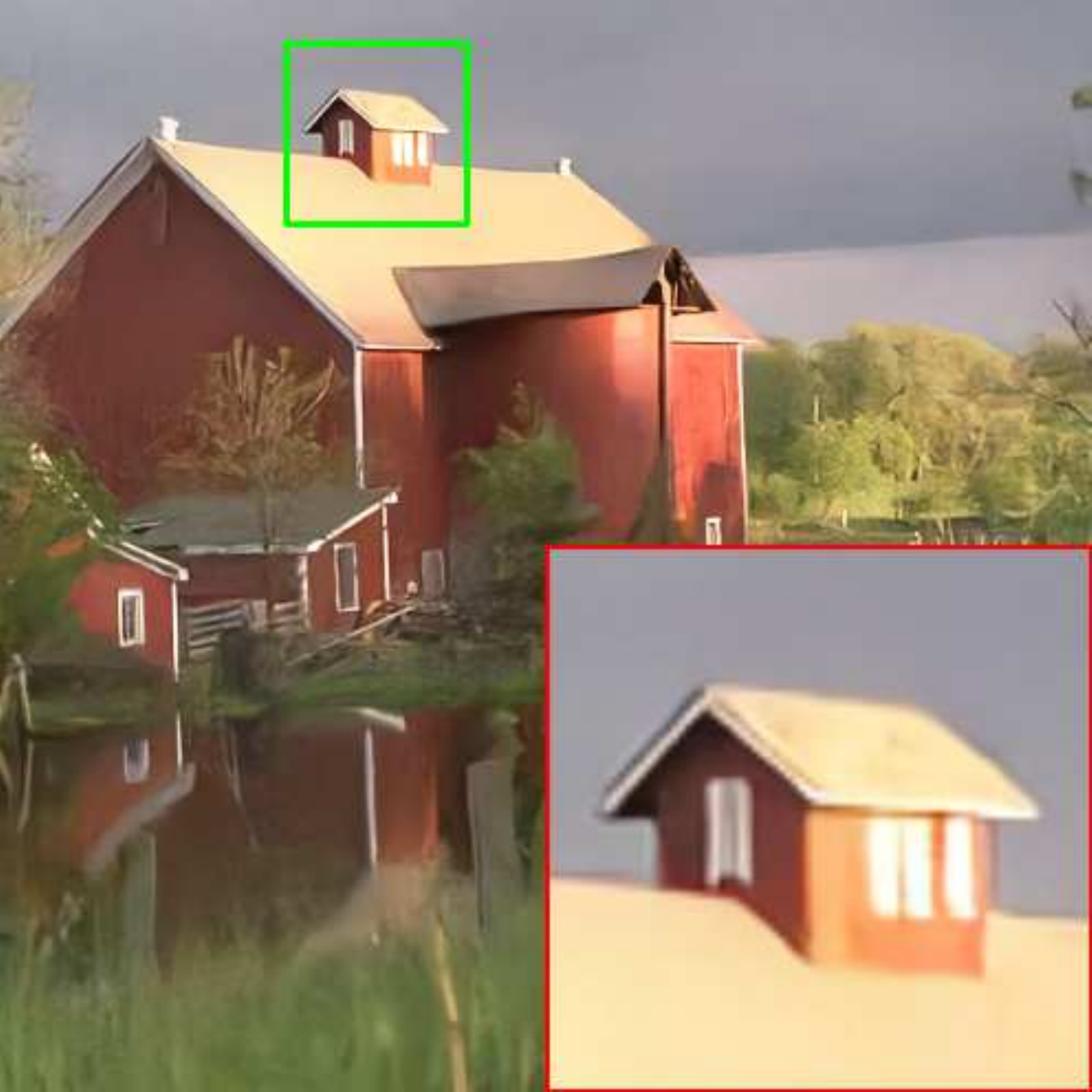}
		\caption{\tiny AirNet \cite{Li2022} /28.89dB}
	\end{subfigure}
    \centering
	\begin{subfigure}{0.194\linewidth}
		\centering
		\includegraphics[width=0.98\linewidth]{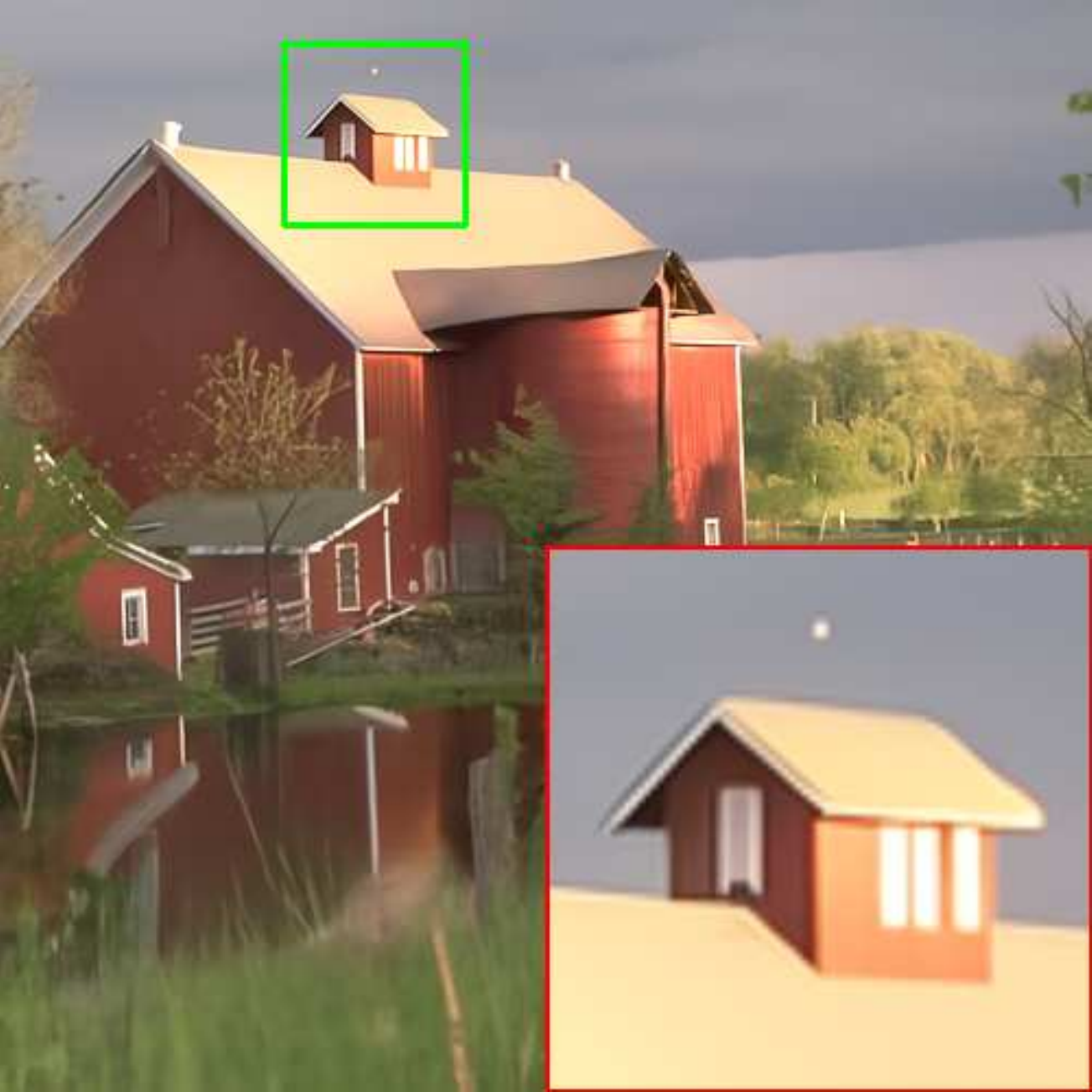}
		\caption{\tiny DRUNet \cite{ZhangL2021} /29.38dB}
	\end{subfigure}
    \centering
	\begin{subfigure}{0.194\linewidth}
		\centering
		\includegraphics[width=0.98\linewidth]{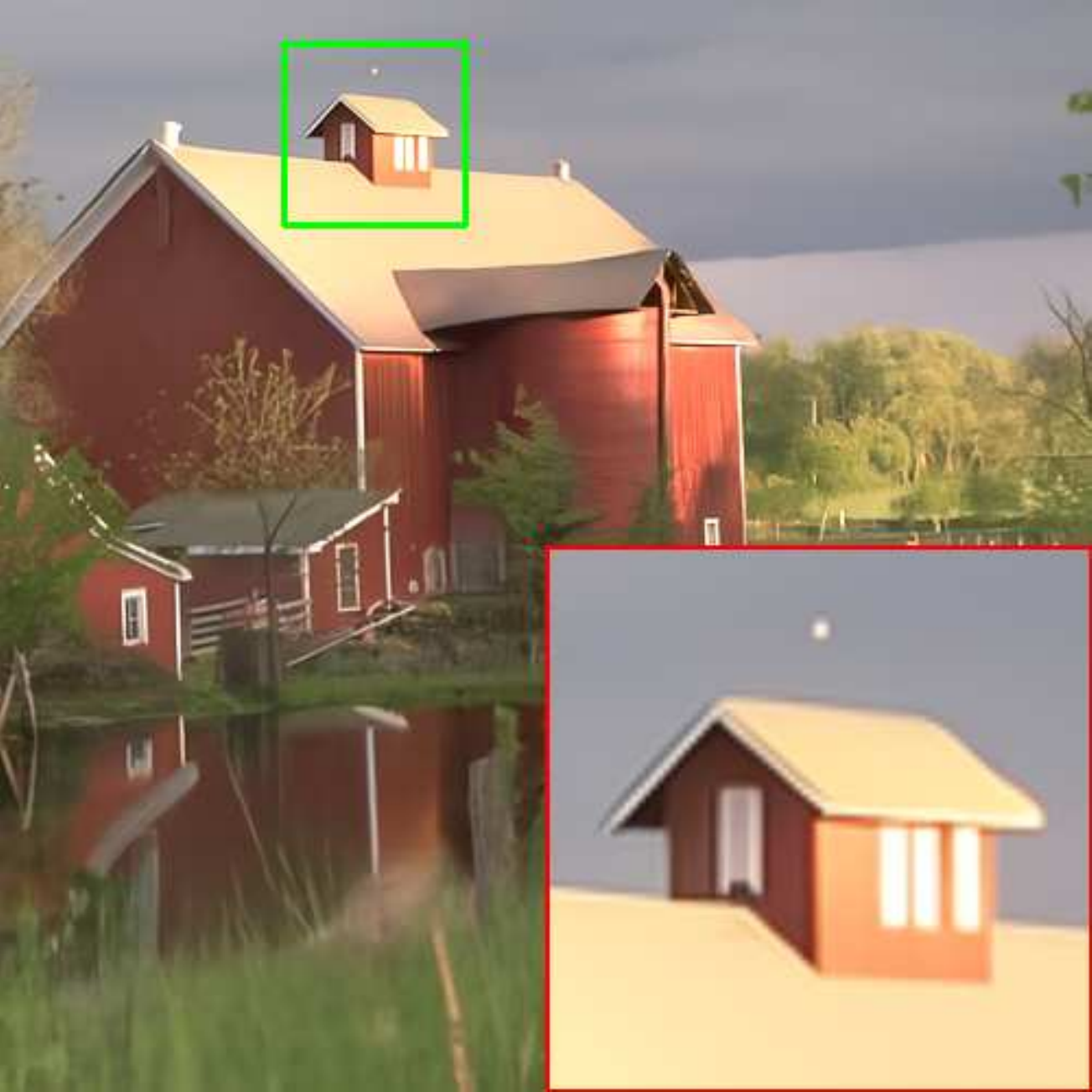}
		\caption{\tiny DCBDNet/29.07dB}
	\end{subfigure}
\caption{Visual results on the image ``kodim22'' from Kodak24 dataset.}
\label{fig:kodim22}
\end{figure*}

The visual results of color image AWGN removal at noise level 50 can be seen in Fig. \ref{fig:kodim22}, where the ``kodim22'' image from the Kodak24 dataset is used. It can be seen that the small white dot in the red box is overly smoothed by CBM3D and AirNet, while it is well preserved by our DCBDNet.

\subsection{Spatially variant AWGN removal}\label{variant_AWGN}
We follow the same technique in FFDNet \cite{Zhang2018} to obtain the additive white Gaussian noise (AWGN) with spatial variation, the bivariate Gaussian probability density function is used to simulate a spatial variant noise distribution, as shown in Eqn. (\ref{eq:Gaussian_function}).
\begin{equation}\label{eq:Gaussian_function}
\begin{aligned}
p &= 3(1 - m)^2 e^{(-m^2 - (n + 1)^2)} \\
  &- 10(\frac{m}{5} - m^3 - n^5)e^{(-m^2 - n^2)} \\
  &- \frac{1}{3} e^{(-(m + 1)^2 - n^2)}.
\end{aligned}
\end{equation}

For an estimated spatially variant noise distribution $p$, the non-uniform noise level map $M$ can be obtained by Eqn. (\ref{eq:noise_level_map}).
\begin{equation}\label{eq:noise_level_map}
\begin{aligned}
M = \lambda\frac{p - min(p)}{max(p) - min(p)},
\end{aligned}
\end{equation}
\noindent where the $\lambda$ is a coefficient used to control the intensity of the noise. $\lambda$ and the normalized distribution $p$ are multiplied to generate the spatial variant noise level map. The spatially variant AWGN can be obtained by Eqn. (\ref{eq:noise_level}).
\begin{equation}\label{eq:noise_level}
\begin{aligned}
N = M\cdot D,
\end{aligned}
\end{equation}
\noindent where $D$ is the standard normal distribution $\mathcal{N}(0, 1)$. The clean image $x$ and the noise level $N$ were then added at the element level to gain spatially variant AWGN image $y$, which is represented as $y = x + N$.

\begin{figure*}[htbp]
	\centering
	\begin{subfigure}{0.194\linewidth}
		\centering
		\includegraphics[width=0.98\linewidth]{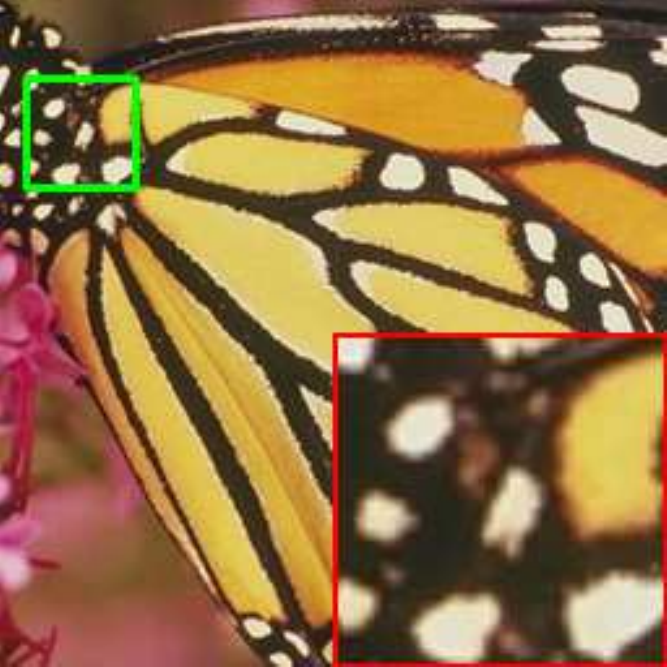}
		\caption{}
	\end{subfigure}
    \centering
	\begin{subfigure}{0.194\linewidth}
		\centering
		\includegraphics[width=0.98\linewidth]{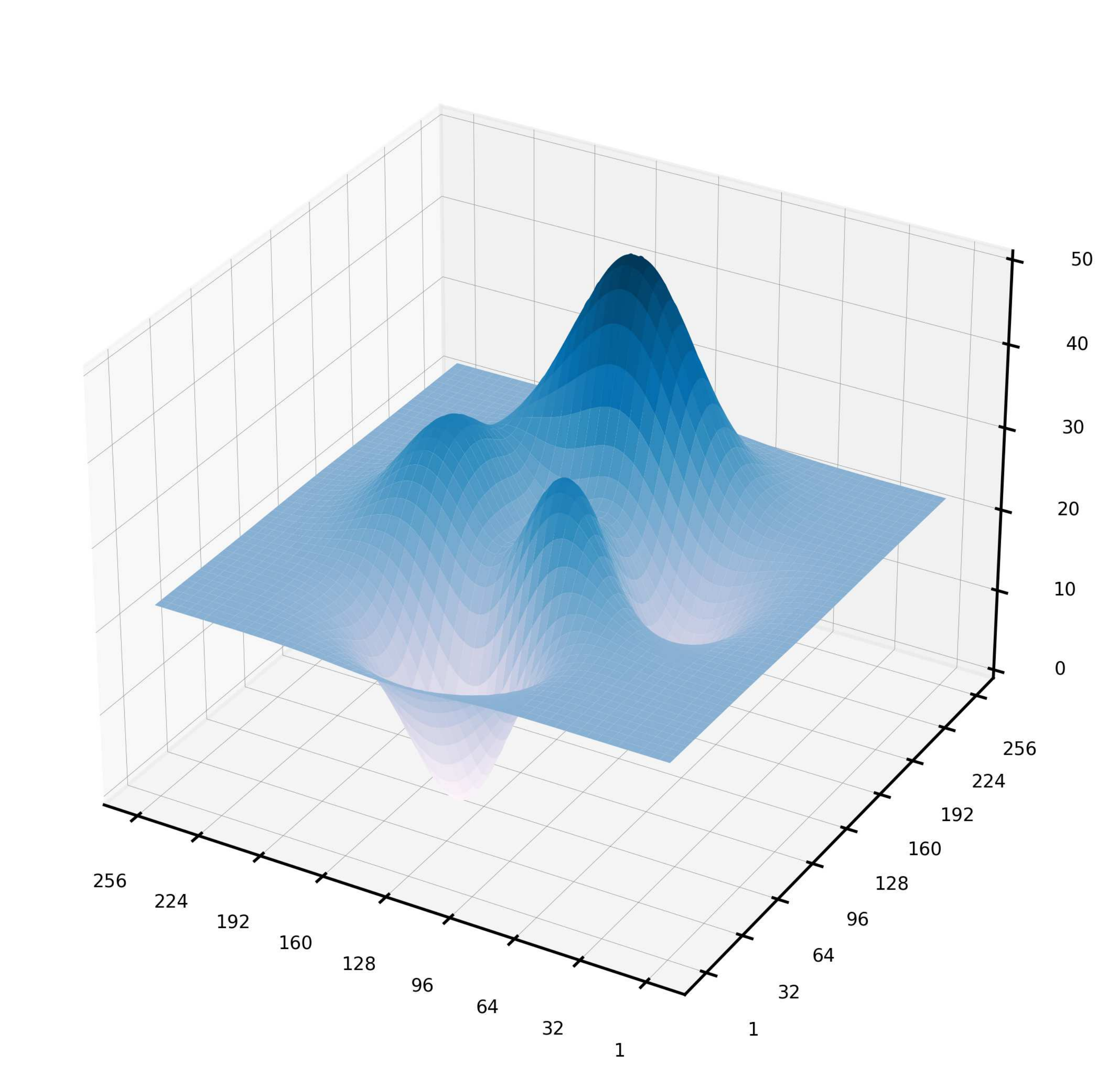}
		\caption{}
	\end{subfigure}
    \centering
	\begin{subfigure}{0.194\linewidth}
		\centering
		\includegraphics[width=0.98\linewidth]{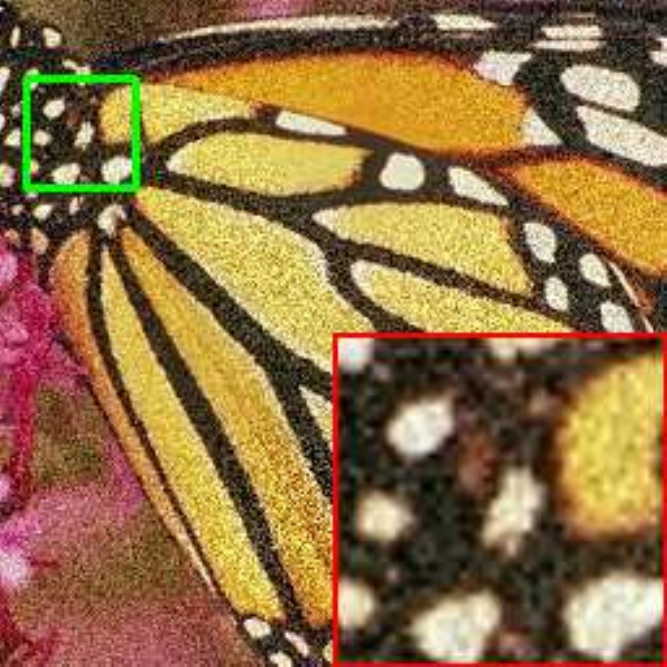}
		\caption{}
	\end{subfigure}
    \centering
	\begin{subfigure}{0.194\linewidth}
		\centering
		\includegraphics[width=0.98\linewidth]{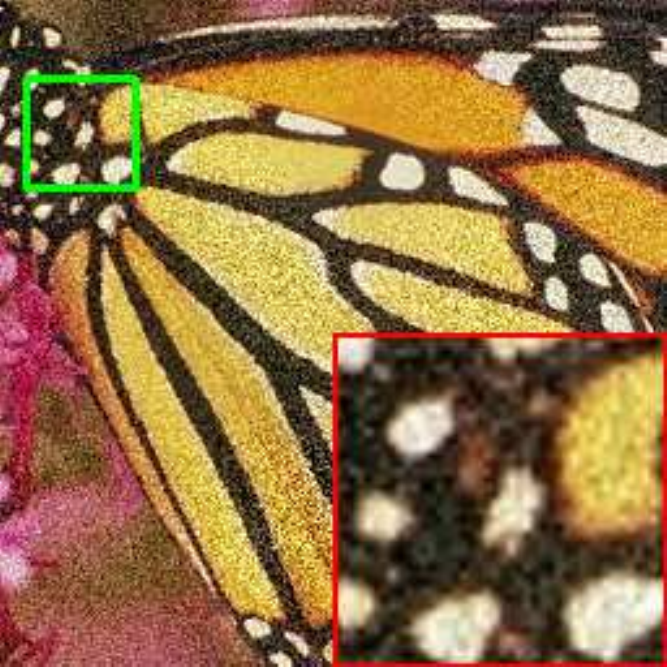}
		\caption{}
	\end{subfigure}
    \centering
	\begin{subfigure}{0.194\linewidth}
		\centering
		\includegraphics[width=0.98\linewidth]{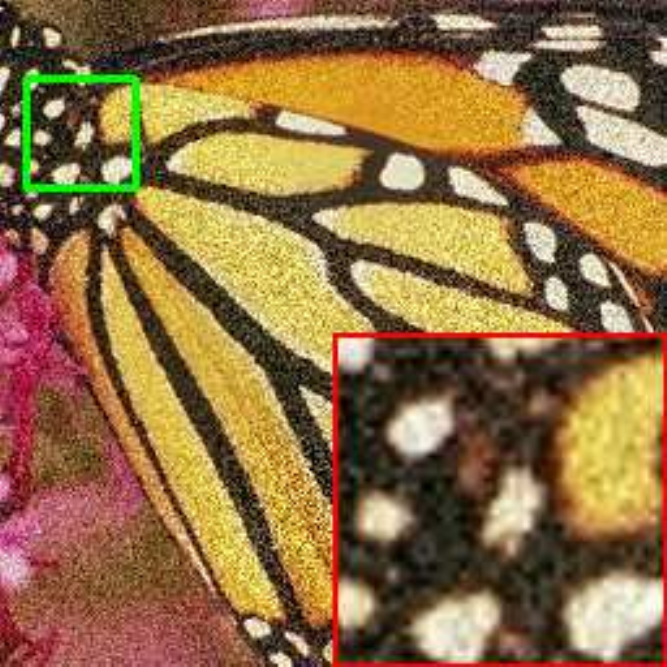}
		\caption{}
	\end{subfigure}
    \begin{subfigure}{0.194\linewidth}
		\centering
		\includegraphics[width=0.98\linewidth]{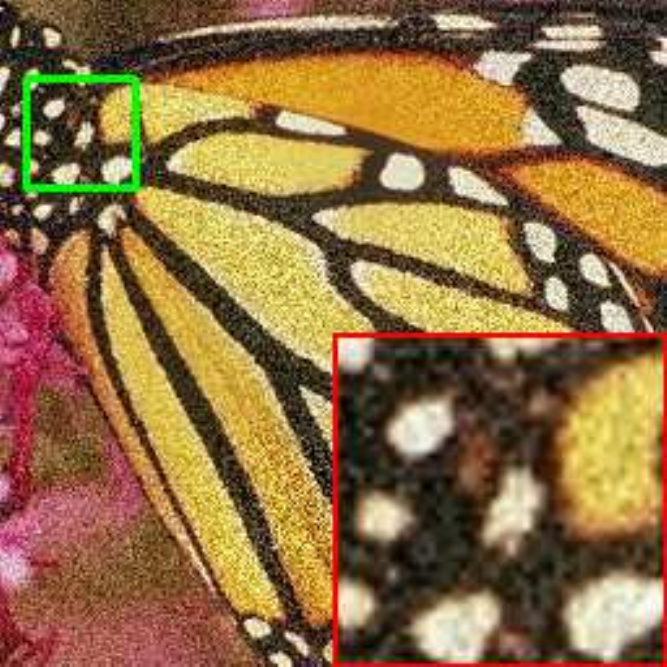}
		\caption{}
	\end{subfigure}
    \centering
	\begin{subfigure}{0.194\linewidth}
		\centering
		\includegraphics[width=0.98\linewidth]{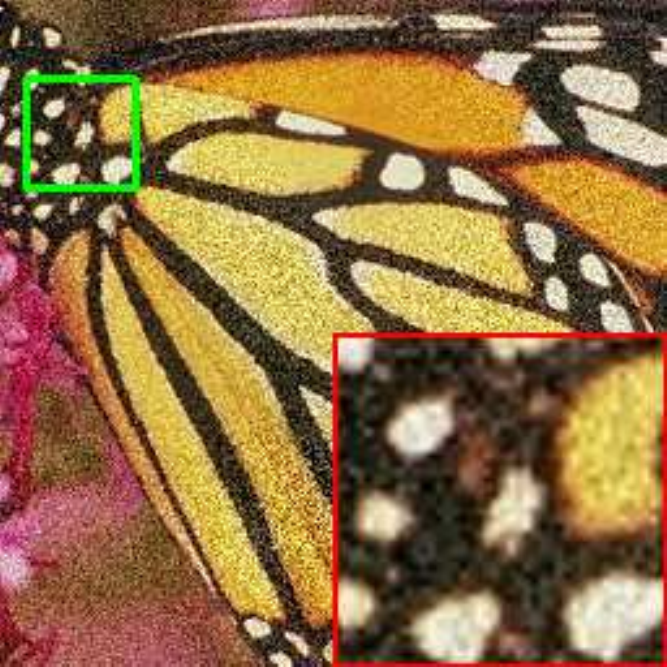}
		\caption{}
	\end{subfigure}
    \centering
	\begin{subfigure}{0.194\linewidth}
		\centering
		\includegraphics[width=0.98\linewidth]{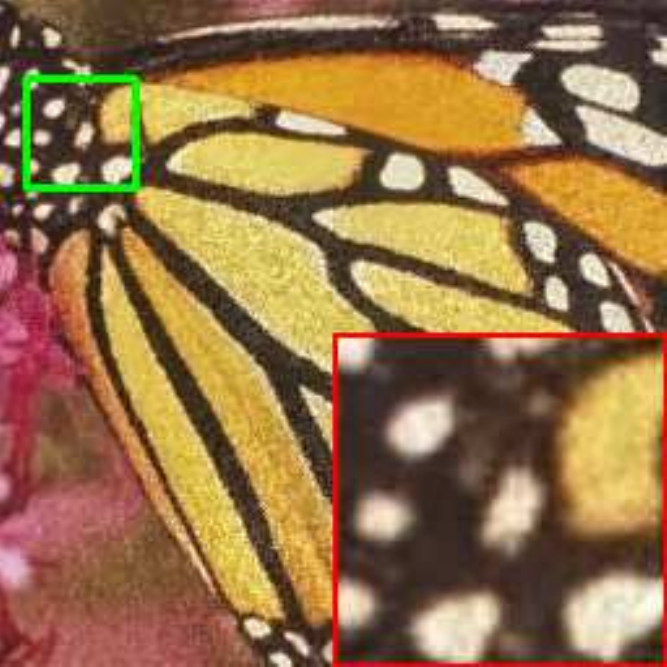}
		\caption{}
	\end{subfigure}
    \centering
	\begin{subfigure}{0.194\linewidth}
		\centering
		\includegraphics[width=0.98\linewidth]{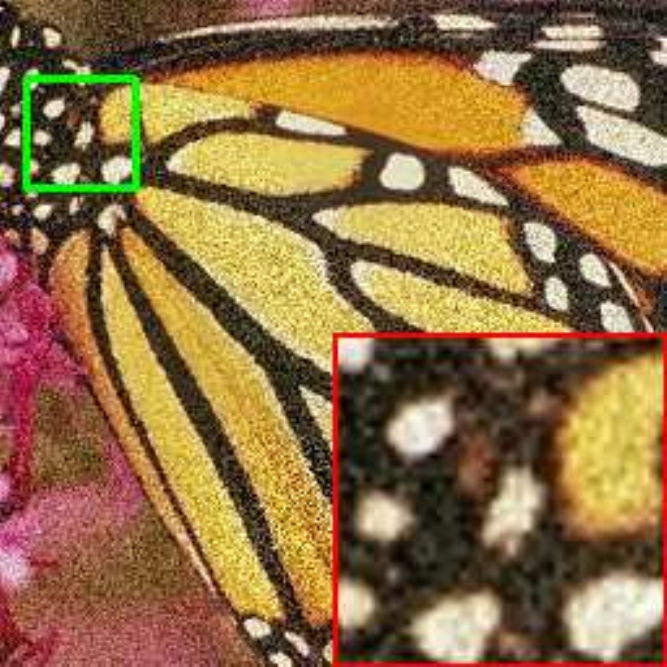}
		\caption{}
	\end{subfigure}
    \centering
	\begin{subfigure}{0.194\linewidth}
		\centering
		\includegraphics[width=0.98\linewidth]{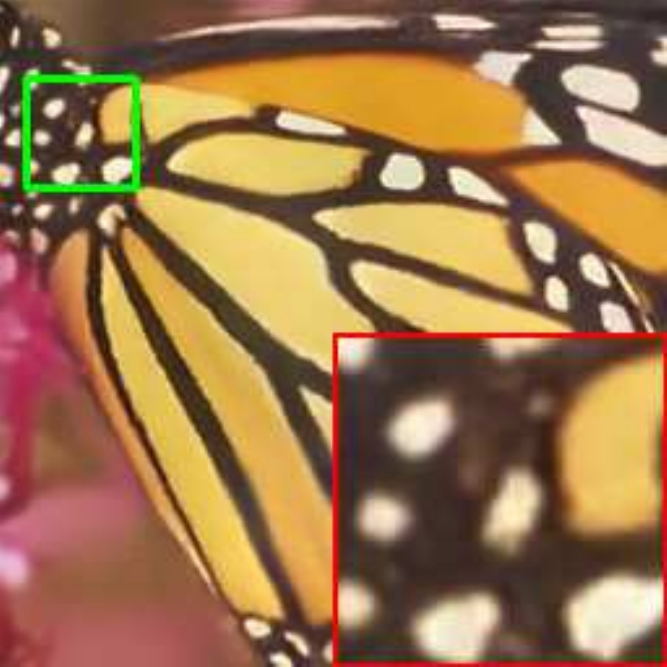}
		\caption{}
	\end{subfigure}
\caption{Visual results on the image ``butterfly'' from Set5 dataset with spatially variant AWGN. (a) Ground-truth image / (PSNR (dB) /SSIM), (b) Non-uniform noise level map, (c) Noisy image degraded by spatially variant AWGN / (20.74/0.552), (d) CDnCNN-B / (20.83/0.555), (e) ADNet / ((20.78/0.592), (f) BUIFD / (20.74/0.590), (g) AirNet / (20.64/0.586), (h) VDN / (24.72/0.743), (i) VDIR / (20.95/0.596), (j) DCBDNet / (23.85/0.847).}
\label{fig:butterfly}
\end{figure*}

To verify the effectiveness of the proposed DCBDNet for spatially variant AWGN removal, we choose the ``butterfly'' image from the Set5 dataset for visual evaluation, and the $\lambda$ in Eqn. (\ref{eq:noise_level_map}) was set as 50. Fig. \ref{fig:butterfly} shows the visual comparison of the compared methods. It can be seen that the DCBDNet and VDN are effective to eliminate the spatially variant AWGN.

\subsection{Real noisy image denoising evaluation}
For the evaluation of the real noisy image denoising, the SIDD validation set, DND sRGB images, RNI15, and Nam datasets were used. The SIDD validation set, DND sRGB images, and Nam dataset contain noisy images and the near noise-free counterparts, the quantitative evaluation can be implemented. Table \ref{tab:SIDD_DND} lists the averaged PSNR and SSIM values of different methods on the SIDD validation set and DND sRGB images. One can see that our proposed DCBDNet model achieves more effective denoising performance than MCWNNM, TWSC, DnCNN-B, CBDNet, and RIDNet, however the VDN and VDIR outperform our model, however both of them have much more complex model structures than ours.

\begin{table*}[htbp]\scriptsize
\centering
\caption{The averaged PSNR(dB), SSIM and FSIM of different denoising methods on the SIDD validation set and DND sRGB images. The top three results on each noise level are marked in red, blue and green in order.}
\label{tab:SIDD_DND}
\begin{tabular}{cccccccccc}
\cline{1-10}
Dataset & Methods & MCWNNM \cite{Xu2017} & TWSC \cite{XuZ2018} & DnCNN-B \cite{Zhang2017} & CBDNet \cite{Guo2019} & RIDNet \cite{Anwar2019} & VDN \cite{YueYZM2019} & VDIR \cite{SohC2022} & DCBDNet\\
\cline{1-10}
\multirow{2}*{SIDD} & PSNR & 33.40 & 35.33 & 23.66 & 30.78 & 38.71 & \textcolor{red}{39.28}	& \textcolor{blue}{39.26} & \textcolor{green}{38.94}\\
\cline{2-10}
\multicolumn{1}{c}{} & SSIM & 0.879 & 0.933 & 0.583 & 0.951 & \textcolor{green}{0.954} & \textcolor{red}{0.957}	& \textcolor{blue}{0.955} & 0.953\\
\cline{1-10}
\multirow{2}*{DND} & PSNR & 37.38 & 37.94	& 37.90	& 38.06 & 39.26 & \textcolor{blue}{39.38} & \textcolor{red}{39.63} & \textcolor{green}{39.37}\\
\cline{2-10}
\multicolumn{1}{c}{} & SSIM & 0.929 & 0.940 & 0.943 & 0.942 & \textcolor{red}{0.953} & \textcolor{blue}{0.952}	& \textcolor{red}{0.953} & \textcolor{green}{0.951}\\
\cline{1-10}
\end{tabular}
\end{table*}

\begin{figure*}[htbp]
	\centering
	\begin{subfigure}{0.24\linewidth}
		\centering
		\includegraphics[width=0.98\linewidth]{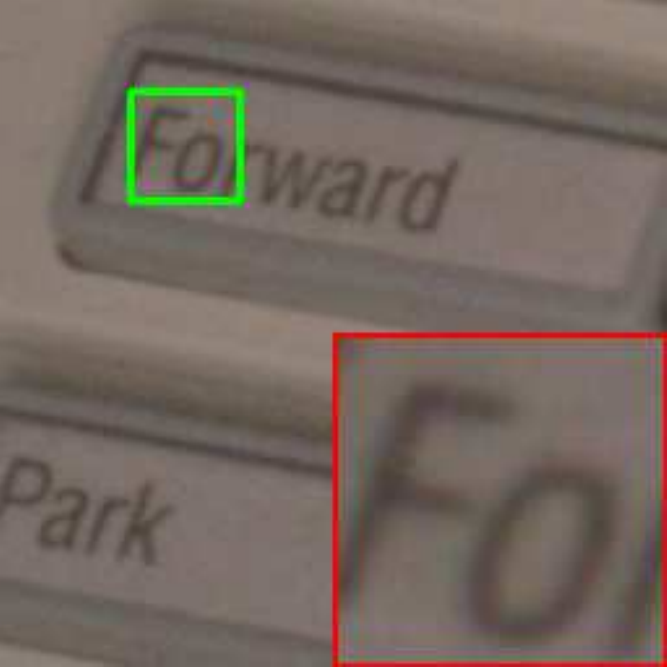}
		\caption{}
	\end{subfigure}
    \centering
	\begin{subfigure}{0.24\linewidth}
		\centering
		\includegraphics[width=0.98\linewidth]{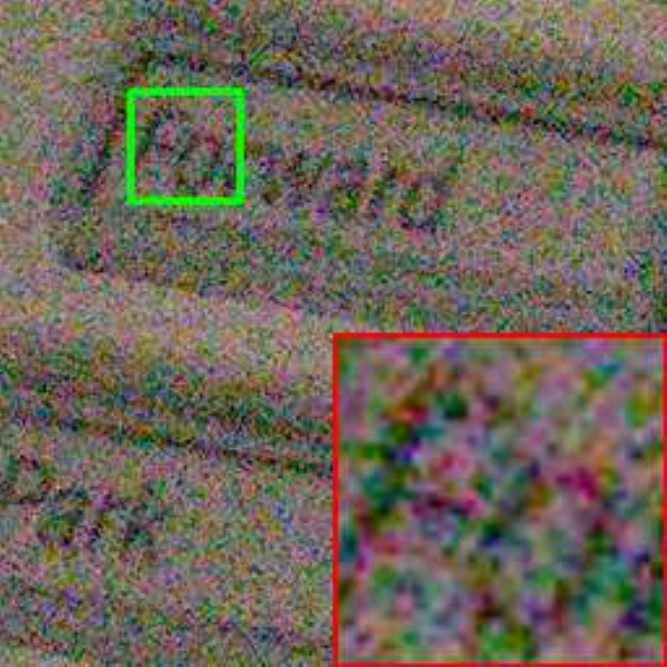}
		\caption{}
	\end{subfigure}
    \centering
	\begin{subfigure}{0.24\linewidth}
		\centering
		\includegraphics[width=0.98\linewidth]{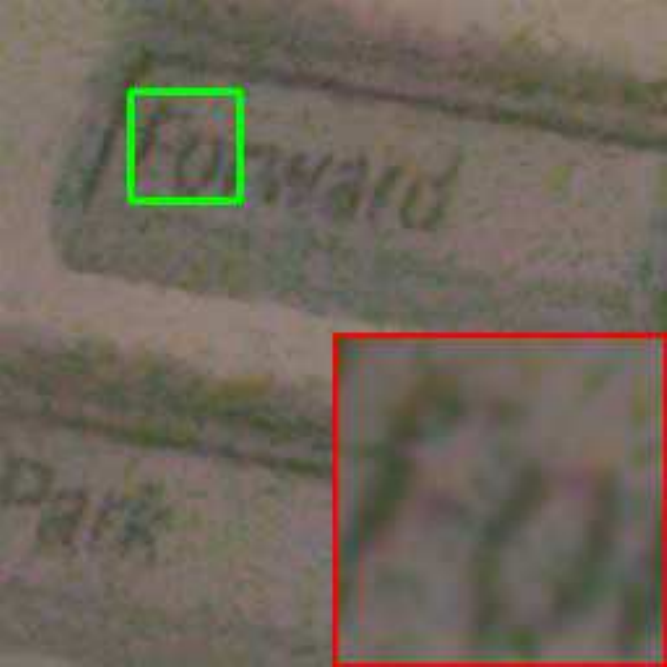}
		\caption{}
	\end{subfigure}
    \centering
	\begin{subfigure}{0.24\linewidth}
		\centering
		\includegraphics[width=0.98\linewidth]{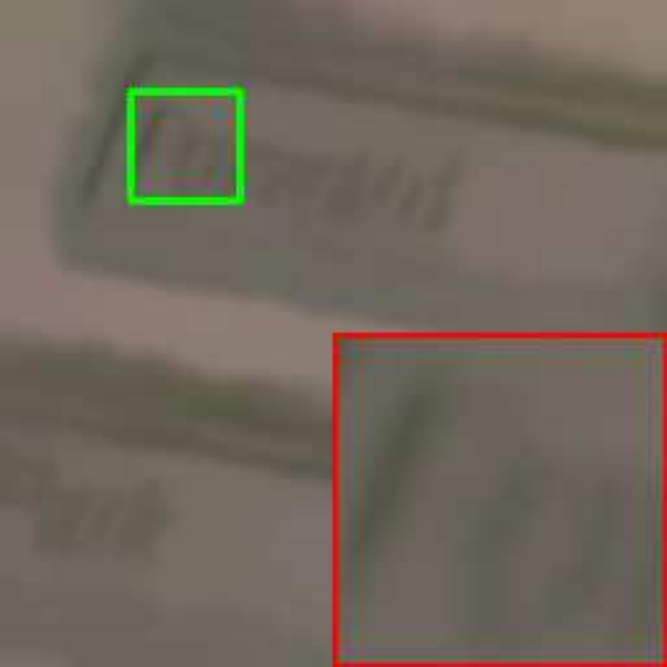}
		\caption{}
	\end{subfigure}
    \centering
	\begin{subfigure}{0.24\linewidth}
		\centering
		\includegraphics[width=0.98\linewidth]{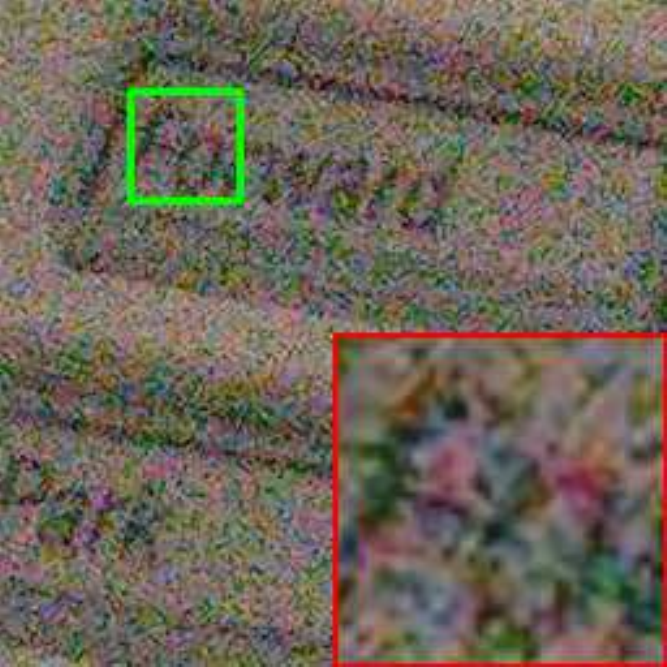}
		\caption{}
	\end{subfigure}
    \centering
	\begin{subfigure}{0.24\linewidth}
		\centering
		\includegraphics[width=0.98\linewidth]{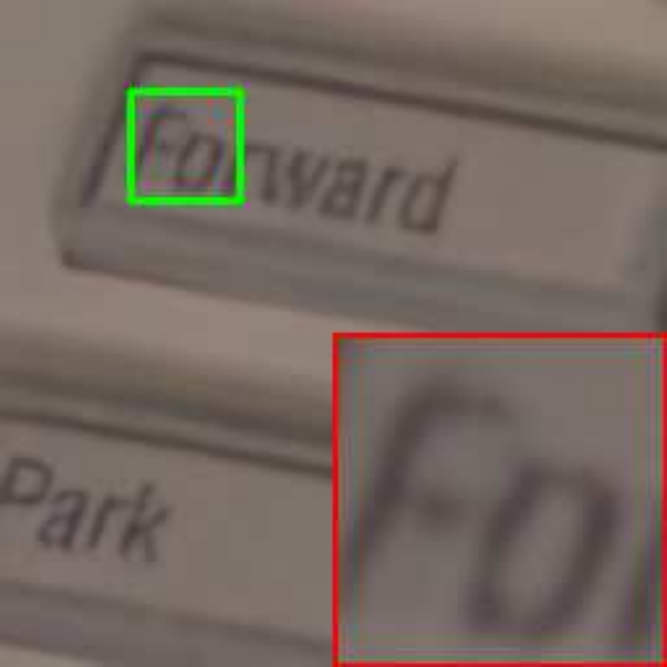}
		\caption{}
	\end{subfigure}
    \centering
	\begin{subfigure}{0.24\linewidth}
		\centering
		\includegraphics[width=0.98\linewidth]{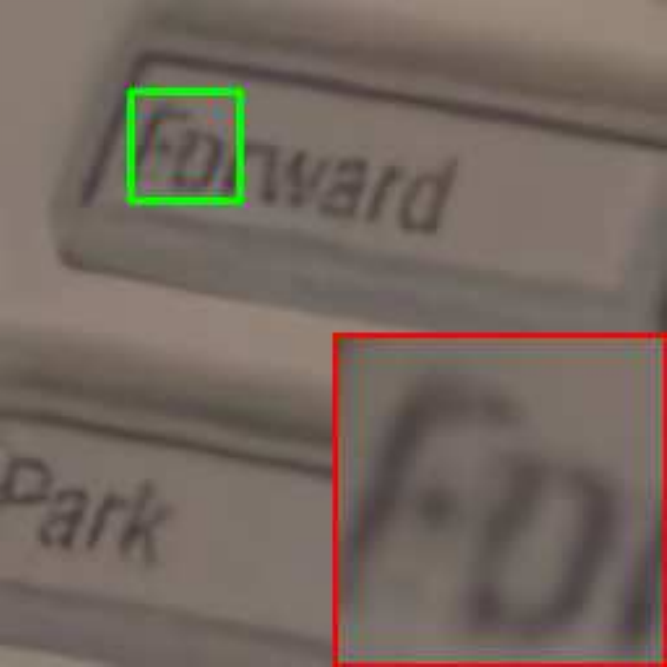}
		\caption{}
	\end{subfigure}
    \centering
	\begin{subfigure}{0.24\linewidth}
		\centering
		\includegraphics[width=0.98\linewidth]{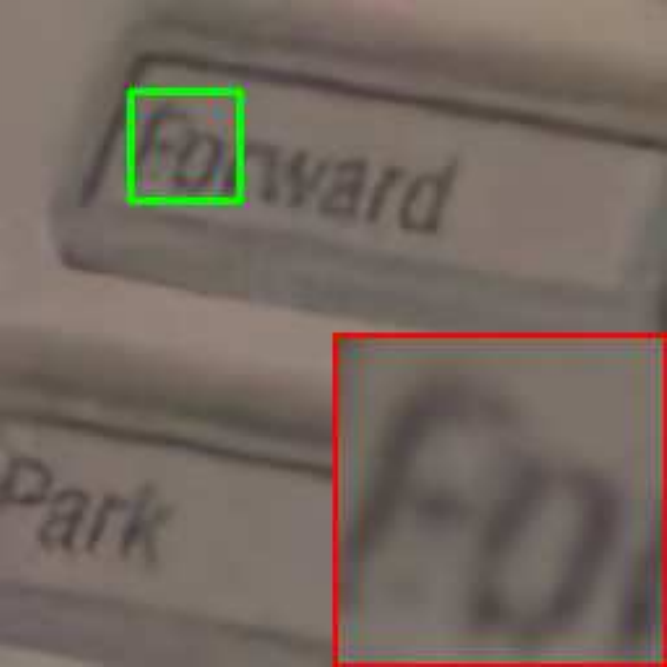}
		\caption{}
	\end{subfigure}
\caption{Visual results on the image ``11\_4'' from SIDD the SIDD validation set. (a) Ground-truth image / (PSNR (dB) /SSIM), (b) Noisy image / (18.25/0.169), (c) MCWNNM / (28.63/0.702),(d) TWSC / (30.41/0.811), (e) CDnCNN-B / (20.76/0.235), (F) VDN / (36.39/0.907), (G) VDIR / (36.35/0.906), (h) DCBDNet / (35.61/0.906).}
\label{fig:11_4}
\end{figure*}

Fig. \ref{fig:11_4} shows the visual comparison from the compared denoising methods. It can be found the VDN, VDIR, and the proposed DCBDNet achieve better visual quality.

\begin{figure*}[htbp]
	\centering
	\begin{subfigure}{0.24\linewidth}
		\centering
		\includegraphics[width=0.98\linewidth]{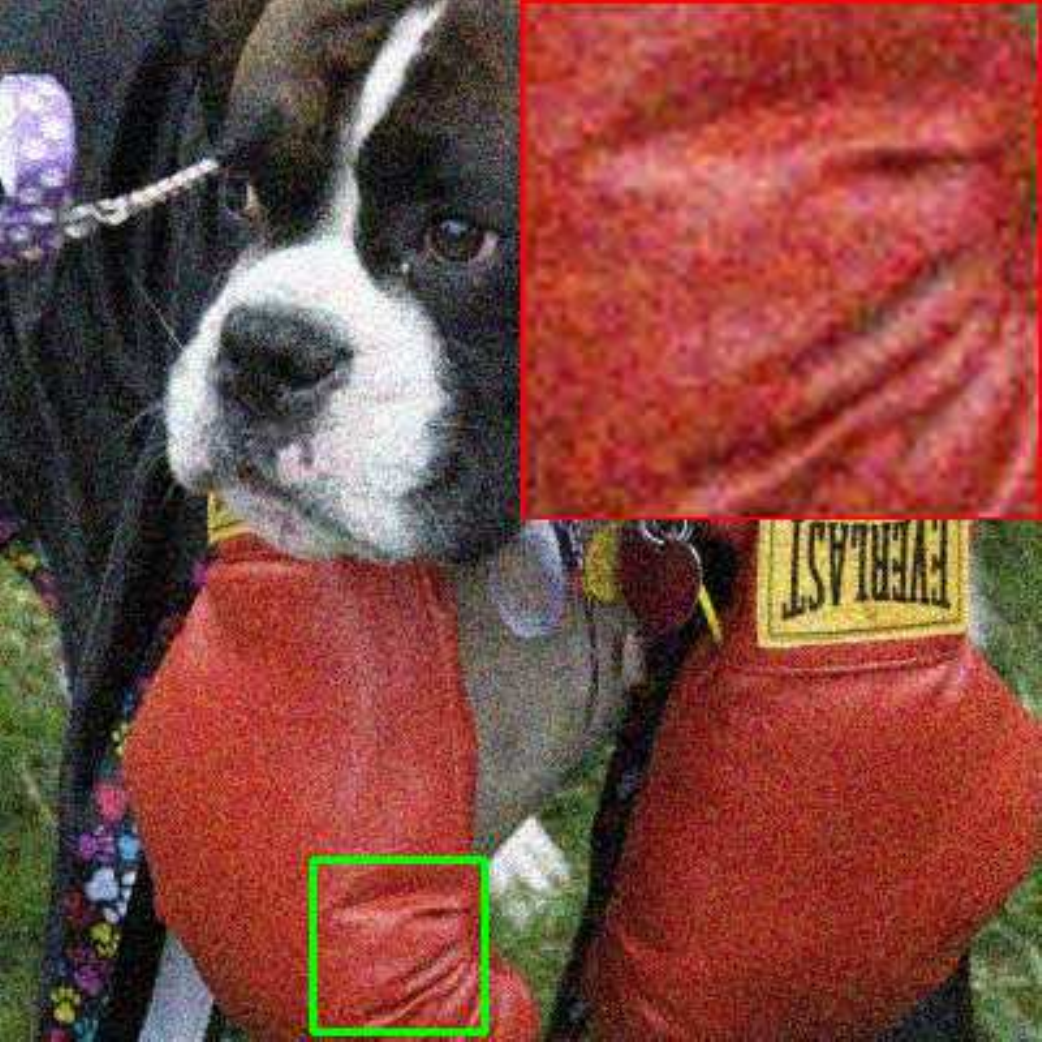}
		\caption{\tiny Noisy}
	\end{subfigure}
    \centering
	\begin{subfigure}{0.24\linewidth}
		\centering
		\includegraphics[width=0.98\linewidth]{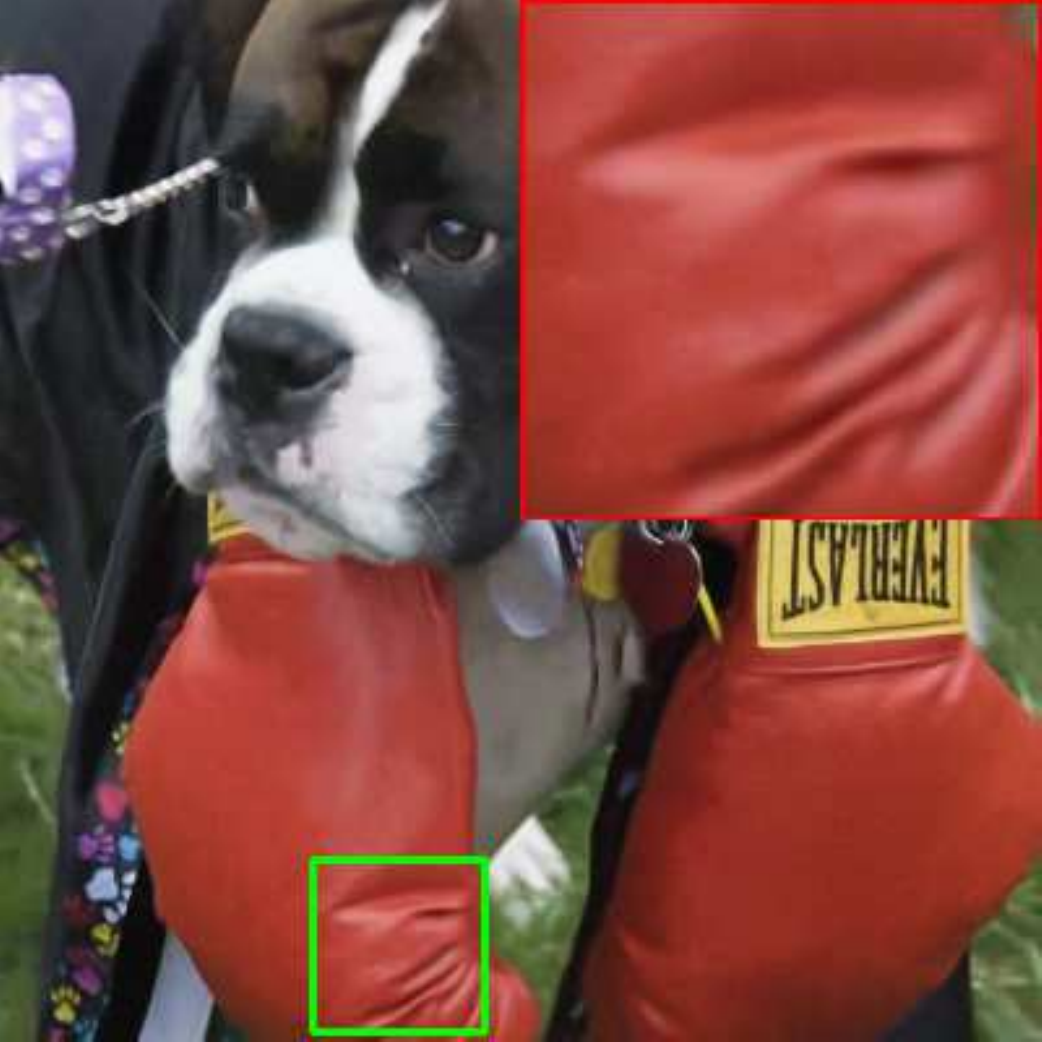}
		\caption{\tiny MCWNNM \cite{Xu2017}}
	\end{subfigure}
    \centering
	\begin{subfigure}{0.24\linewidth}
		\centering
		\includegraphics[width=0.98\linewidth]{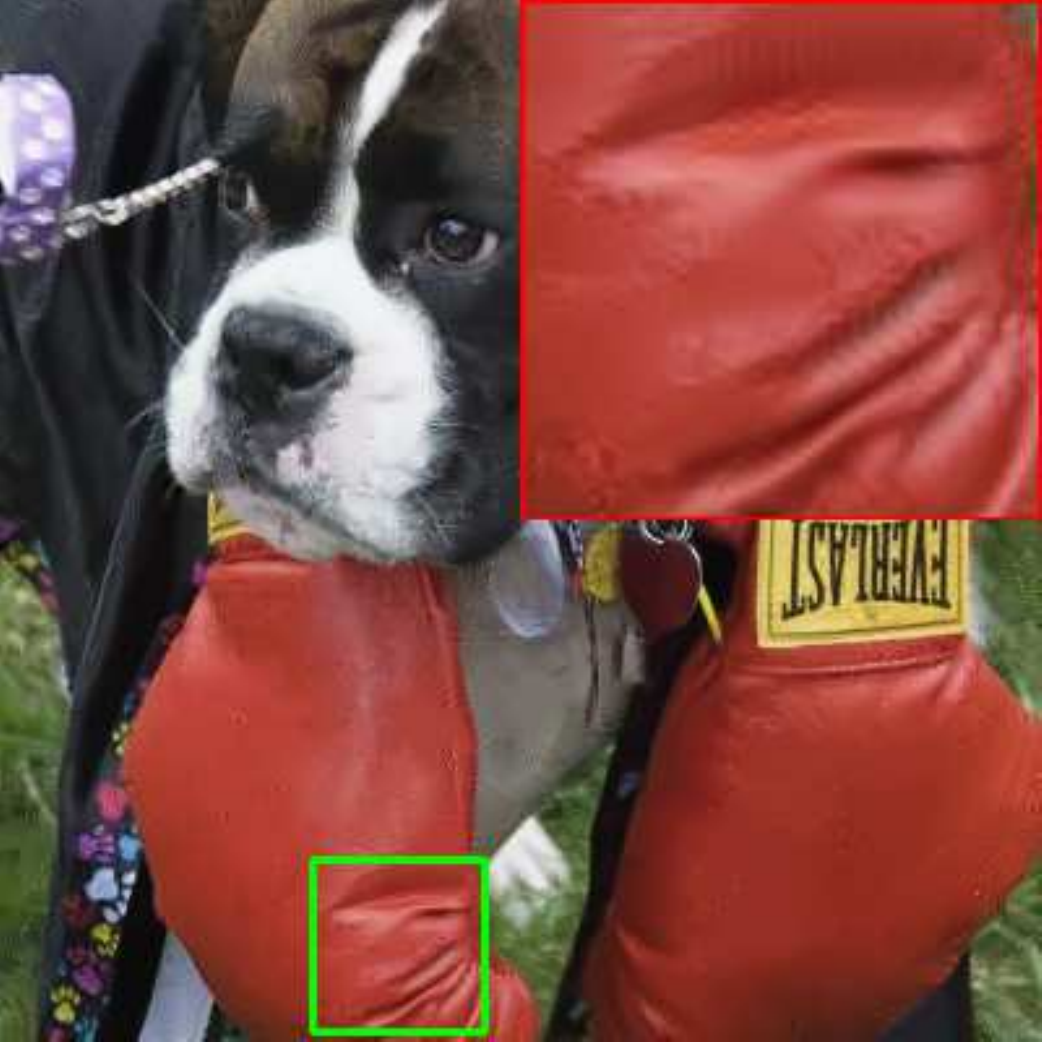}
		\caption{\tiny TWSC \cite{XuZ2018}}
	\end{subfigure}
    \centering
	\begin{subfigure}{0.24\linewidth}
		\centering
		\includegraphics[width=0.98\linewidth]{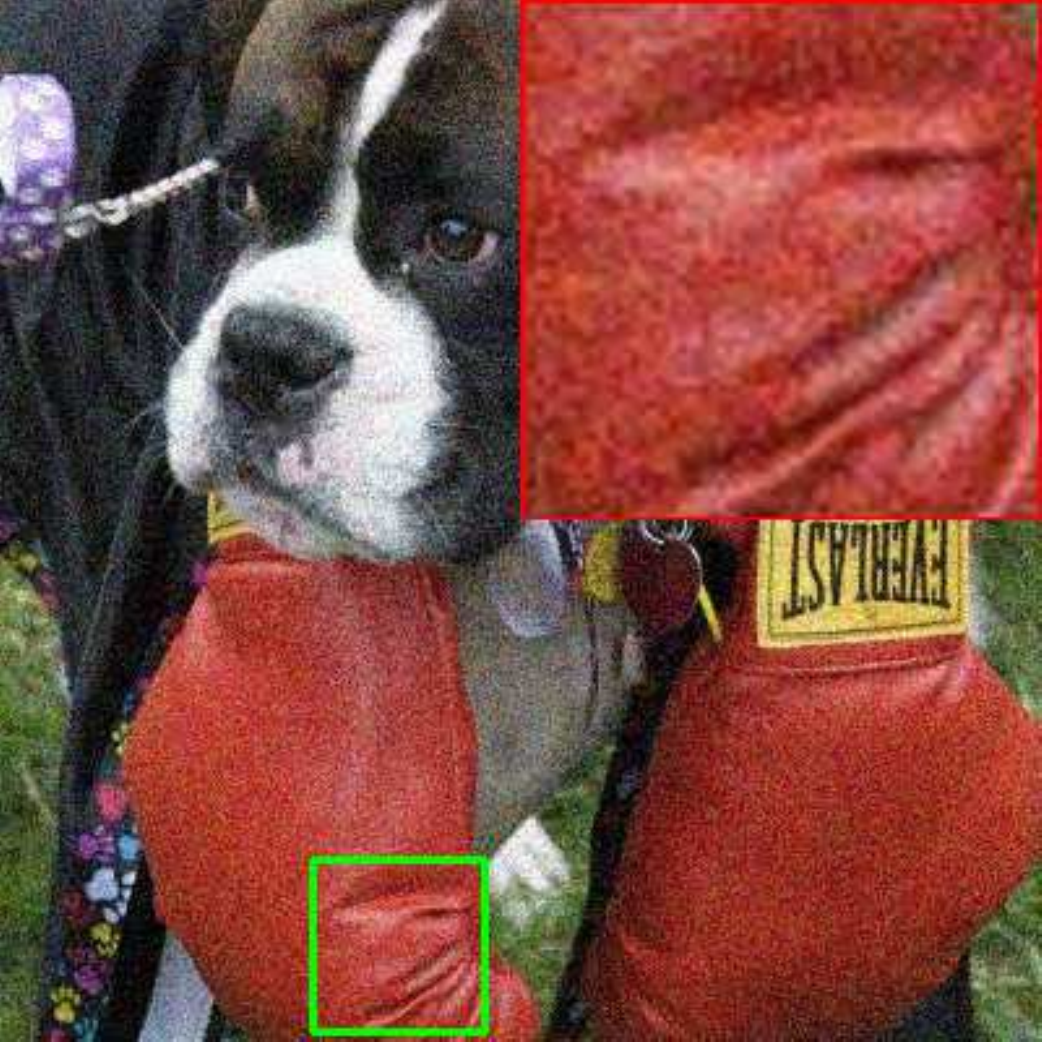}
		\caption{\tiny CDnCNN-B \cite{Zhang2017}}
	\end{subfigure}
    \centering
	\begin{subfigure}{0.24\linewidth}
		\centering
		\includegraphics[width=0.98\linewidth]{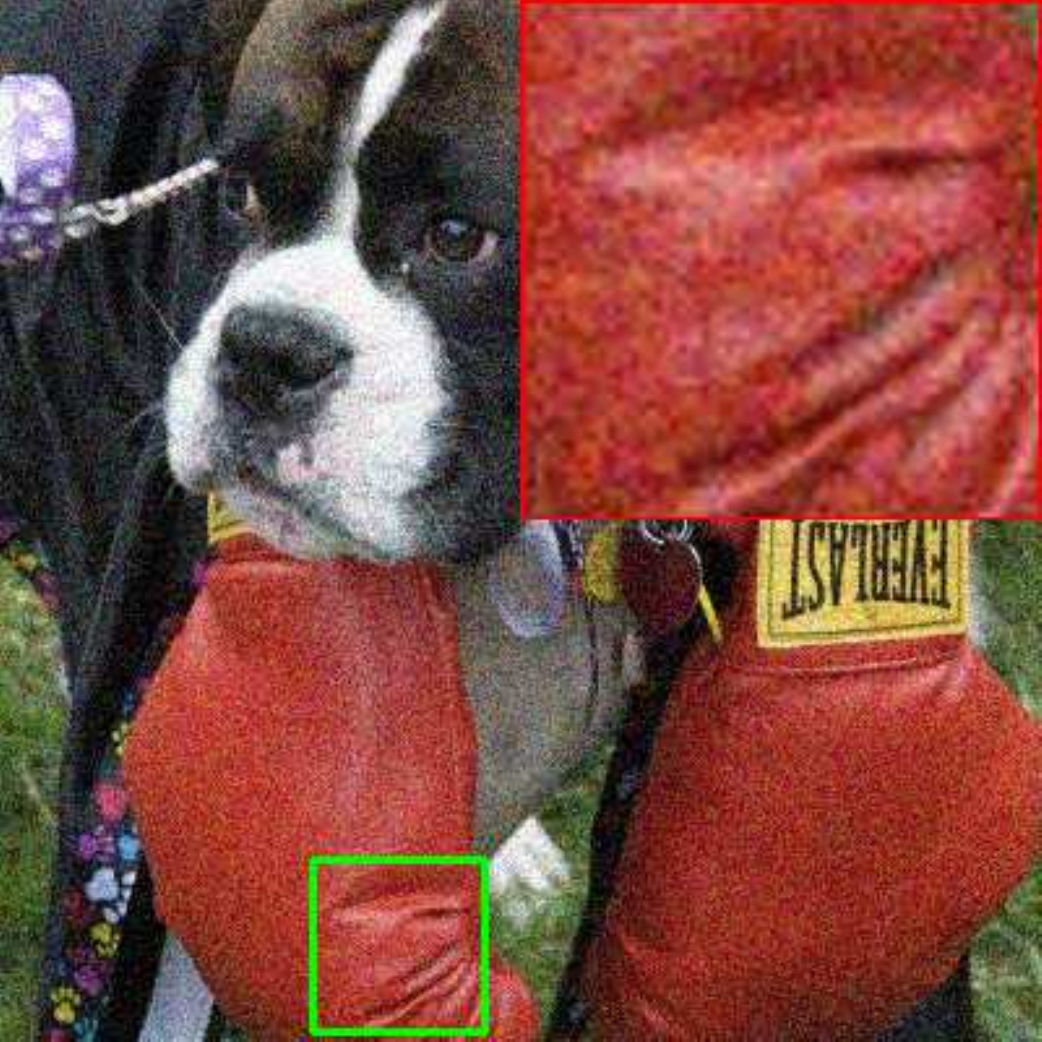}
		\caption{\tiny ADNet \cite{TianX2020}}
	\end{subfigure}
    \centering
	\begin{subfigure}{0.24\linewidth}
		\centering
		\includegraphics[width=0.98\linewidth]{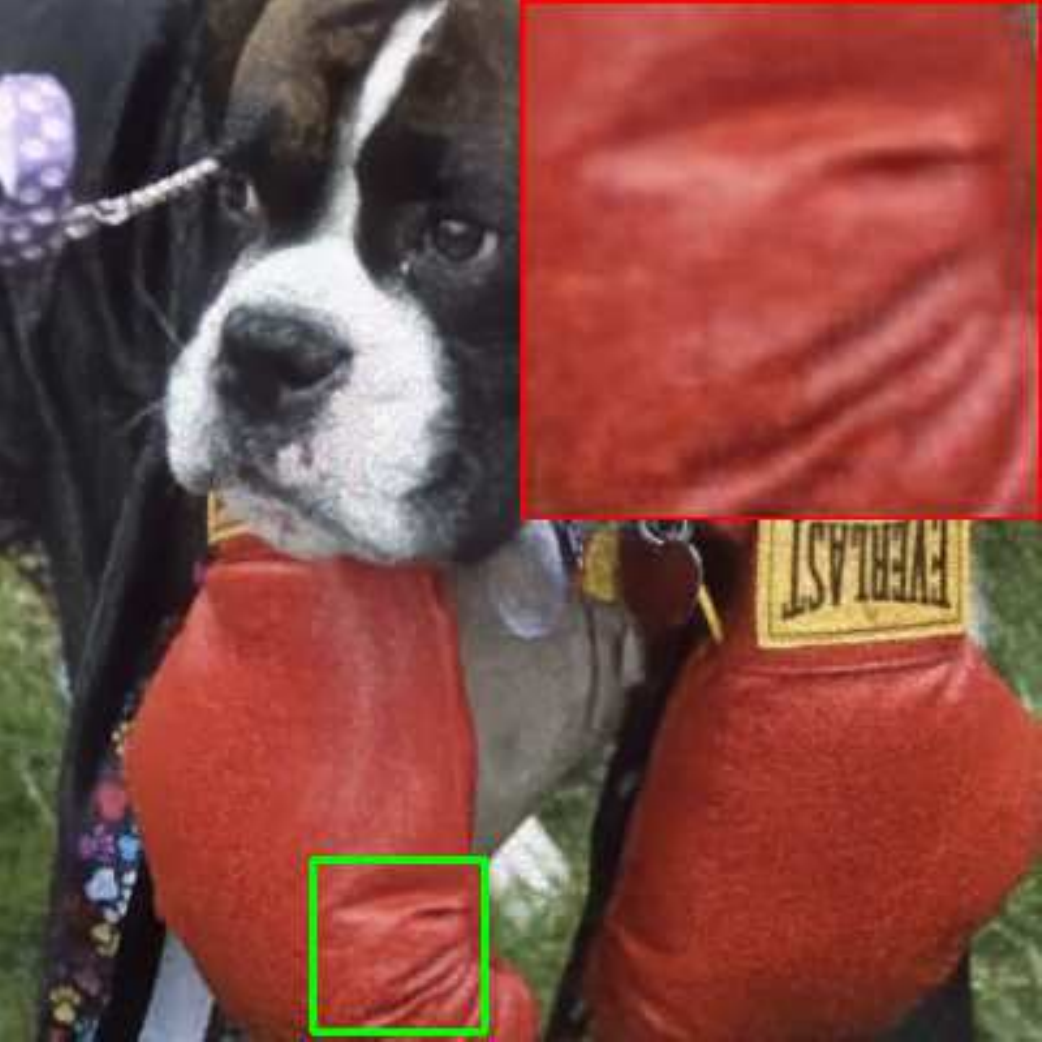}
		\caption{\tiny VDN \cite{ZhangZ2017}}
	\end{subfigure}
    \centering
	\begin{subfigure}{0.24\linewidth}
		\centering
		\includegraphics[width=0.98\linewidth]{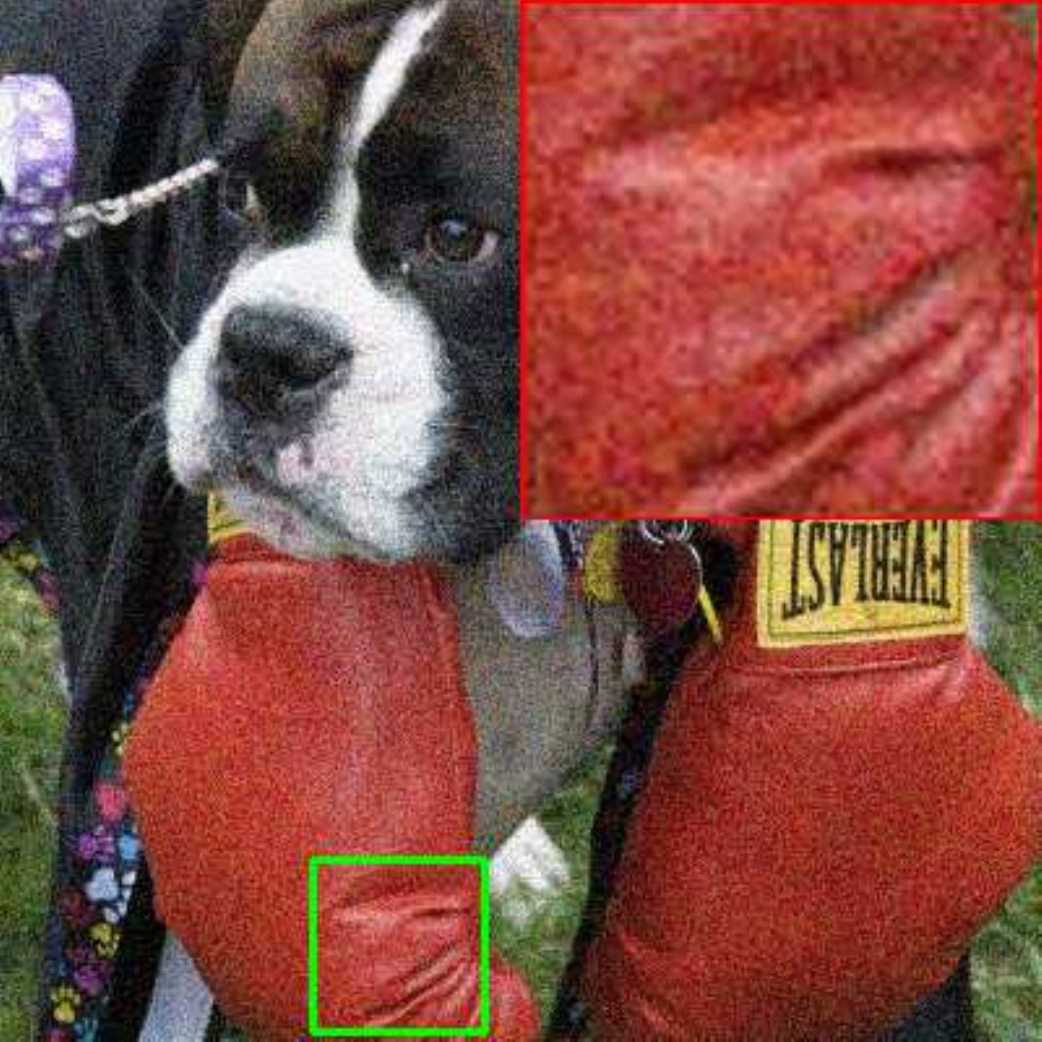}
		\caption{\tiny VDIR \cite{YueYZM2019}}
	\end{subfigure}
    \centering
	\begin{subfigure}{0.24\linewidth}
		\centering
		\includegraphics[width=0.98\linewidth]{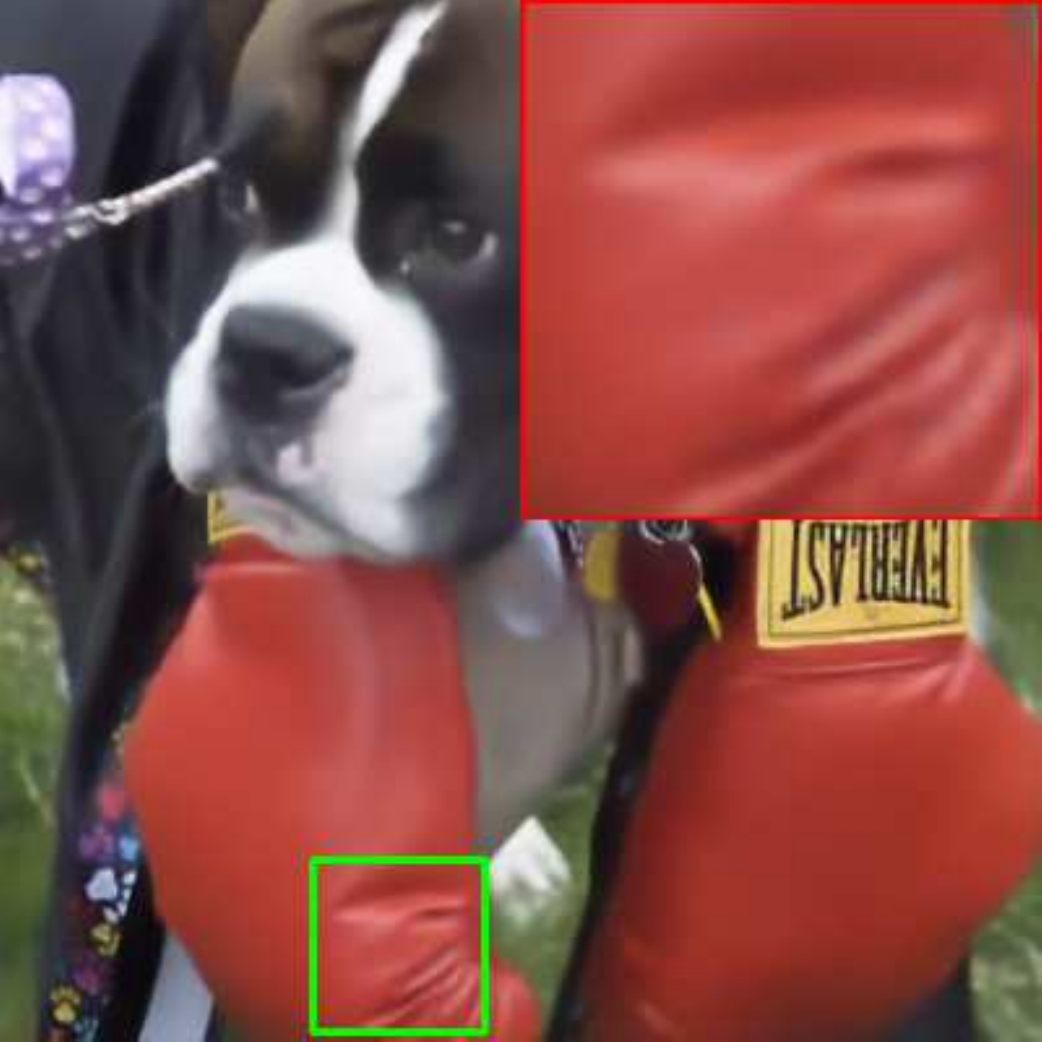}
		\caption{\tiny DCBDNet}
	\end{subfigure}
\caption{Visual results on the image ``Dog'' from RNI15 dataset.}
\label{fig:Dog}
\end{figure*}

The RNI15 dataset consists of 15 real noisy images without the corresponding ground truth images, therefore the quantitative comparison can not be implemented. The visual results of the compared denoising models on the image ``Dog" can be seen in Fig. \ref{fig:Dog}, where one can see that the TWSC, CDnCNN-B, VDN, and VDIR produced inferior visual results. Furthermore, the MCWNNM and the proposed DCBDNet obtained visual appealing results.

\begin{figure*}[htbp]
	\centering
	\begin{subfigure}{0.194\linewidth}
		\centering
		\includegraphics[width=0.98\linewidth]{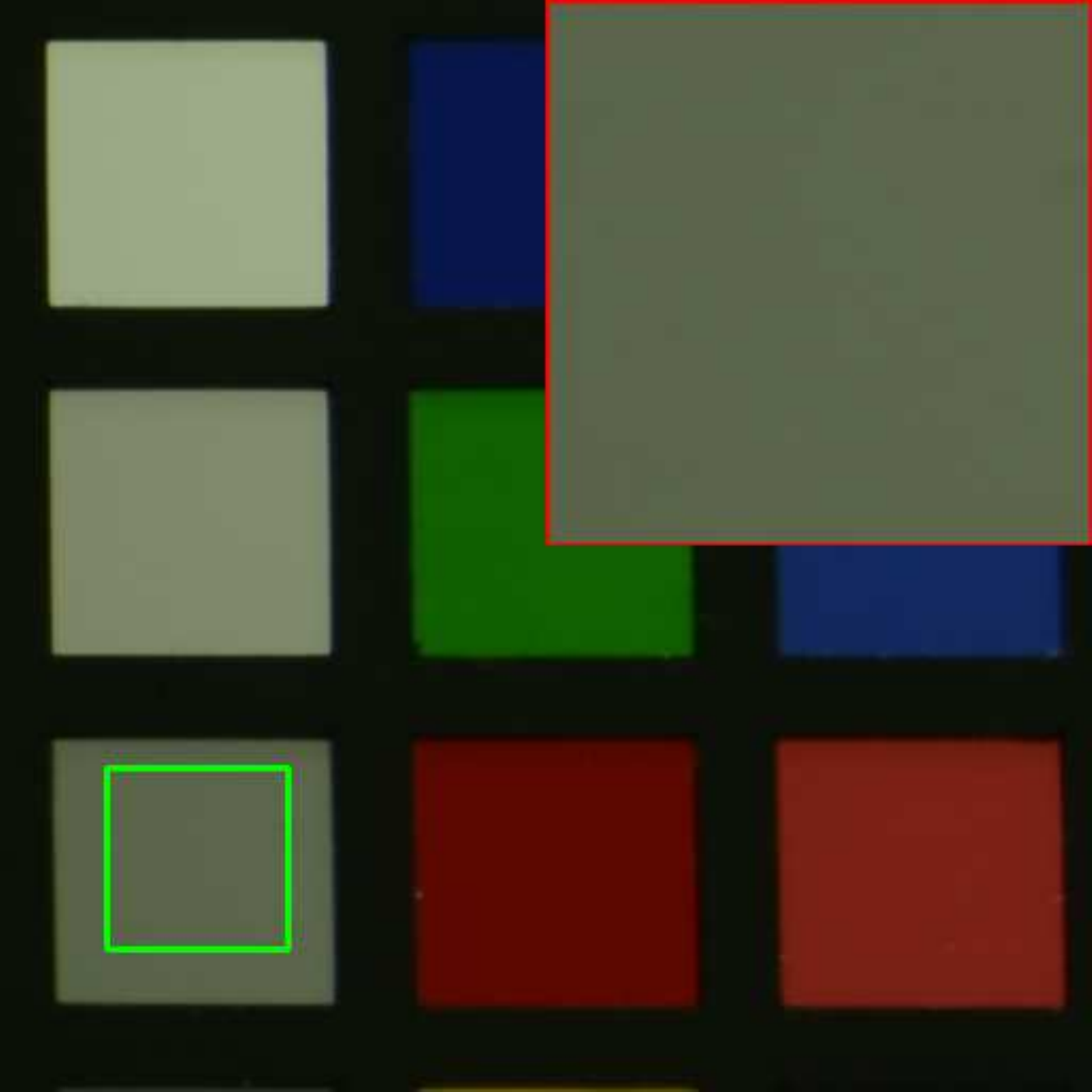}
		\caption{}
	\end{subfigure}
    \centering
	\begin{subfigure}{0.194\linewidth}
		\centering
		\includegraphics[width=0.98\linewidth]{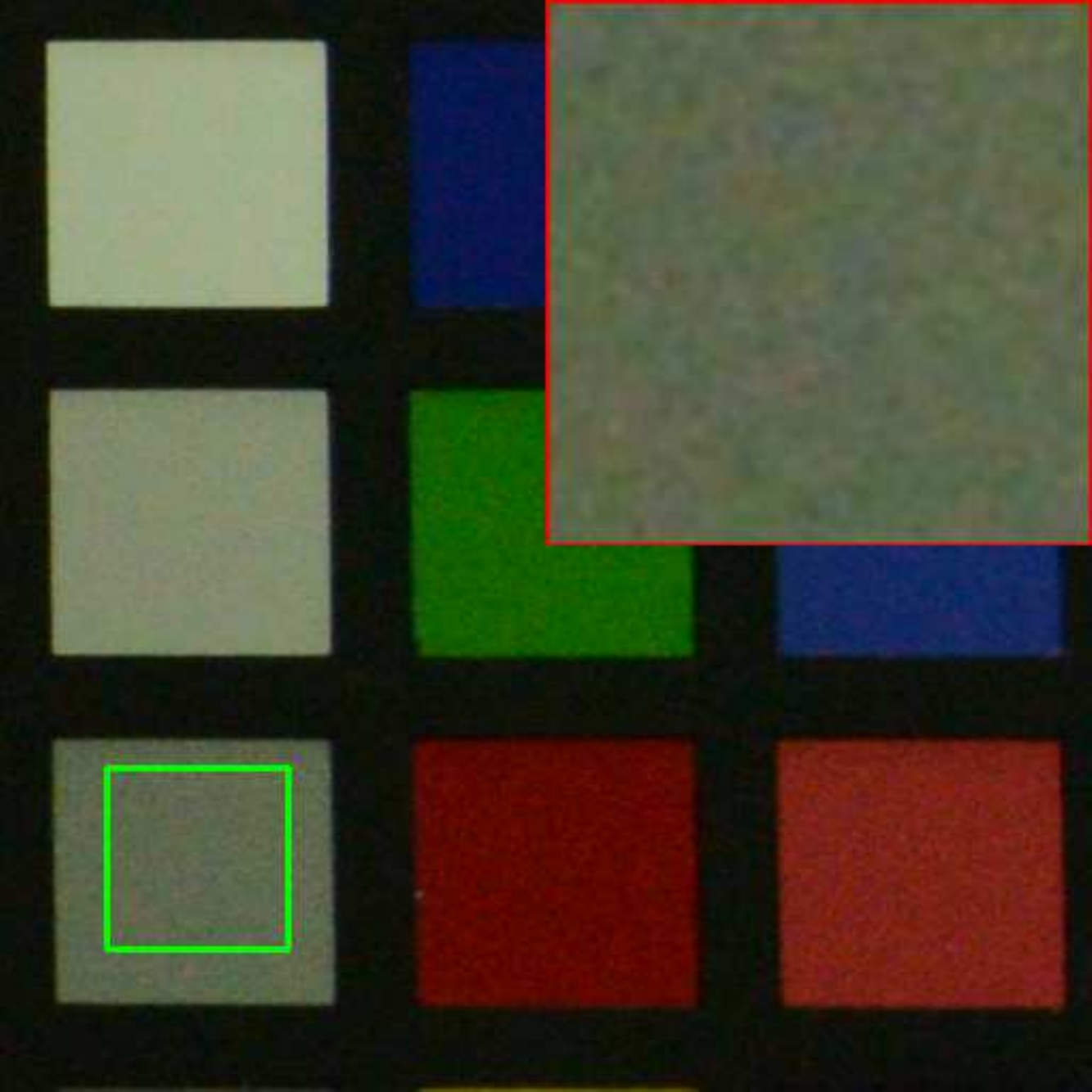}
		\caption{}
	\end{subfigure}
    \centering
	\begin{subfigure}{0.194\linewidth}
		\centering
		\includegraphics[width=0.98\linewidth]{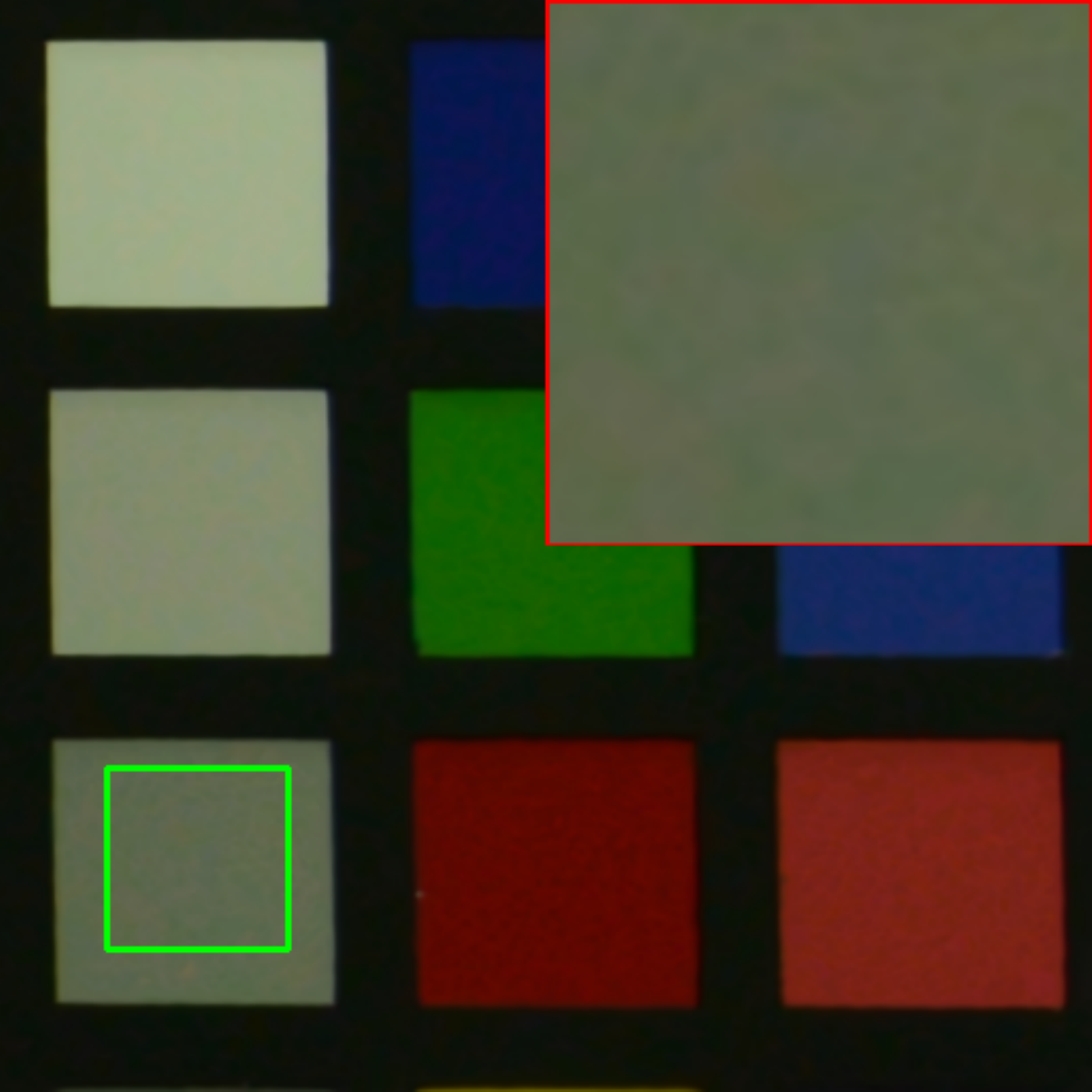}
		\caption{}
	\end{subfigure}
    \centering
	\begin{subfigure}{0.194\linewidth}
		\centering
		\includegraphics[width=0.98\linewidth]{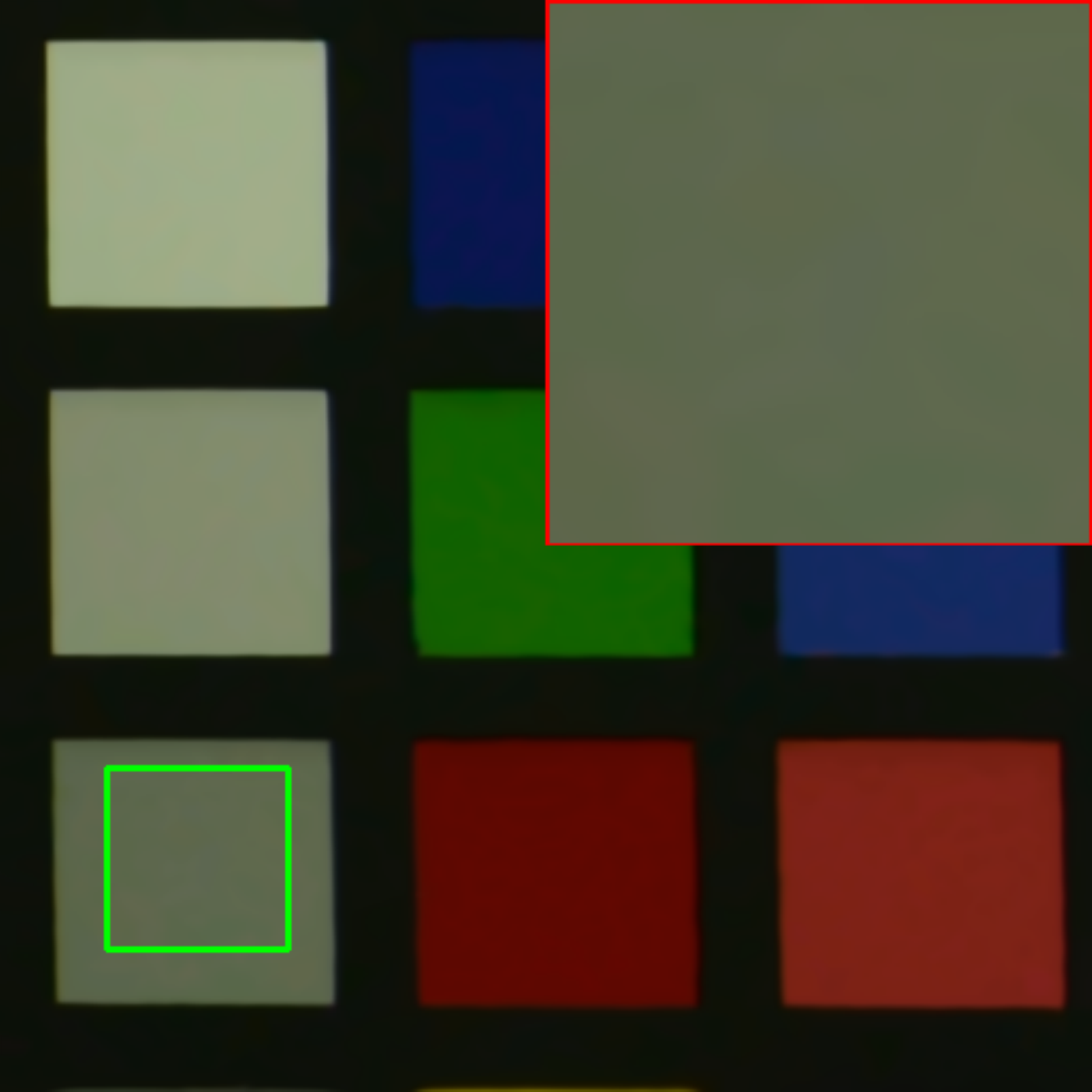}
		\caption{}
	\end{subfigure}
    \centering
	\begin{subfigure}{0.194\linewidth}
		\centering
		\includegraphics[width=0.98\linewidth]{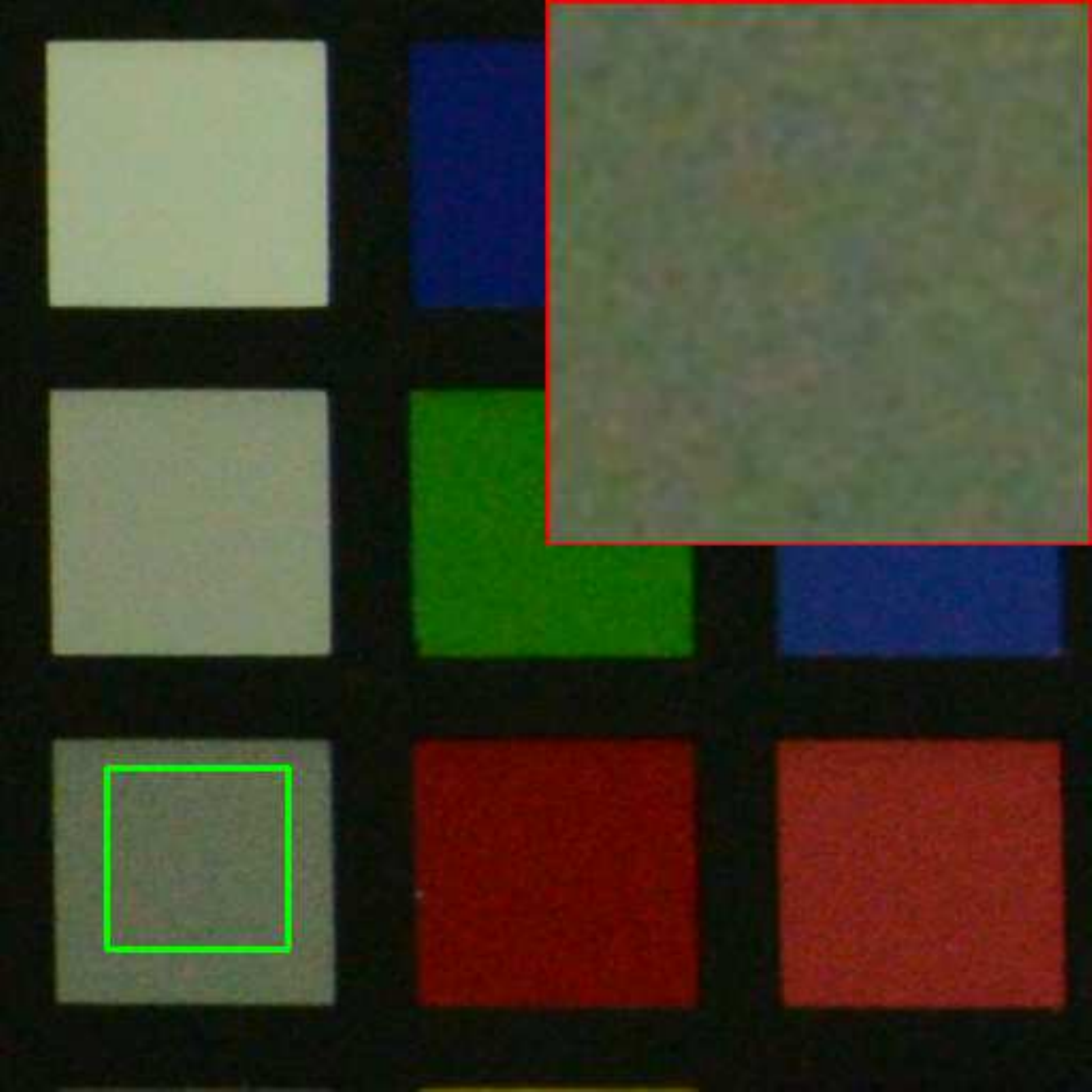}
		\caption{}
	\end{subfigure}
    \centering
	\begin{subfigure}{0.194\linewidth}
		\centering
		\includegraphics[width=0.98\linewidth]{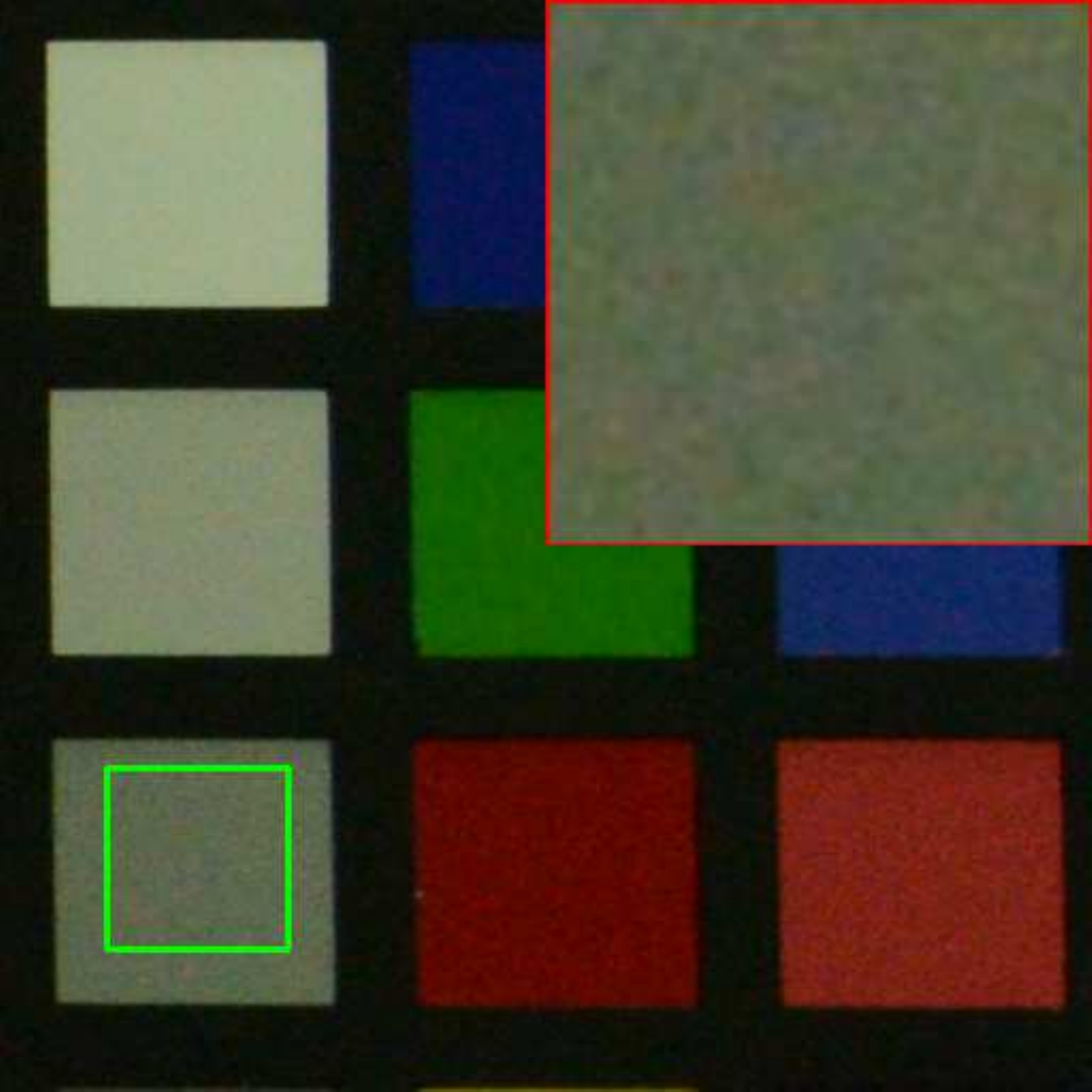}
		\caption{}
	\end{subfigure}
    \centering
	\begin{subfigure}{0.194\linewidth}
		\centering
		\includegraphics[width=0.98\linewidth]{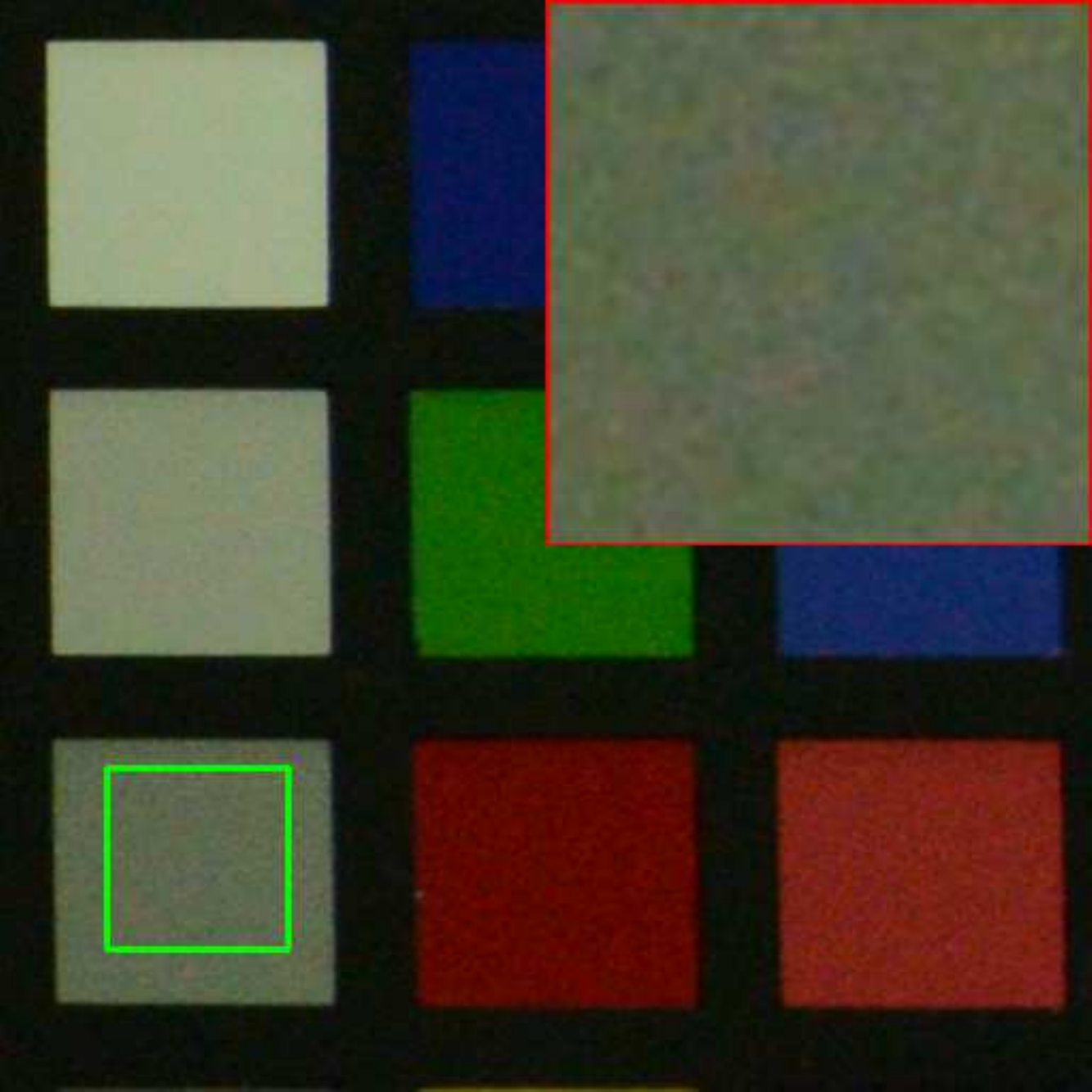}
		\caption{}
	\end{subfigure}
    \centering
	\begin{subfigure}{0.194\linewidth}
		\centering
		\includegraphics[width=0.98\linewidth]{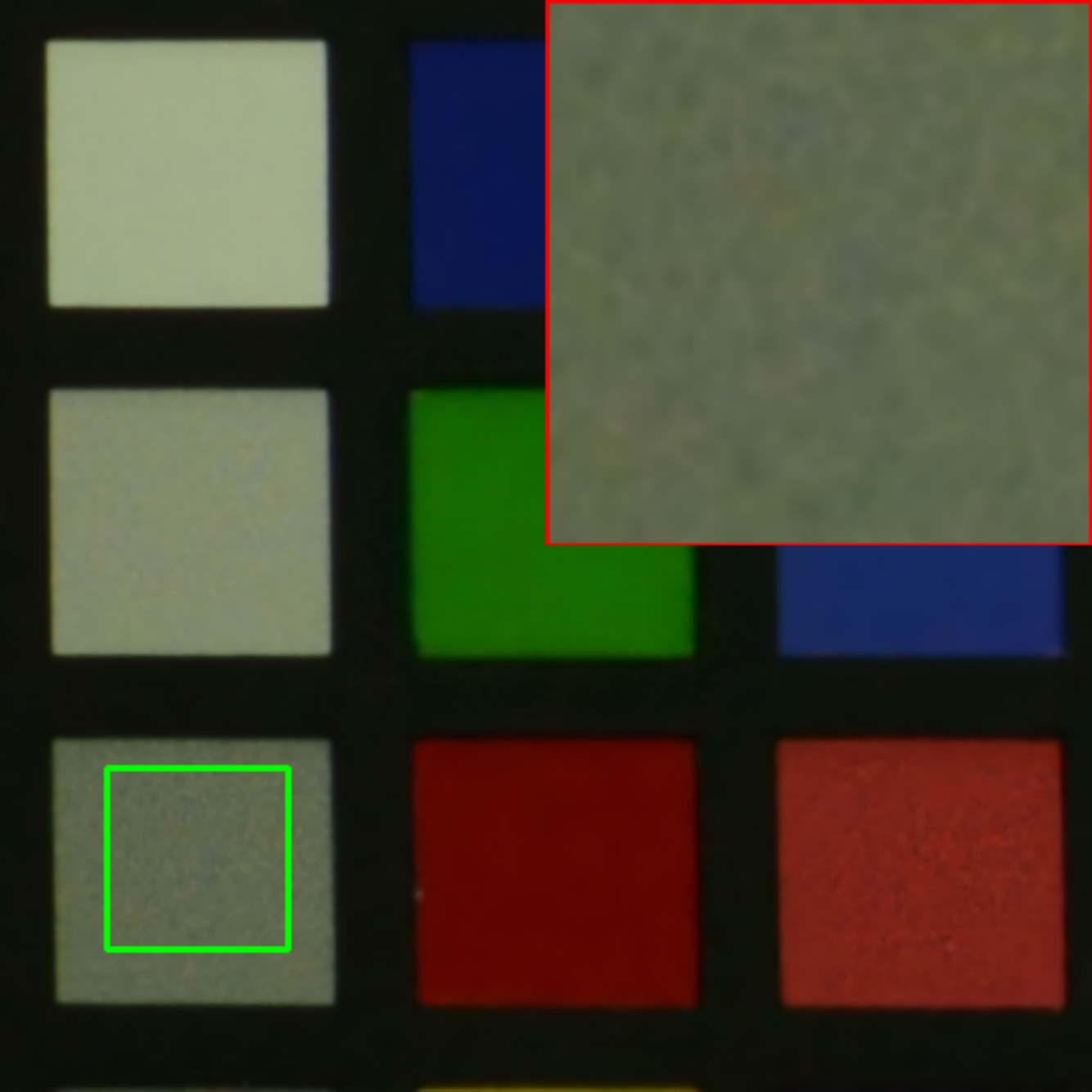}
		\caption{}
	\end{subfigure}
    \centering
	\begin{subfigure}{0.194\linewidth}
		\centering
		\includegraphics[width=0.98\linewidth]{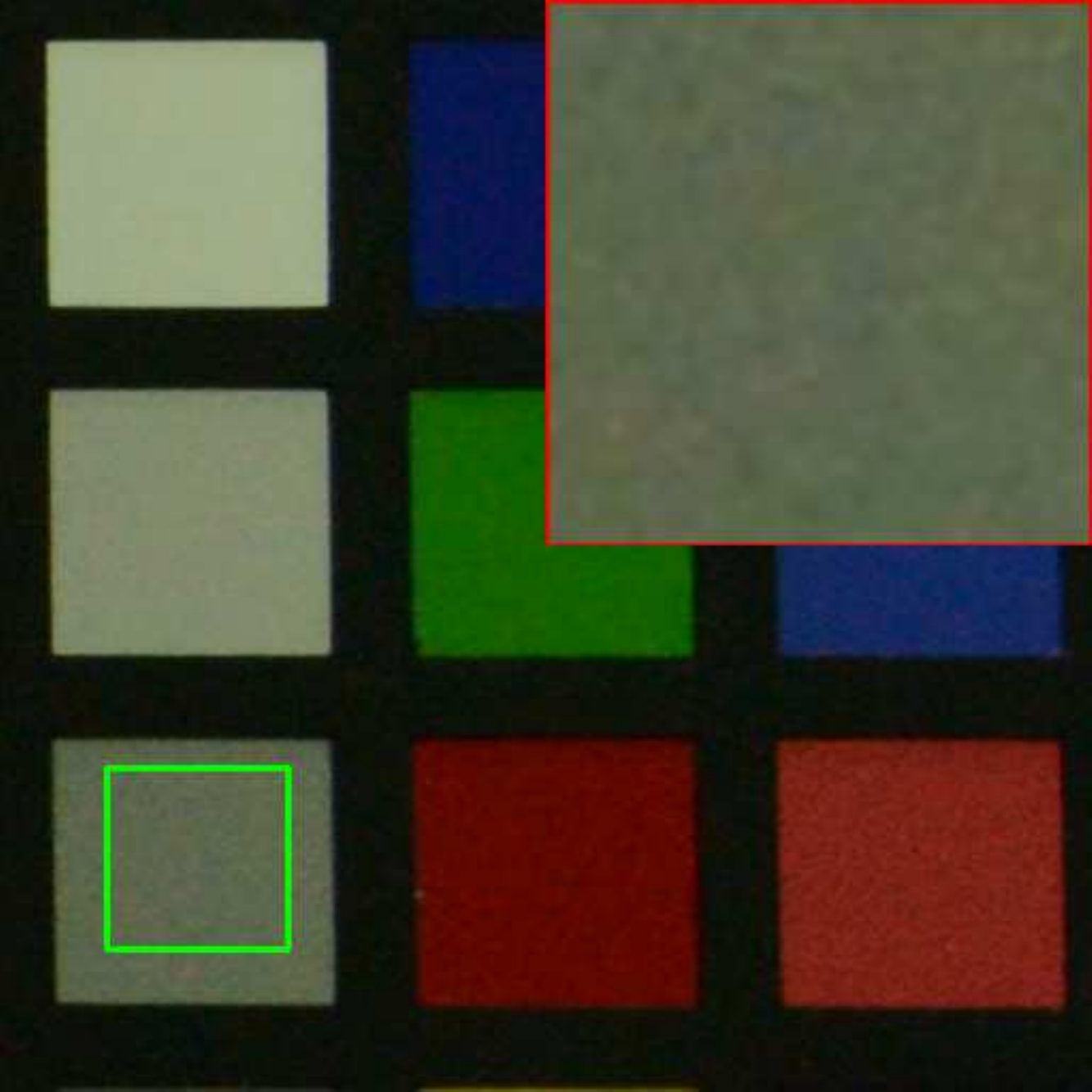}
		\caption{}
	\end{subfigure}
    \centering
	\begin{subfigure}{0.194\linewidth}
		\centering
		\includegraphics[width=0.98\linewidth]{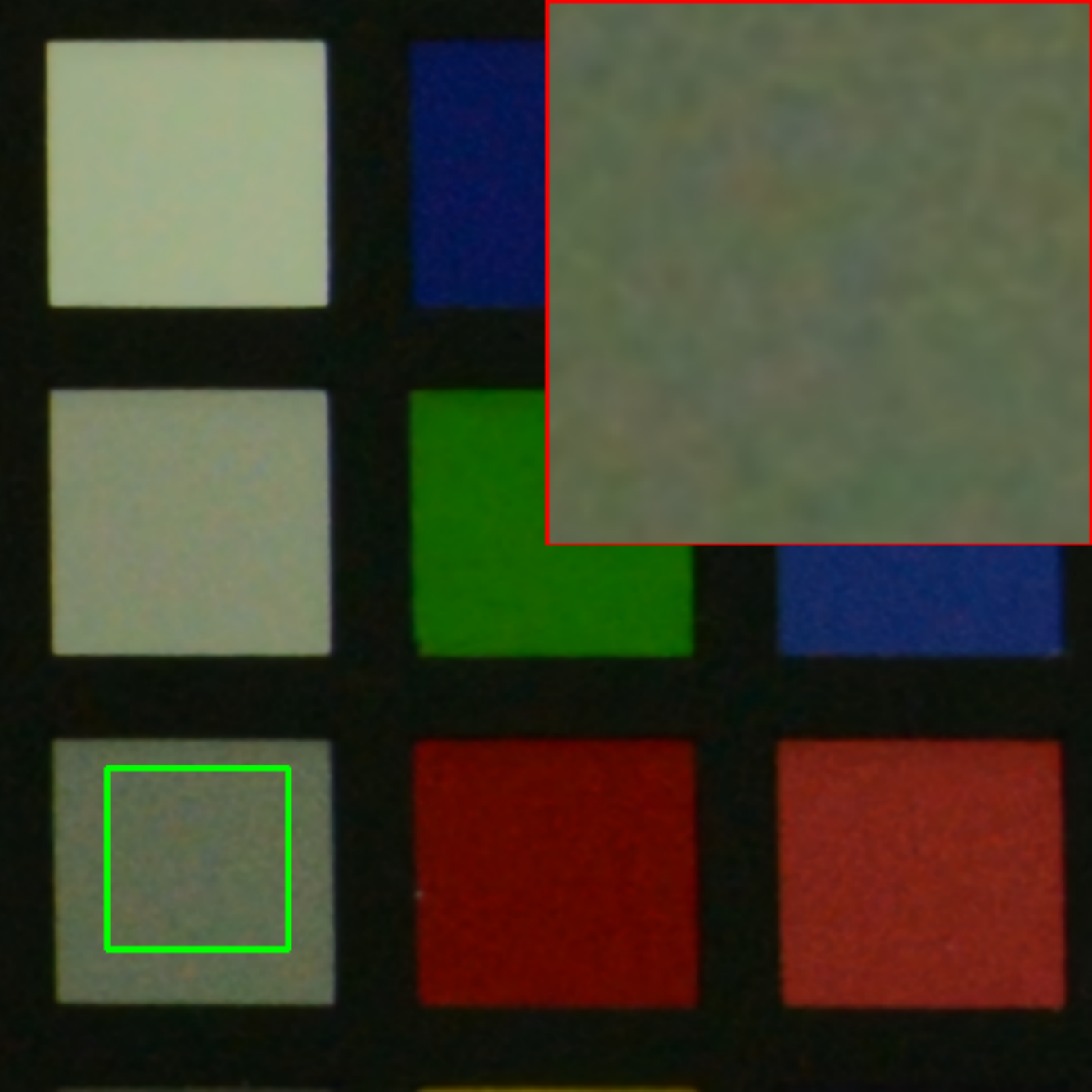}
		\caption{}
	\end{subfigure}
\caption{Denoising results on the image ``Vinegar'' from Nam dataset. (a) Ground-truth image / (PSNR(dB)/SSIM), (b) Noisy image / (34.93/0.841), (c) MCWNNM / (39.51/0.964), (d) TWSC / (41.05/0.983), (e) CDnCNN-B / (35.41/0.856), (f) BUIFD / (35.88/0.868), (g) ADNet / (35.07/0.844), (I) VDN / (37.53/0.967), (i) VDIR / (36.88/0.909), (j) DCBDNet / (38.59/0.954).}
\label{fig:Vinegar}
\end{figure*}

Fig. \ref{fig:Vinegar} displays the denoising results of the compared denoising methods on the image ``Vinegar" from the Nam dataset, where one can see that the MCWNNM, TWSC, VDN, and the proposed DCBDNet obtained better visual quality and PSNR/SSIM values. It also should be noted that even nowadays the DNN based methods are much more popular in image denoising, however the traditional methods like the MCWNNM and TWSC still can obtain remarkable denoising performance in real denoising tasks.

\subsection{Network complexity analysis}
In our experiments, the network complexity was evaluated from the perspectives of model running time, floating point operations per second (FLOPs), and numbers of network parameters. It should be noted that we used the original source codes of the compared denoising methods released by the authors, therefore the BM3D, WNNM, MCWNNM, and TWSC were implemented in Matlab (R2020a) environment, and the DnCNN-B, BUIFD, IRCNN, FFDNet, BRDNet, ADNet, DudeNet, CBDNet, RIDNet, VDN, VDIR, DeamNet, AirNet, DRUNet, and the proposed DCBDNet were evaluated in PyCharm (2021) environment.

We first evaluated the running time of different denoising methods, the results are listed in Table \ref{tab:time}. Three randomly chosen grayscale and color images with different sizes were utilized for evaluating the running time at noise level 25, and the running time on each image of every evaluated model was obtained by averaging the running time of 20 implementations. Our experiments neglected the memory transfer time between CPU and GPU. It can be observed that the denoising speed of BM3D, WNNM, MCWNNM, and TWSC using CPU is much slower than the methods running on GPU. The models can achieve leading PSNR values such as the DeamNet and DRUNet require a longer denoising time, especially when the size of the image is bigger. More importantly, these models equipped with complex structures also require much longer training time.

Table \ref{tab:Parameters_FLOPs} lists the number of parameters and FLOPs of the tested denoising methods on grayscale and color images, respectively. For the network parameters, it can be seen that DnCNN-B, IRCNN, FFDNet, and ADNet have smaller numbers of parameters than our DCBDNet, however the quantitative results of our DCBDNet are superior to these methods. It can be discovered that with a smaller number of parameters, our DCBDNet still can outperform some models with more network parameters in some noise levels, such as the BUIFD, DudeNet, RIDNet, CBDNet, BRDNet, and AirNet. In terms of FLOPs, we also tested the different models on two images of the same size with different channel numbers (grayscale and color). It can be found that the FLOPs of our DCBDNet are lower than most of the state-of-the-art methods, nevertheless our DCBDNet can still obtain competitive denoising performance. The experimental results also demonstrate that our DCBDNet can achieve a desirable balance between model complexity and denoising performance.

\begin{table*}[htbp]
\centering
\caption{Running time (in seconds) of the evaluated denoising methods on three grayscale and color images with different sizes.}
\label{tab:time}
\begin{tabular}{|c|c|c|c|c|c|c|c|}
\hline
\multirow{2}*{Methods} & \multirow{2}*{Device}	& \multicolumn{2}{c|}{$256\times256$}	& \multicolumn{2}{c|}{$512\times512$}	& \multicolumn{2}{c|}{$1024\times1024$}\\
\cline{3-8}
 \multicolumn{1}{|c|}{} &  & Gray & Color & Gray & Color & Gray & Color\\
\hline
BM3D \cite{Dabov2007} & CPU & 0.458	& 0.593	& 2.354	& 3.771	& 9.782	& 12.818\\
\hline
WNNM \cite{Gu2014} & CPU & 63.867 & - & 277.003 & - & 1150.842  & - \\
\hline
MCWNNM \cite{Xu2017} & CPU & - & 62.777 & - & 277.623 & - & 1120.112  \\
\hline
TWSC \cite{XuZ2018} & CPU & 12.314 & 34.41 & 53.155 & 140.964 & 221.507 & 608.492\\
\hline
DnCNN-B \cite{Zhang2017} & GPU & 0.032	& 0.032	& 0.037	& 0.037	& 0.057	& 0.057\\
\hline
BUIFD \cite{Helou2020} & GPU & 0.035	& 0.037	& 0.050	& 0.053	& 0.112	& 0.123\\
\hline
IRCNN \cite{ZhangZ2017} & GPU  & 0.030	& 0.030	& 0.030	& 0.030	& 0.030	& 0.030\\
\hline
FFDNet \cite{Zhang2018} & GPU & 0.031	& 0.030	& 0.031	& 0.030	& 0.032	& 0.030\\
\hline
ADNet \cite{TianX2020} & GPU & 0.031 & 0.033 & 0.035 & 0.045 & 0.051 & 0.093\\
\hline
VDN \cite{YueYZM2019} & GPU & 0.144 & 0.162 & 0.607 & 0.597 & 2.367 & 2.376\\
\hline
VDIR \cite{SohC2022} & GPU & - & 0.385 & - & 1.622 & - & 6.690\\
\hline
DeamNet \cite{Ren2021} & GPU  &  0.054 & - & 0.121 & - & 0.392 & -\\
\hline
AirNet \cite{Li2022} & GPU  &  - & 0.143 & - & 0.498 & - & 2.501\\
\hline
DRUNet \cite{ZhangL2021} & GPU  & 0.068 & 0.068 & 0.106 & 0.107 & 0.276 & 0.280\\
\hline
DCBDNet & GPU & 0.050 & 0.051 & 0.078 & 0.081 & 0.183 & 0.187 \\
\hline
\end{tabular}
\end{table*}

\begin{table}[htbp]
\centering
\caption{The number of model parameters (in K) and the FLOPs (in G) for grayscale and color image denoising of different models.}
\label{tab:Parameters_FLOPs}
\begin{tabular}{|c|c|c|c|c|}
\hline
\multirow{2}*{Methods} & \multicolumn{2}{c|}{Parameters} & \multicolumn{2}{c|}{FLOPs}\\
\cline{2-5}
 \multicolumn{1}{|c|}{}  & Gray & Color & $256\times256\times1$ & $256\times256\times3$\\
\hline
DnCNN-B \cite{Zhang2017} & 666 & 668 & 21.93 & 22.12 \\
\hline
BUIFD \cite{Helou2020} & 1186 & 1196 & 35.31 & 35.65\\
\hline
IRCNN \cite{ZhangZ2017} & 186 & 188 & 6.08  & 6.15\\
\hline
FFDNet \cite{Zhang2018} & 485 & 852 & 3.97 & 3.14\\
\hline
CBDNet \cite{Guo2019} & - & 4365 & - & 20.14\\
\hline
RIDNet \cite{Anwar2019} & 1497 & 1499 & 48.90 & 48.98\\
\hline
VDN \cite{YueYZM2019} & 7810 & 7817 & 24.47 & 24.70\\
\hline
VDIR \cite{SohC2022} & - & 2227 & - & -\\
\hline
BRDNet \cite{Tian2020} & 1113 & 1117 & 36.48 & 36.63\\
\hline
ADNet \cite{TianX2020} & 519 & 521 & 17.08 & 17.15\\
\hline
DudeNet \cite{Tian2021} & 1077 & 1079 & 35.35 & 35.43\\
\hline
DeamNet \cite{Ren2021} & 2226 & - & 72.85 & - \\
\hline
AirNet \cite{Li2022} & - & 8930 & - & 150.64\\
\hline
DRUNet \cite{ZhangL2021} & 32639 & 33641 & 71.71 & 71.79\\
\hline
DCBDNet & 1004 & 1013 & 24.38 & 24.68\\
\hline
\end{tabular}
\end{table}

\section{Conclusion}\label{Conclusion}
In this paper, we propose a novel dual convolutional blind denoising network with skip connection (DCBDNet). The proposed DCBDNet contains a noise estimation network and a dual convolutional neural network (CNN) with two sub-networks. The proposed denoising model incorporates a noise estimation network to estimate the noise level map to enhance its flexibility, which makes the DCBDNet can achieve blind denoising. The dual CNN not only expands the network width to enhance the learning ability of the DCBDNet model, but also can capture the complementary image features for improving denoising performance. In addition, the u-shaped structure and dilated convolution are utilized in two sub-networks respectively to enlarge the receptive fields. Skip connections are adopted in both sub-networks for image feature superposing, and for avoiding the gradient vanishing or exploding. Experimental results illustrate that our proposed DCBDNet can achieve competitive denoising performance both quantitatively and qualitatively compared to state-of-the-art denoising methods. Moreover, the proposed DCBDNet has a shorter running time, fewer FLOPs, and fewer model parameters compared to many other denoising methods. Therefore our proposed DCBDNet can provide an option for practical image denoising. In the future, we aim to further investigate the feature learning ability of the model, especially for the images containing rich repetitive textures.

\section{Acknowledgements}
The work is funded by the Natural Science Foundation of China No. 61863037, No. 41971392, and the Applied Basic Research Foundation of Yunnan Province under grant No. 202001AT070077.
\bibliography{sn-bibliography}
\end{document}